%% file: paper_release.tex
\documentclass{article}
\usepackage[table]{xcolor}
\usepackage{microtype}
\usepackage{graphicx}
\usepackage{subfigure}
\usepackage{booktabs} %

\usepackage{hyperref}

\usepackage{amsmath}
\usepackage{amssymb}
\usepackage{subfiles}
\usepackage{graphicx} %
\usepackage{fancyvrb}
\usepackage{bm}
\usepackage{framed}
\usepackage{tcolorbox}
\usepackage{xcolor}

\usepackage{makecell} 
\usepackage{wrapfig} 
\usepackage{bm}
\usepackage{xspace}

\usepackage{tabularx}
\usepackage{listings}

\makeatletter
\DeclareRobustCommand\onedot{\futurelet\@let@token\@onedot}
\def\@onedot{\ifx\@let@token.\else.\null\fi\xspace}

\def\eg{\emph{e.g}\onedot}

\def\etc{\emph{etc}\onedot}

\makeatother

\newcolumntype{S}{>{\centering\arraybackslash}m{0.9cm}}
\newcolumntype{M}{>{\centering\arraybackslash}m{1.2cm}}
\newcolumntype{L}{>{\centering\arraybackslash}m{1.4cm}}
\definecolor{mygray}{gray}{.95}
\definecolor{mylightergray}{gray}{.99}
\definecolor{mygreen}{RGB}{10, 179, 33}
\usepackage{caption}

\def\p{\emph{protagonist}\xspace}
\def\o{\emph{oracle}\xspace}

\def\tb{\emph{True Belief}\xspace}
\def\fb{\emph{False Belief}\xspace}

\def\tpfo{$\text{T}_p\text{F}_o$\xspace}

\def\fob{\emph{Forward Belief}\xspace}
\def\foa{\emph{Forward Action}\xspace}
\def\bab{\emph{Backward Belief}\xspace}

\newcolumntype{Y}{>{\centering\arraybackslash}X} %

\usepackage[accepted]{icml2024}

\usepackage{amsmath}
\usepackage{amssymb}
\usepackage{mathtools}
\usepackage{amsthm}

\usepackage[capitalize,noabbrev]{cleveref}

\theoremstyle{plain}

\theoremstyle{definition}

\theoremstyle{remark}

\usepackage[textsize=tiny]{todonotes}

\begin{document}

\twocolumn[
\icmltitle{Language Models Represent Beliefs of Self and Others}

\begin{icmlauthorlist}
\icmlauthor{Wentao Zhu}{1}
\icmlauthor{Zhining Zhang}{1}
\icmlauthor{Yizhou Wang}{1,2,3,4}
\end{icmlauthorlist}

\icmlaffiliation{1}{Center on Frontiers of Computing Studies, School of Compter Science, Peking University}
\icmlaffiliation{2}{Inst. for Artificial Intelligence, Peking University}
\icmlaffiliation{3}{Nat'l Eng. Research Center of Visual Technology, Peking University}
\icmlaffiliation{4}{Nat'l Key Lab of General Artificial Intelligence, Peking University}

\icmlcorrespondingauthor{Wentao Zhu}{wtzhu@pku.edu.cn}
\icmlcorrespondingauthor{Yizhou Wang}{yizhou.wang@pku.edu.cn}

\icmlkeywords{Machine Learning, ICML}

\vskip 0.3in
]

\printAffiliationsAndNotice{}  %

\begin{abstract}
Understanding and attributing mental states, known as Theory of Mind (ToM), emerges as a fundamental capability for human social reasoning. While Large Language Models (LLMs) appear to possess certain ToM abilities, the mechanisms underlying these capabilities remain elusive. In this study, we discover that it is possible to linearly decode the belief status from the perspectives of various agents through neural activations of language models, indicating the existence of internal representations of self and others' beliefs. By manipulating these representations, we observe dramatic changes in the models' ToM performance, underscoring their pivotal role in the social reasoning process. Additionally, our findings extend to diverse social reasoning tasks that involve different causal inference patterns, suggesting the potential generalizability of these representations.~\footnote{Project page: \url{https://walter0807.github.io/RepBelief/}}

\end{abstract}

\input{icml2024/chaps/1_intro}

\input{icml2024/chaps/2_related}

\input{icml2024/chaps/3_method}

\input{icml2024/chaps/4_experiments}

\input{icml2024/chaps/5_conclusion}

\clearpage
\bibliography{example_paper}
\bibliographystyle{icml2024}

\newpage
\appendix
\onecolumn

\input{icml2024/chaps/6_appendix}

\end{document}

%% file: icml2024/chaps/1_intro.tex
\section{Introduction}

Developing machine systems that can engage in sophisticated social reasoning in a human-like manner represents one of the paramount goals in artificial intelligence.
At the core of such an endeavor is the necessity for these systems to possess a \emph{``Theory of Mind"} (ToM) capability, which involves recognizing and attributing mental states — such as beliefs, desires, intentions, and emotions — to oneself and others, while acknowledging that others may possess mental states distinct from one's own~\cite{leslie1987pretense, wellman2001meta}.
This foundational capability is crucial not only for the nuanced navigation of human social interactions but also for enabling machines to engage in cooperative, adaptive, and sympathetic behaviors in diverse social environments~\cite{kleiman2016coordinate, rabinowitz2018machine, zhu2023social}. 

The recent advancements in Large Language Models (LLMs) appear to be a promising approach towards this objective, as emerging research indicates that LLMs exhibit reasonable ToM capabilities~\cite{kosinski2023theory, bubeck2023sparks}. These studies suggest that LLMs could, to some extent, predict and understand human intentions and beliefs, thereby demonstrating a foundational level of social reasoning. 
Meanwhile, some other research underscores that these capabilities tend to be superficial and fragile~\cite{shapira2023clever, ullman2023large, ma2023tomchallenges, verma2024theory}. Critics argue that while LLMs may mimic the outward appearance of understanding social contexts and mental states, akin to the ``Clever Hans''~\cite{pfungst1911clever,kavumba2019choosing} and ``Stochastic Parrot''~\cite{bender2021dangers} analogies, this performance may not stem from a deep, genuine comprehension similar to human ToM. Instead, it may simply reflect the models' ability to replicate patterns observed in their training data. 

These observations highlight a critical gap in our understanding of LLM social reasoning capabilities extending beyond mere black-box tests. Key questions remain unanswered, such as whether LLMs develop an internal representation of others' mental states, and whether it is feasible to distinguish between the mental states of others and those of the LLMs when the two have a conflict due to reasons such as information mismatch. 
Addressing these questions not only helps us gain a deeper insight on how LLMs understand others' mental states and perform social reasoning, but is also meaningful for the trustworthiness and alignment of AI systems~\cite{wang2023decodingtrust, ngo2023alignment, ji2023ai}.

In this work, we undertake a preliminary exploration to understand the ToM capabilities of LLMs by studying their internal representations, going beyond merely analyzing the text responses they generate. Firstly, we seek to identify if LLMs have internal representations of others' beliefs and their own (\S~\ref{sec:rep}). If the answer is true, models potentially possess the ability to recognize others' mental states and differentiate them from their own prior to generating a final response. 
Specifically, we prompt the model with a short story in third-person narrative paired with a belief, which may or may not hold true, and attempt to classify the belief's validness from both the main character's perspective and the model's (``God's-eye view''), based on the model's intermediate activations. 
Furthermore, we explore the possibility of modifying the internal representations to steer the model's behavior towards or away from reflecting others' mental states (\S~\ref{sec:intervention}). Lastly, we assess how our findings generalize across various social reasoning tasks with different causal inference patterns (\S~\ref{sec:generalization}).

%% file: icml2024/chaps/2_related.tex
\section{Related Work}

\subsection{Human Theory-of-Mind}

Theory of Mind (ToM), recognized as a cornerstone of human social cognition, facilitates individuals to infer the mental states of others, including their beliefs, desires, and intentions. Research indicates that infants as young as 12 months exhibit the ability to ascribe mental states to others, showcasing early development of ToM \cite{onishi200515, spelke2007core}. The false-belief task~\cite{baron1985does, wellman2001meta} stands as a critical experimental approach for evaluating ToM. In this task, participants are required to predict a protagonist's actions based on her incorrect beliefs, which are separate from the participant's own knowledge.
Cognitive scientists design meticulous experiments to dissect the nuances of reasoning related to agents' desires and beliefs, employing rigorous control conditions to eliminate simplistic heuristic explanations ~\cite{goodman2009cause, baker2009action, baker2017rational, jara2020naive}. Furthermore, neuroscientific studies seek to pinpoint the neural basis of social cognition, particularly highlighting the roles of the dorsal medial prefrontal cortex (dmPFC) and the temporoparietal junction (TPJ) \cite{frith2006neural, dohnel2012functional, molenberghs2016understanding}. ~\citeauthor{jamali2021single} further reveal that single neurons in dmPFC could encode information about others’ beliefs.
The significance of ToM extends beyond individual interactions, influencing the spread of culture and the unity of social groups. The cognitive mechanisms enabled by ToM play a crucial role in forming and sustaining social norms, fostering cooperative behavior, and perpetuating shared cultural practices \cite{tomasello2005understanding, henrich2016secret}.

\subsection{Machine Theory-of-Mind}
Developing machine systems that exhibit human-like ToM ability has been a long-standing endeavor in artificial intelligence research. Notably, ~\citeauthor{rabinowitz2018machine} design a ToMnet that utilizes meta-learning to build models
of the agents it encounters based solely on observations of their behaviors.~\citeauthor{track2018satisficing} introduces the concept of Satisficing Theory of Mind. ~\citeauthor{shum2019theory} explore the application of Bayesian inference to decipher group behaviors and anticipate group dynamics. ~\citeauthor{wang2022tomc} propose to integrate ToM reasoning within multi-agent communication frameworks, enhancing the cooperative capabilities. These studies primarily focus on deducing the mental states of others explicitly and forming neural representations thereof.
The remarkable achievements of LLMs have spurred further exploration into the ToM capabilities of these models. Research works in this field ~\cite{kosinski2023theory,shapira2023clever,ullman2023large} predominantly evaluates model performance using various prompts related to false-belief tests, yielding diverse outcomes.
Additional studies ~\cite{moghaddam2023boosting, wilf2023think} advocate for enhancing the ToM performance of LLMs via strategic prompting.
To facilitate a more uniform assessment of machine ToM abilities, numerous benchmarks have been established, including ToM-QA~\cite{nematzadeh2018evaluating}, ToMi~\cite{le-etal-2019-revisiting}, SocialIQA~\cite{sap2019socialiqa}, BIB~\cite{gandhi2021baby}, Agent~\cite{shu2021agent}, BigToM~\cite{gandhi2023bigtom}, ToMChallenges~\cite{ma2023tomchallenges}, MMToM-QA~\cite{jin2024mmtomqa}, and T4D~\cite{zhou2023far}.
Our research diverges from the prevailing focus by delving into the intrinsic mechanisms of LLM ToM reasoning, specifically through the examination of internal neural representations.

\begin{figure*}[t]
  \centering
  \includegraphics[width=\linewidth]{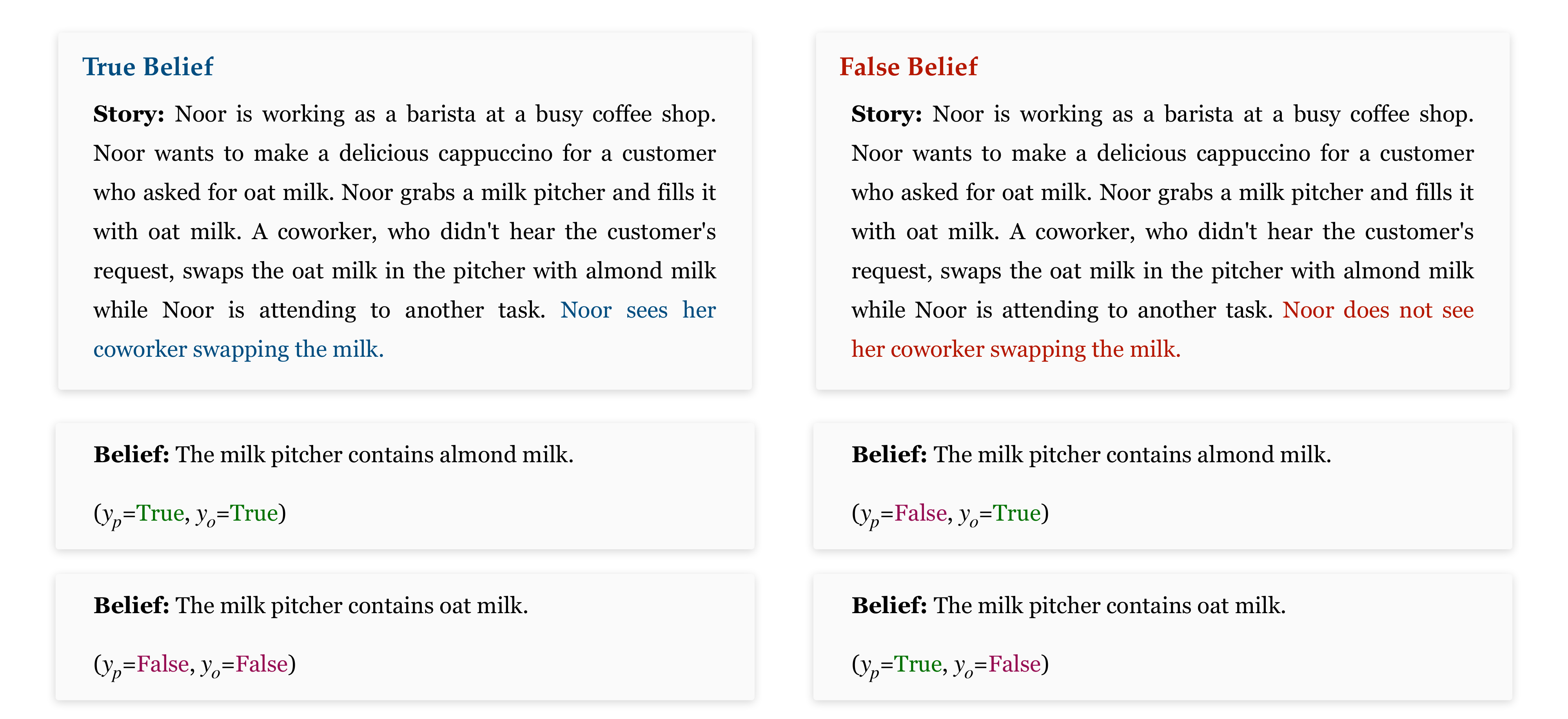}
  \caption{
  Example case of belief representation probing. Left: a ``True Belief'' story where the protagonist shares the same belief with oracle. Right: a ``False belief'' story where the protagonist has different belief with oracle. For both stories, we concatenate them with the two beliefs respectively and directly feed to the model. The ground-truth belief label from the protagonist's perspective $y_p$ and the oracle's perspective $y_o$ can be decided accordingly.
  }
  \label{fig:prompt}
\end{figure*}

%% file: icml2024/chaps/3_method.tex
\begin{figure*}[t]
  \centering
  \includegraphics[width=0.9\linewidth]{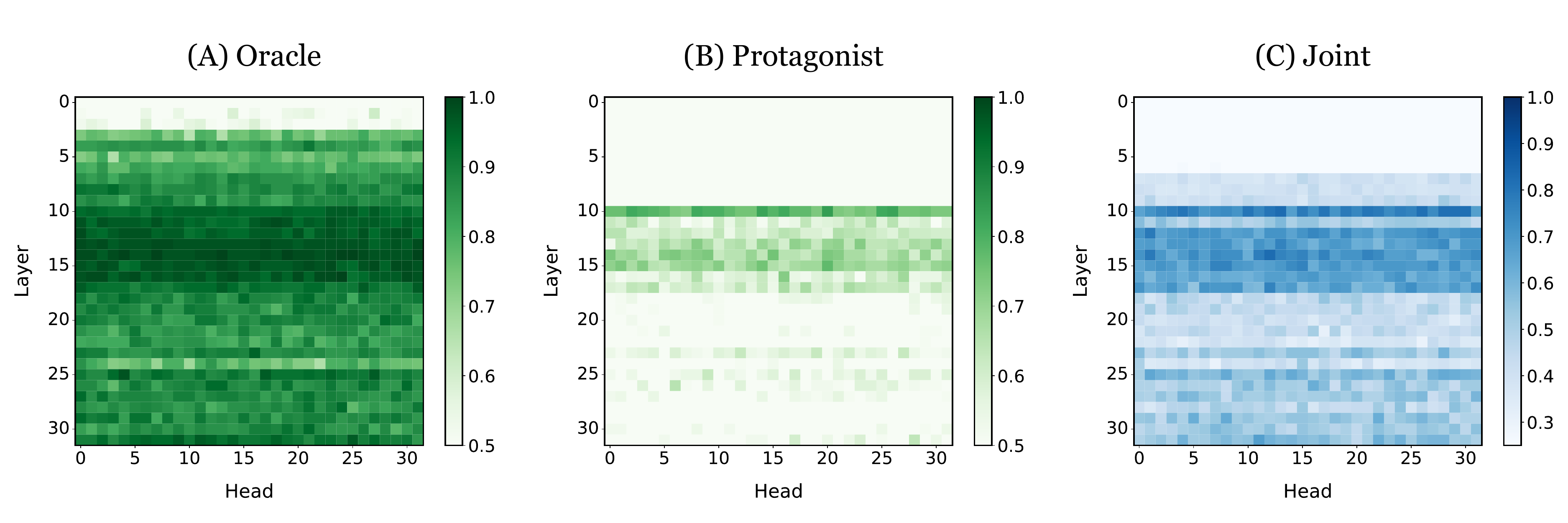}
  \caption{
  Probe accuracies based on the attention head activations in all layers of Mistral-7B. (A) Belief status estimation for \emph{oracle} using logistic regression (binary). (B) Belief status estimation for \emph{protagonist} using logistic regression (binary). (C) Joint belief status estimation for both agents using multinomial logistic regression (quaternary).
  }
  \label{fig:heads}
\end{figure*}

\begin{figure*}[t]
  \centering
  \includegraphics[width=0.85\linewidth]{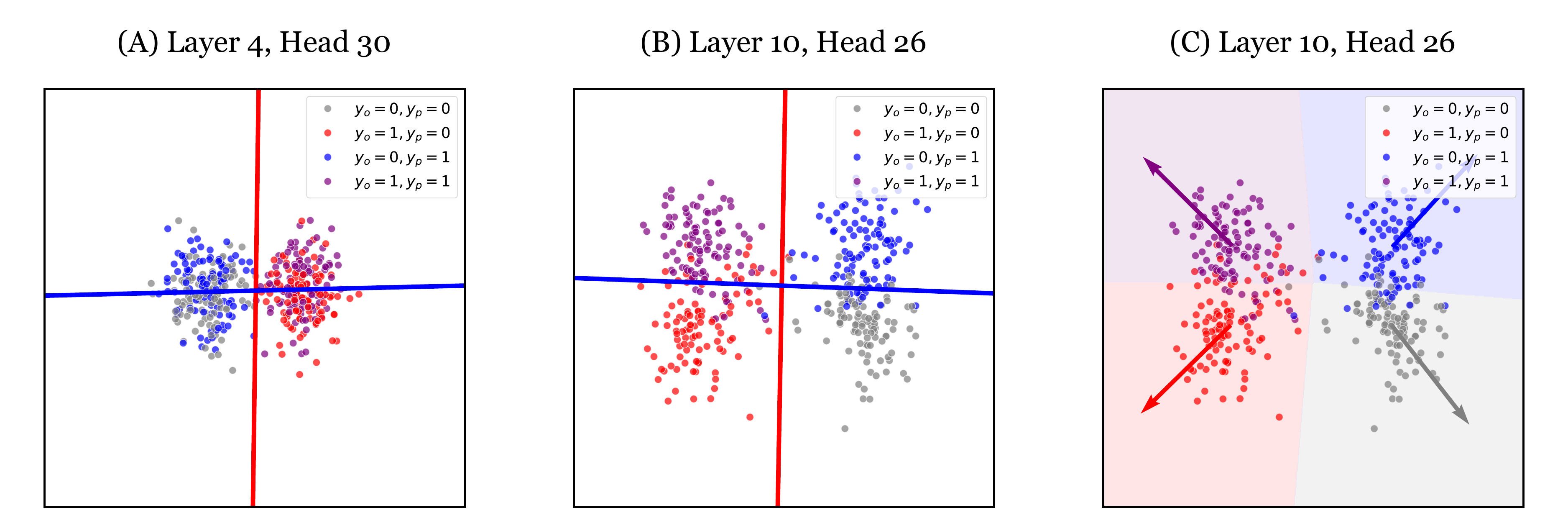}
  \caption{Illustration of linear separability of the belief representations. We show visual explanations for typical representation spaces: In (A), \o belief status can be precisely estimated with a linear model, while \p cannot. The red and blue lines represent linear decision boundaries for \o and \p, respectively; In (B), the belief statuses of both \o and \p can be accurately modeled using linear models; (C) further shows the decision boundaries for joint belief status estimation using multinomial linear regression model, with arrows indicating the probing weight directions for each class.
  }
  \label{fig:cca}
\end{figure*}

\section{Belief Representations in Language Models}
\label{sec:rep}

We first explore if and how LLMs characterize the beliefs of different agents. Previous works in neural network interpretability ~\cite{bau2020units, burns2022discovering, moschella2022relative, li2023emergent} suggest that there often exist interpretable directions in the latent representation space of the model. Therefore, some research works propose to linearly project the learned representation to the target directions to uncover meaningful variables~\cite{mikolov2013linguistic, goh2021multimodal, elhage2022toy, gurnee2023language, park2023linear}.
Motivated by this insight, we start by training linear classifier probes \cite{alain2016understanding, belinkov2022probing} on the latent representations of a language model to estimate the likelihood of a belief from a certain agent's perspective.

\subsection{Setup}

\paragraph{Model.} We employ Mistral-7B-Instruct~\cite{jiang2023mistral} which is an instruction fine-tuned autoregressive language model with state-of-the-art performance. We focus on the activations of self-attention heads that enable Transformer-based language models to transfer information across various token positions~\cite{vaswani2017attention, elhage2021mathematical, todd2023function}. 

\paragraph{Dataset.} We utilize the BigToM dataset~\cite{gandhi2023bigtom} which is constructed with a causal template and an example scenario including prior desires, actions, beliefs, and a causal event that changes the state of the environment. 
The protagonist could be aware or unaware of the causal event, which results in different beliefs. In this section, we focus on the ``Forward Belief'' setting, where the model infers the belief of the agent given the agent’s percepts of the causal event.
We train and evaluate the probes on a held-out subset without access to the stories in the test set of the benchmark.

\subsection{Probing}
\label{sec:probing}

\paragraph{Feature Extraction.}

Our goal is to decode the belief status of different agents from the activations of attention heads, given a narrative and a corresponding belief statement.
Specifically, we focus on two agents, namely \emph{protagonist}, the central figure of the narrative, and \emph{oracle}, which represents an omniscient spectator's perspective.
By prompting the model with pairs of story and belief without explicit directives, we capture the attention head activations at the final token position, denoted as $\bm{X} \in \mathbb{R} ^ {L \times H \times D}$. Here, $L$, $H$, and $D$ represent the number of layers, the number of attention heads per layer, and the dimensionality of the attention head features, respectively. Concurrently, we acquire the corresponding ground-truth belief labels $y_p$ and $y_o$, as illustrated in~\cref{fig:prompt}.

\paragraph{Binary Probing.}

We first train individual linear probes for each attention head at every layer to fit the belief labels $y_p$ and $y_o$ separately. For ease of explanation, we denote the activation of a particular attention head as $\bm{x} \in \mathbb{R}^{N \times D}$ where $N$ is the size of the dataset, and the ground-truth belief labels as $\bm{y} \in \{0, 1\}^N$. We employ a logistic regression model to predict the probability of the belief being true: 

\begin{equation}
\small
    \hat{\bm{y}} = \sigma(\bm{x} \bm{W} + b),
\end{equation}
where $\sigma(\cdot)$ is the logistic sigmoid function, $\bm{W} \in \mathbb{R}^D$ is the weight vector, $b \in \mathbb{R}$ is the bias.
The optimization of parameters $\bm{W}$ and $b$ is achieved through minimizing the cross-entropy loss

\begin{equation}
\small
    \mathcal{L}(\bm{W}, b) = -\frac{1}{N} \left( \bm{y}^T \log(\hat{\bm{y}}) + (\bm{1} - \bm{y})^T \log(\bm{1} - \hat{\bm{y}}) \right).
\end{equation}

\Cref{fig:heads} (A) and (B) display the validation accuracies of the linear probes. It reveals that a large number of attention heads can accurately capture the \o's belief status. These informative attention heads are distributed across various layers, particularly excluding the initial layers, with those in the middle layers demonstrating superior accuracy. It implies that the language model indeed develops intermediate representations that reflect its own belief status based on the full information provided.
In contrast, the majority of attention heads only reach baseline accuracy in predicting the \p's belief status, performing no better than random guess. However, it is worth noting that a specific group of attention heads in the middle layers exhibits remarkably better performance, achieving over 80\% validation accuracy. This phenomenon suggests that these attention heads implicitly encode the belief status of other agents in a linearly-decodable way.

\paragraph{Multinomial Probing.}

We further explore the possibility of estimating the belief statuses of both agents from the activations simultaneously. For this purpose, we combine the belief labels of both agents into a four-dimensional variable representing their joint belief statuses using one-hot encoding, where each dimension corresponds to a unique combination of $y_o$ and $y_p$. For brevity, we define this joint belief variable as $\bm{y}_m \in \{0, 1\}^{N \times 4}$. A multinomial logistic regression model is utilized to predict the joint belief $\hat{\bm{y}_m}$. Specifically, the class probabilities are derived by applying the softmax function to the linear transformations of $\bm{x}$:

\begin{equation}
\hat{\bm{y}_m} = \text{softmax}(\bm{x} \bm{W}_m + \bm{b}_m),
\end{equation}
where $\bm{W}_m \in \mathbb{R}^{D \times 4}$ is the weight matrix for multinomial logistic regression, and $\bm{b}_m \in \mathbb{R}^4$ is the bias vector. These parameters can be optimized by minimizing the cross-entropy loss of all classes

\begin{equation}
\mathcal{L}(\bm{W}_m, \bm{b}_m) = -\frac{1}{N} \text{Tr} \left( \bm{y}_m^{T} \log(\hat{\bm{y}_m}) \right).
\end{equation}

\Cref{fig:heads} (C) illustrates the accuracies of multinomial probing on the validation set, which demonstrates that it is possible to train linear probes with decent accuracy in the quaternary classification task for specific attention heads. In other words, there exist individual attention heads (mostly in the middle layers) that could encode the belief statuses of both agents not only independently but also in conjunction, such as indicating whether \p and \o share identical beliefs. This observation underscores the nuanced capability of these attention heads to represent complex relational information between agents' beliefs. We present additional statistical analysis on belief probing in~\cref{sec:stats}.

\paragraph{Visualizing the Belief Representations.}
In order to better understand the belief representations in the attention head activation space, we further visualize the linear regressors. 
We perform canonical-correlation analysis (CCA) to reduce the dimensionality of the activations to two and plot the linear decision boundaries in the reduced space. ~\Cref{{fig:cca}} demonstrates two representative categories of the attention heads. The first category predominantly encodes the belief of \o, showing a bias toward this perspective. The second category, on the other hand, more comprehensively captures the beliefs of both agents (B). We further visualize the decision boundaries of multinomial probes in (C). 
Although geometries in the high-dimensional space can be much more complicated, the 2D visualizations offer some basic intuitions of linearly-separable belief representations.
\Cref{fig:heads} reveals that, although a significant portion of the attention heads fall into the first category, a distinct subset aligns with the second category.

%% file: icml2024/chaps/4_experiments.tex
\section{Manipulating the Belief Representations}

Although the probing results support the presence of belief representations for different agents within the attention head activation spaces, it remains unclear if these representations contribute to the overall social reasoning process. In this section, we aim to explore the functional roles of belief representations by explicitly manipulating them. We design experiments to address the following questions: Can we alter the social reasoning capabilities of language models by manipulating their internal representations? If so, how can this be achieved? And, how does the practice impact different types of social reasoning tasks?

\subsection{ToM Evaluation}
\begin{figure}[t]
  \centering
  \includegraphics[width=\linewidth]{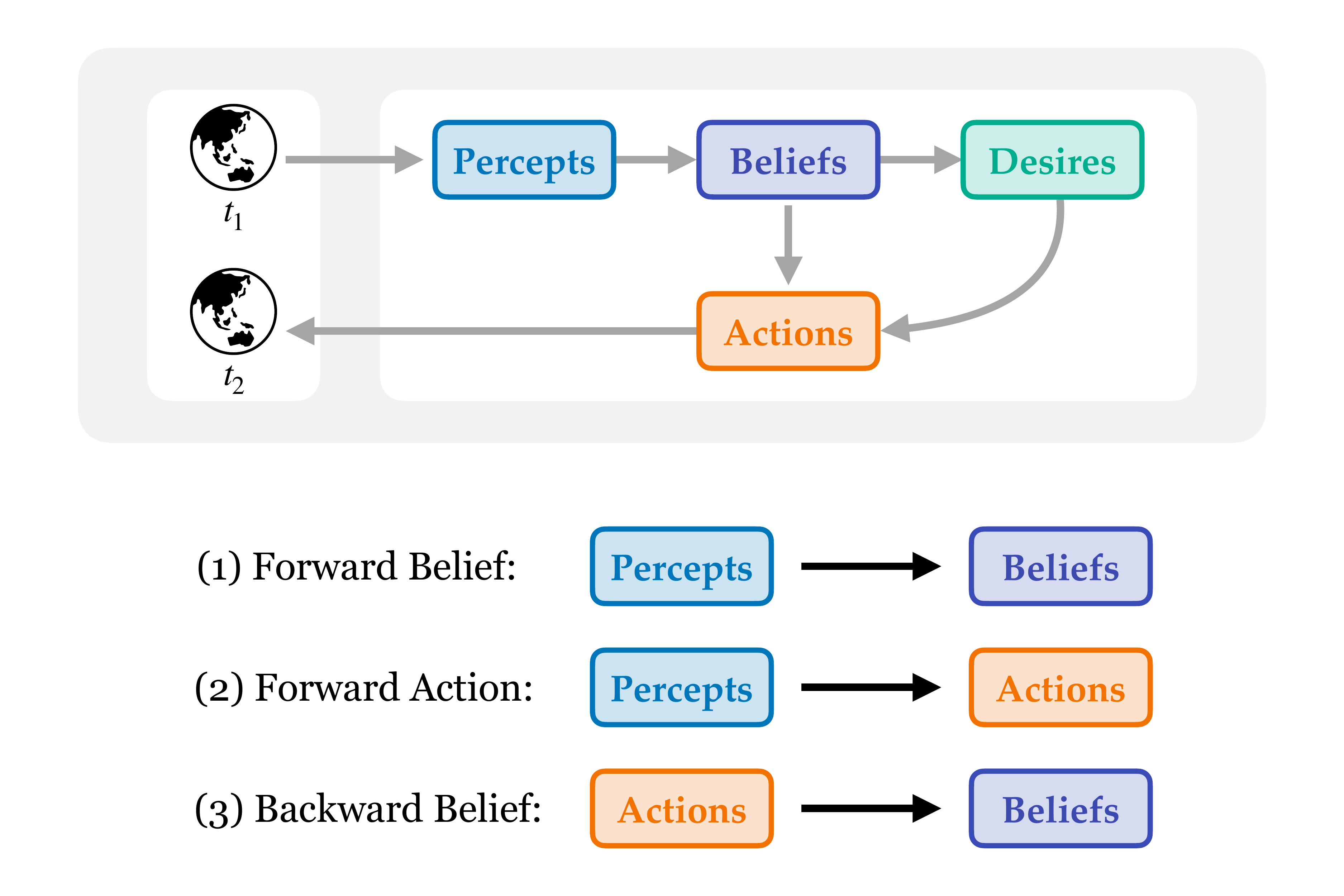}
  \caption{Different social reasoning tasks and the underlying causal graph.}
  \label{fig:causal}
  \vspace{-0.3cm}
\end{figure}

We evaluate the ToM capabilities of language models using the BigToM~\cite{gandhi2023bigtom} benchmark. We focus on the 0-shot setting and do not explicitly reveal the agent’s initial belief. We study three social reasoning tasks, namely \fob, \foa, and \bab, each focuses on different causal inference patterns as shown in~\cref{fig:causal}.

\begin{enumerate}
\setlength\itemsep{0em} %
    \item The \fob task entails deducing the agent's beliefs given its percepts of a causal event. This inference can be expressed as: $P(\text{Belief} \mid \text{Percept})$. 
    \item The \foa task involves predicting the agent's future action based on the percepts. This process implicitly requires an initial inference of the agent's beliefs, followed by the deduction of the agent's action considering both percepts and desires: $\sum_{\text{Belief}} P(\text{Action} \mid \text{Percept}, \text{Desire}, \text{Belief})$.
    \item The \bab task aims to infer the agent's beliefs from the observed actions. This task poses a significant challenge as it demands a joint inference of unknown beliefs and percepts based on an observed action: $\sum_{\text{Percept}} \sum_{\text{Belief}} P(\text{Action} \mid \text{Desire}, \text{Percept}, \text{Belief})$.
\end{enumerate}

All the tasks are presented as reading comprehension with a story in third-person narrative, followed by a question and two options. We evaluate the models based on their accuracy in responding to these questions, specifically under two conditions for each narrative scenario: \tb (TB) and \fb (FB). We also evaluate the percentage of scenarios where the model correctly answers both TB and FB questions.

\begin{table*}[t]
\centering
\caption{Model performance comparison on the BigToM benchmark. TB = True Belief. FB = False Belief. Bold items denote the best setting in each subset.}
\vspace{1em}
\label{tab:bigtom}
\footnotesize
\begin{tabularx}{\linewidth}{l@{\hspace{10pt}}Y@{\hspace{10pt}}YYY@{\hspace{10pt}}YYY@{\hspace{10pt}}YYY}
\toprule
Model & \multicolumn{3}{c}{Forward Belief} & \multicolumn{3}{c}{Forward Action} & \multicolumn{3}{c}{Backward Belief} \\
\cmidrule(r){2-4} \cmidrule(lr){5-7} \cmidrule(l){8-10}
& TB & FB & Both & TB & FB & Both & TB & FB & Both \\
\midrule
LLaMA-65b & 0.68 & 0.62 & 0.51 & 0.82 & 0.47 & 0.45 & 0.56 & 0.53 & 0.40 \\
text-davinci-003   & 0.82 & 0.82 & 0.65 & 0.96 & 0.27 & 0.25 & 0.54 & 0.59 & 0.24 \\
Claude      & 0.97 & 0.82 & 0.81 & 0.98 & 0.28 & 0.27 & 0.79 & 0.48 & 0.33 \\
Claude-2    & 0.88 & 0.75 & 0.68 & 0.95 & 0.36 & 0.34 & 0.75 & 0.50 & 0.39 \\
GPT-3.5   & 0.81 & 0.69 & 0.53 & 0.97 & 0.19 & 0.17 & 0.55 & 0.45 & 0.18 \\
GPT-4     & 0.99 & 0.98 & 0.97 & 0.98 & 0.81 & 0.79 & 0.86 & 0.53 & 0.40 \\
\midrule
Mistral-7B (baseline) & 0.95 & 0.33 & 0.31 & 0.92 & 0.30 & 0.26 & \textbf{0.91} & 0.22 & 0.16 \\
Mistral-7B (+ random) & \textbf{0.97} & 0.33 & 0.32 & 0.92 & 0.29 & 0.25 & 0.91 & 0.19 & 0.14 \\
Mistral-7B (+ protagonist) & 0.96 & 0.30 & 0.29 & 0.91 & 0.30 & 0.25 & 0.90 & 0.22 & 0.15 \\
Mistral-7B (- oracle) & 0.84 & 0.49 & 0.41 & \textbf{0.93} & 0.29 & 0.25 & 0.50 & 0.37 & 0.26 \\
Mistral-7B (+ \tpfo) & 0.85 & \textbf{0.66} & \textbf{0.58} & 0.88 & \textbf{0.41} & \textbf{0.31} & 0.61 & \textbf{0.44} & \textbf{0.41} \\
\midrule
DeepSeek-7B (baseline) & \textbf{0.73} & 0.47 & 0.37 & \textbf{0.77} & 0.48 & 0.31 & 0.64 & 0.50 & 0.29 \\
DeepSeek-7B (+ random) & 0.72 & 0.49 & 0.40 & 0.76 & 0.47 & 0.30 & \textbf{0.65} & 0.49 & 0.26 \\
DeepSeek-7B (+ protagonist) & 0.73 & 0.50 & 0.42 & 0.76 & 0.49 & 0.31 & 0.65 & 0.52 & 0.27 \\
DeepSeek-7B (- oracle) & 0.65 & 0.46 & \textbf{0.44} & 0.73 & 0.57 & 0.34 & 0.65 & 0.48 & 0.28 \\
DeepSeek-7B (+ \tpfo) & 0.63 & \textbf{0.74} & 0.38 & 0.75 & \textbf{0.60} & \textbf{0.39} & 0.54 & \textbf{0.66} & \textbf{0.31} \\
\bottomrule
\end{tabularx}
\end{table*}

\begin{figure*}[t]
  \centering
  \includegraphics[width=0.95\linewidth]{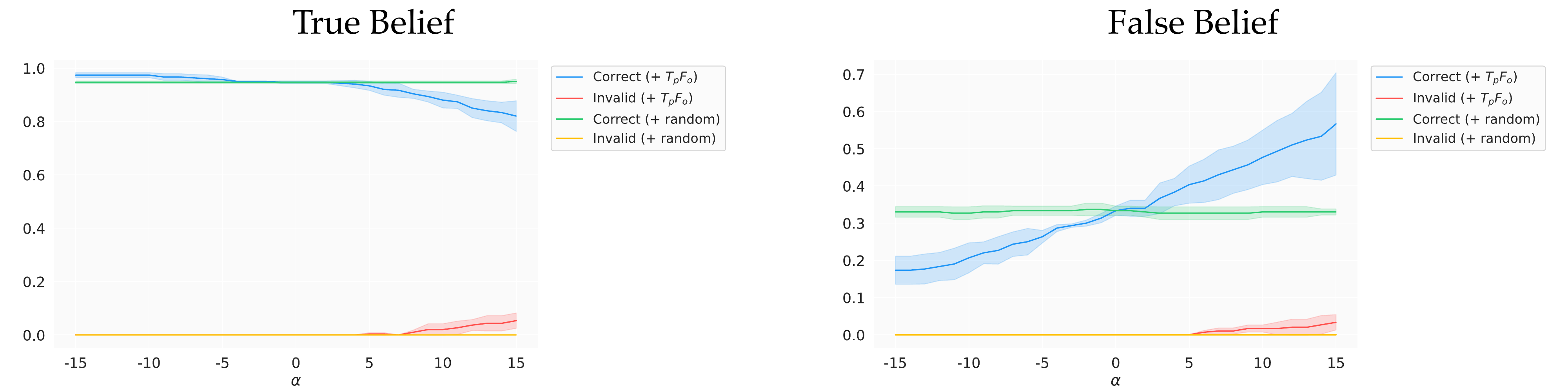}
  \caption{Impact of varying intervention strength $\alpha$ on the \fob task using Mistral-7B. ``Invalid'' denotes the answer is not recognized by the grading mechanism as the model fails to provide answer in the required format, \eg, delivering uncertain responses.}
  \label{fig:alphas}
\end{figure*}

\begin{figure*}[t]
  \centering
  \includegraphics[width=\linewidth]{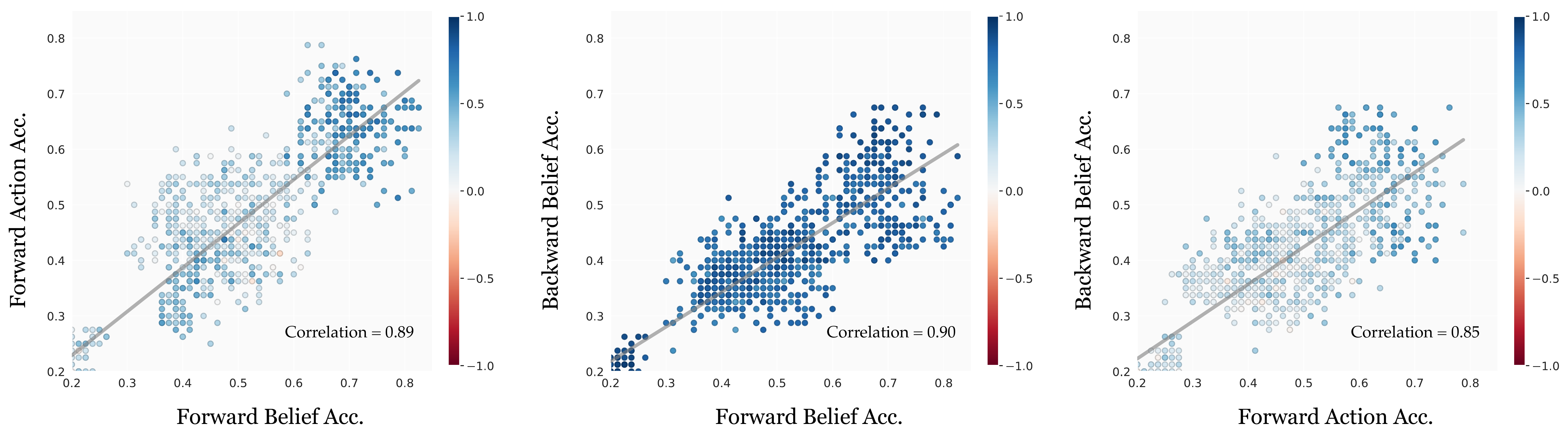}
  \caption{Pairwise comparisons of multinomial probing results on Mistral-7B. Each point represents a specific attention head. The point position denotes its probe accuracies in the two tasks, and point color denotes the cosine similarity between the (+ \tpfo) probe directions of the two tasks.}
  \label{fig:tasks}
\end{figure*}

\begin{figure}[t]
  \centering
  \includegraphics[width=\linewidth]{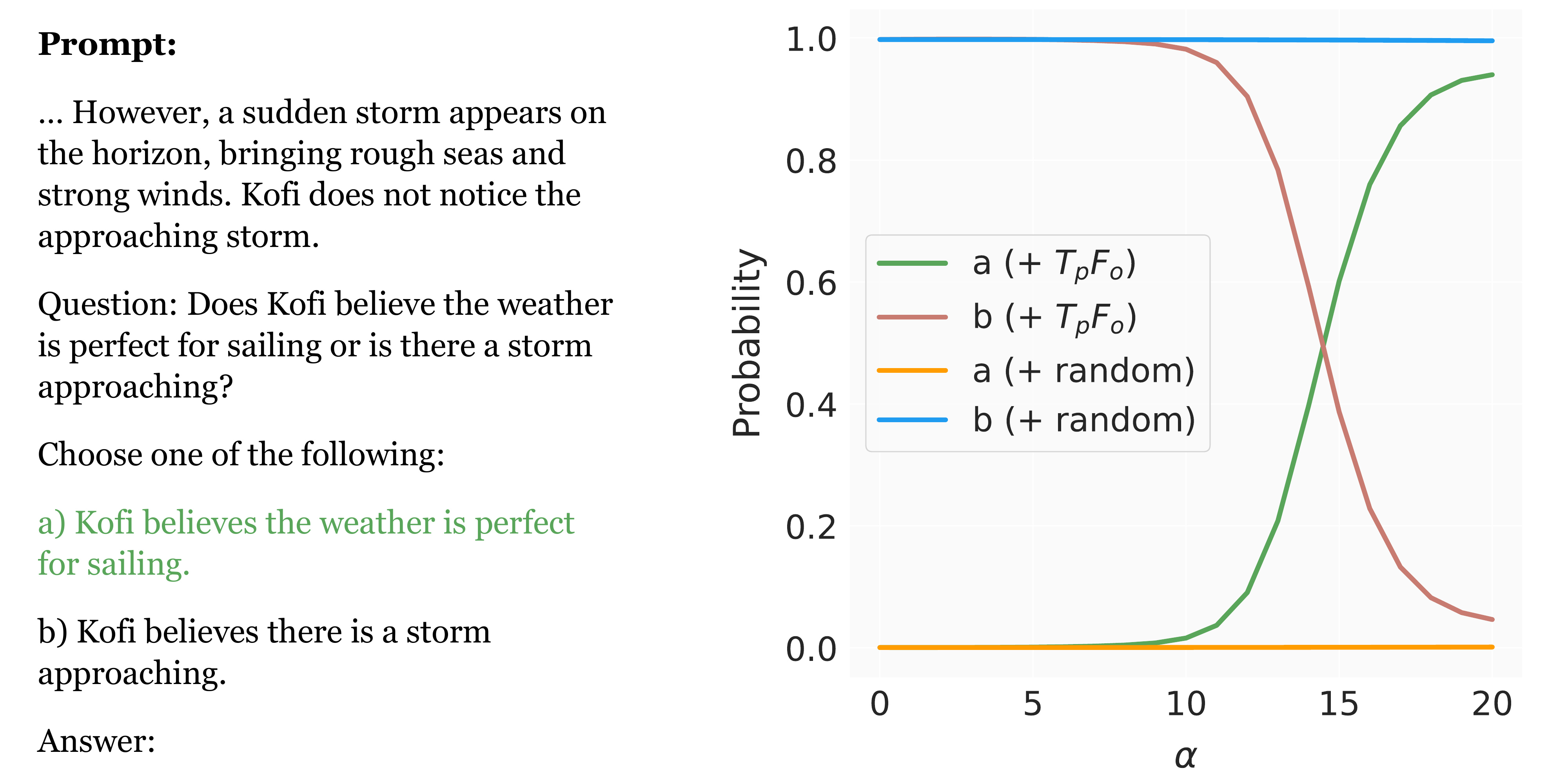}
  \caption{(Left) Question prompt of the \fb condition for the \fob task. The story background is omitted for simplicity. (Right) Changes of next-token probability with regard to different intervention strength $\alpha$ on Mistral-7B.  
  }
  \label{fig:logits}
\end{figure}

\begin{figure*}[t]
  \centering
  \includegraphics[width=\linewidth]{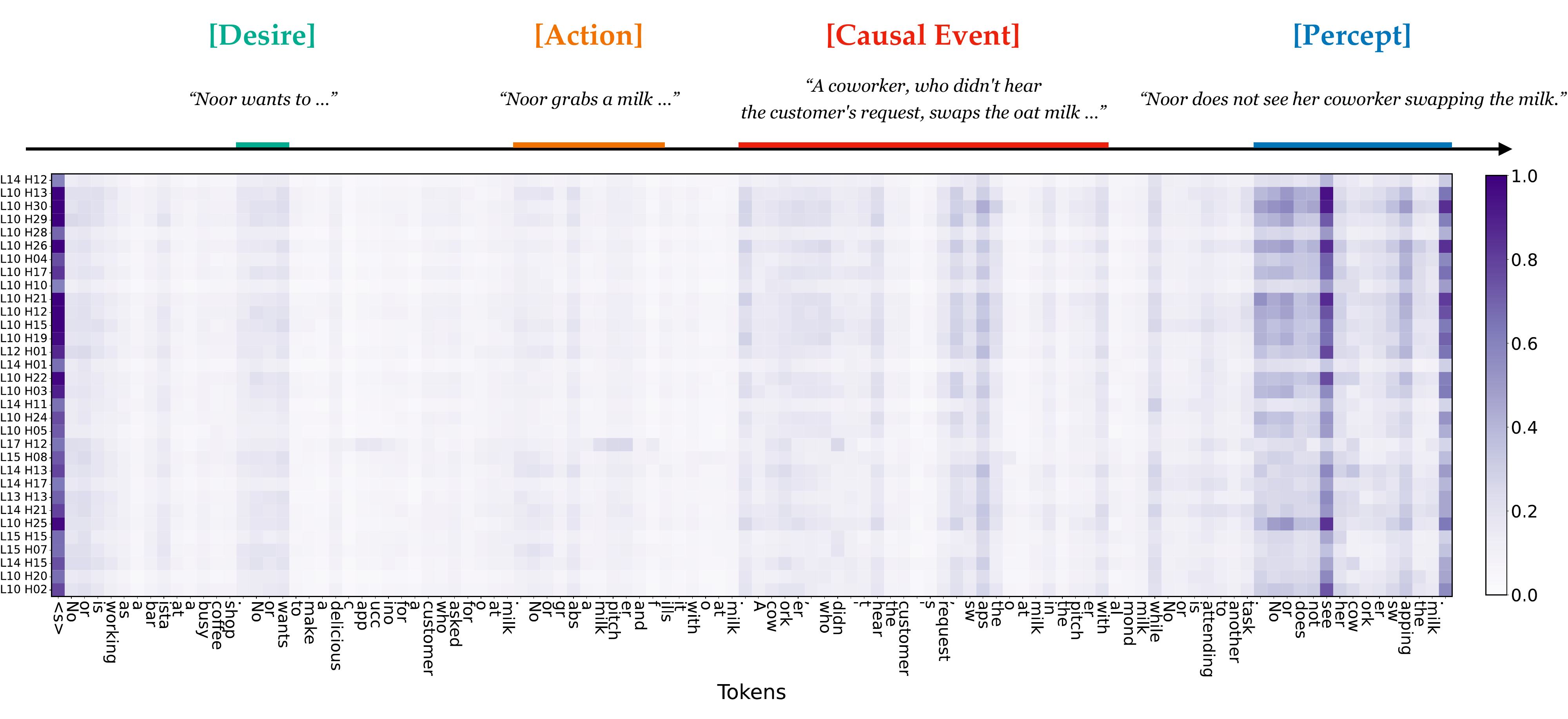}
  \caption{Magnitudes of gradients on the token embeddings with respect to the projection of attention head activations over the corresponding joint belief directions. Each line represents a specific attention head in Mistral-7B. We highlight the prominent segments and the corresponding causal variables.
  }
  \label{fig:tokens}
\end{figure*}

We conduct experiments on two language models, Mistral-7B-Instruct~\cite{jiang2023mistral} and DeepSeek-LLM-7B-Chat~\cite{bi2024deepseek}. Both models are tested using the most deterministic setting with a temperature of $0$ following ~\cite{gandhi2023bigtom}. As the baseline results in~\cref{tab:bigtom} show, both models exhibit a distinct performance gap between \tb and \fb conditions when tested directly. It suggests that the models fail to recognize that other agents may hold beliefs different from their own due to perception differences.
Specifically, in the classical \fb test, Mistral is more biased towards the wrong answer, while DeepSeek's choices are closer to random guess. These tendencies might be attributed to the biased internal belief representations as previously discovered in \S~\ref{sec:probing}. The probe accuracies for DeepSeek are presented in~\cref{sec:probe}.

\subsection{Activation Intervention}
~\label{sec:intervention}
\subsubsection{Strategy.} We apply inference-time intervention~\cite{li2023inferencetime} to manipulate the activations at multi-head attention (MHA) stage of the Transformer models. It involves first selecting the top-$K$ heads based on their probing accuracy on the validation set, then steering their activations towards certain directions for $\alpha \times$ the standard deviation for next token prediction autoregressively. 
Mathematically, the intervention for the $l$-th layer of can be written as 

\begin{equation}
    x_{l+1} = x_l + \sum_{h=1}^H Q_l^h \left( \text{Att}_l^h(P_l^h x_l) + \alpha \sigma_l^h \theta_l^h \right),
\end{equation}   

where $x_i$ is the stream activation of the $i$-th layer, $H$ is the number of attention heads within the layer. For each head $h$, $P_l^h$ maps stream activation into a lower-dimensional head space, and $Q_l^h$ maps it back. $\text{Att}$ is an operator where communication with other input tokens happens. The intervention happens after $\text{Att}$ and before $Q_l^h$, where $\alpha$ is the step length of intervention, $\sigma_l^h$ is the standard deviation of activations along the target direction, $\theta_l^h$ is the target direction.
We set $K$ and $\alpha$ with grid search following previous works, and present the ablations in~\cref{sec:hyper}. Our primary focus is on identifying effective directions for altering the model behavior purposefully. Specifically, we explore the usage of the following directions: 

\begin{itemize}
\setlength\itemsep{0em} %
    \item Random directions within the activation spaces $\mathbb{R}^{D}$.
    \item Weight directions for \o and \p, respectively, derived from binary probing. We focus on the direction to maximize the probability of predicting \p's belief as True (+ protagonist, corresponding to the upwards direction vertical to the blue boundary in~\cref{fig:cca} (B)), and the direction to minimize the probability of predicting \o's belief as True (- oracle, corresponding to the rightwards direction vertical to the red boundary in~\cref{fig:cca} (B))

    \item Weight directions for joint belief status prediction, derived from multinomial probing. We focus on the direction which maximizes the probability of recognizing \p's belief as True and meanwhile distinguishes it from \o's belief (+ \tpfo, corresponding to the blue arrow in~\cref{fig:cca} (C)).
\end{itemize}

For random directions, we use the top-$K$ informative heads identified with multinomial probing.
For different social reasoning tasks, we separately probe the attention head activations by constructing the corresponding prompt templates. For example, for the \foa task, we utilize the story-action pairs, while for the \bab task, we include the \p's next-step actions in the story. More details can be found in~\cref{sec:prompts}.

\subsubsection{Results.} We evaluate different activation intervention strategies and present the results in Table~\ref{tab:bigtom}. For different tasks, we perform intervention based on their respective probing results.
The (+ random) results indicate that random perturbations of attention heads have marginal impact on model performance, which is in line with the findings in ~\cite{li2023inferencetime}. 
Directly targeting activations towards the \p belief direction also fails to significantly change the model performances, possibly due to overpowering \o belief signals with heavily-biased belief representation. To investigate this, we conduct intervention along the \o belief direction reversely (- oracle), and discover a noteworthy change of model behaviors. Furthermore, we explore a direction that could distinguish between the belief of both agents, amplifying the \p's belief likelihood and weakening the \o's belief likelihood. This direction could be derived from the corresponding dimension of the weight matrix in the multinomial logistic regression probes. We find that intervention towards this direction (+ \tpfo) remarkably changes the model performance, effectively improving the overall ToM reasoning capabilities.

Additionally, we seek to better understand the functionality of identified belief directions through continuous interventions along these vectors. As demonstrated in~\cref{fig:alphas}, The (+ \tpfo) direction exhibits significant impacts on benchmark performance, underscoring its pivotal role in ToM reasoning process. Specifically, steering towards this direction consistently enhances ToM accuracy in \fb cases, while slightly decreases the \tb accuracy (partially due to an increase of invalid responses). At a finer level, we explore changes in model behavior by examining the fluctuations in the probabilities of next-token predictions.~\cref{fig:logits} illustrates how steering head activations to the specific directions could influence next-token predictions and invert the selection of choices.

\subsection{Varying Social Reasoning Tasks}
~\label{sec:generalization}

Furthermore, we investigate how various social reasoning tasks differ in terms of their underlying representations and whether these representations could generalize across different tasks or not.

\begin{table}[t]
\centering
\caption{Cross-task intervention results on Mistral-7B. We perform activation intervention towards the joint belief directions identified in the \fob task and evaluate the other two tasks.}
\vspace{1em}
\label{tab:generalization}
\footnotesize
\begin{tabularx}{\linewidth}{@{}lYYYYYY@{}}
\toprule
Model & \multicolumn{3}{c}{Forward Action} & \multicolumn{3}{c}{Backward Belief} \\
\cmidrule(lr){2-4} \cmidrule(l){5-7}
& TB & FB & Both & TB & FB & Both \\
\midrule
Mistral-7B & 0.92 & 0.30 & 0.26 & \textbf{0.91} & 0.22 & 0.16 \\
+ random & \textbf{0.93} & 0.31 & 0.26 & 0.90 & 0.22 & 0.16 \\
+ transferred & 0.90 & \textbf{0.40} & \textbf{0.33} & 0.73 & \textbf{0.38} & \textbf{0.22} \\
\midrule
DeepSeek-7B & \textbf{0.77} & 0.48 & 0.31 & 0.64 & 0.50 & 0.29 \\
+ random & 0.76 & 0.42 & 0.29 & \textbf{0.65} & 0.47 & \textbf{0.32} \\
+ transferred & 0.75 & \textbf{0.57} & \textbf{0.37} & 0.57 & \textbf{0.57} & 0.28 \\
\bottomrule
\end{tabularx}
\end{table}

First of all, we explore the interrelationships among the identified ToM representations across different tasks. The probing results of each task are presented in~\cref{sec:probe}.~\cref{fig:tasks} illustrates a strong correlation between the probing accuracies in different task scenarios, suggesting related representational capabilities under various causal inference conditions. Remarkably, the top-performing heads in one task tend to include the predictive features in another task as well, and the directions they identify exhibit high similarity. Considering that all three tasks implicitly or explicitly involve belief inference, we propose that a subset of attention head spaces might contain belief representations which potentially contribute to a range of social reasoning tasks.

This hypothesis motivates us to conduct a generalization test across different social reasoning tasks. We specifically intervene in the (+ \tpfo) directions identified under \fob conditions when evaluating the other two tasks.~\cref{tab:generalization} indicates that the directions identified in one task do generalize to others, suggesting that these directions might encapsulate a more universal function as belief representations.

Moreover, we seek to understand why these directions could act as generalizable belief representations. 
Specifically, we first prompt the model with story narratives, then respectively project the attention head activations onto the target (+ \tpfo) directions of the top probes. 
We then back-propagate the projection norm through the model and calculate the gradient magnitudes in input token embeddings, which approximately reflects the relevance of individual input tokens to the target directions. 
\Cref{fig:tokens} reveals that the identified directions in attention head activation spaces primarily focus on tokens denoting key causal variables, including the protagonist's desires and initial actions, the causal event that changes the environmental states, and the protagonist’s percept status of the causal event. These elements collectively facilitate a comprehensive inference of both agents' beliefs. These observations may shed light on the generalization potential of these directions across various social reasoning tasks. Despite the diverse causal inference patterns required by these tasks, they share common underlying causal variables and all necessitate inference regarding the agent's belief status, whether explicitly or implicitly. We present additional studies of these directions on other ToM reasoning scenarios and unrelated language tasks in ~\cref{sec:tomi,sec:other_tasks}.

%% file: icml2024/chaps/5_conclusion.tex
\section{Discussions}

In this study, we investigate the ToM capabilities in LLMs, specifically examining their ability to internally represent and attribute beliefs. We discover that LLMs can distinguish between different belief states of multiple agent perspectives through their intermediate activations with simple linear models. Additionally, we show that manipulation of these representations significantly affects the model's social reasoning performances. Finally, we demonstrate the generalization of the internal belief representations in diverse social reasoning task scenarios. 

Our study contributes to the ongoing dialogue on the social reasoning capabilities of LLMs, providing new insights into their ability to simulate ToM through internal representations.
Looking ahead, our study opens avenues for further investigation, including the development and processing of belief representations during training, their scalability in more complex LLMs like mixture of expert (MoE) models~\cite{jiang2024mixtral}, and methods to enhance machine ToM capabilities in alignment with human values. 
While our research provides valuable insights, it comes with its limitations. The scope of our exploration was confined to certain types of LLMs and specific social reasoning tasks, which may not capture the full spectrum of ToM capabilities. Future work should aim to address these gaps, broadening the understanding of ToM in AI systems across various models and more complex contexts.

\section*{Acknowledgements}
This work is supported by National Science and Technology Major Project (2022ZD0114904). We thank the anonymous reviewers for insightful discussions. We thank the awesome open-source toolbox nnsight~\cite{nnsight} which is used to extract the Transformer internal representations.

\section*{Impact Statement} Our study on the Theory of Mind (ToM) capabilities in Large Language Models (LLMs) illuminates the potential for more empathetic AI, enhancing human-machine interactions in various sectors. Ethically, it necessitates careful consideration to prevent misuse and bias propagation, ensuring AI's societal impact is positive. Responsible development and transparent deployment are imperative to safeguard against unintended consequences and maintain trust in AI advancements. Additionally, this study should not be misinterpreted by the media and the general public as evidence that LLMs exhibit consciousness and self-awareness.

%% file: icml2024/chaps/6_appendix.tex
\section{Statistical Analysis on Belief Probing}
\label{sec:stats}
To further verify the existence of attention heads that encode belief status of different agents, we conduct robust statistical analysis.
Specifically, we train and evaluate the linear probes across different data splits using 100 random seeds. We report the mean accuracies and 95\% confidence intervals (CI) for the top attention heads in ~\cref{tab:stat1,tab:stat2,tab:stat3}. We calculate the validation accuracy of linear probes for both True Belief (TB) and False Belief (FB) scenarios respectively. The probes are trained with full training set with both TB and FB. 
Furthermore, we apply a multiple hypothesis testing correction for our analysis of the top-10 attention heads. We adopt the stringent Bonferroni correction method that rigorously control the Family-Wise Error Rate (FWER), thereby significantly reducing the likelihood of type I errors. Under this framework, each attention head is scrutinized against a null hypothesis positing that its accuracy does not surpass a specific baseline (75\%), versus an alternative hypothesis that asserts superior accuracy.
The results robustly validates that a specific subset of attention heads exhibits the capacity to predict agents' belief states with a validation accuracy exceeding 75\%. Significantly low p-values with the Bonferroni correction strongly reject the null hypothesis, affirming the statistical significance and reliability of our findings.

\begin{table}[htbp]
\centering
\caption{The top predictive attention heads in belief status estimation for \textit{oracle} using logistic regression (binary). We use Mistral-7B model in the \fob task. Random guessing is 50\%.}
\footnotesize
\resizebox{0.85\linewidth}{!}{
\begin{tabular}{cccccccc}
\toprule
Position & Mean Acc. & CI & Corrected P-Value & Mean Acc (TB) & CI (TB) & Mean Acc (FB) & CI (FB) \\
\midrule
(14, 31) & 97.8 & (97.5, 98.1) & 0.0000 & 97.7 & (97.3, 98.2) & 97.8 & (97.4, 98.2) \\
(13, 26) & 97.5 & (97.1, 97.8) & 0.0000 & 97.1 & (96.6, 97.5) & 97.8 & (97.4, 98.2) \\
(14, 11) & 97.5 & (97.1, 97.8) & 0.0000 & 97.5 & (97.0, 98.0) & 97.4 & (97.0, 97.8) \\
(14, 18) & 97.4 & (97.0, 97.7) & 0.0000 & 97.6 & (97.1, 98.0) & 97.1 & (96.7, 97.6) \\
(14, 8) & 97.3 & (96.9, 97.6) & 0.0000 & 97.1 & (96.6, 97.6) & 97.4 & (97.0, 97.8) \\
(14, 26) & 97.2 & (96.9, 97.6) & 0.0000 & 96.8 & (96.3, 97.3) & 97.6 & (97.2, 98.1) \\
(15, 23) & 97.2 & (96.9, 97.6) & 0.0000 & 97.3 & (96.8, 97.7) & 97.2 & (96.7, 97.7) \\
(13, 2) & 97.2 & (96.8, 97.6) & 0.0000 & 97.2 & (96.7, 97.7) & 97.1 & (96.6, 97.5) \\
(13, 1) & 97.2 & (96.8, 97.6) & 0.0000 & 97.1 & (96.5, 97.6) & 97.2 & (96.7, 97.7) \\
(14, 0) & 97.1 & (96.8, 97.5) & 0.0000 & 97.3 & (96.8, 97.8) & 96.9 & (96.5, 97.4) \\
\bottomrule
\end{tabular}
}
\label{tab:stat1}
\end{table}

\begin{table}[htbp]
\centering
\caption{The top predictive attention heads in belief status estimation for \textit{protagonist} using logistic regression (binary). We use Mistral-7B model in the \fob task. Random guessing is 50\%.}
\footnotesize
\resizebox{0.85\linewidth}{!}{
\begin{tabular}{cccccccc}
\toprule
Position & Mean Acc. & CI & Corrected P-Value & Mean Acc (TB) & CI (TB) & Mean Acc (FB) & CI (FB) \\
\midrule
(10, 16) & 78.3 & (77.4, 79.2) & 0.0000 & 80.7 & (79.5, 82.0) & 76.2 & (74.9, 77.5) \\
(10, 2) & 77.4 & (76.6, 78.3) & 0.0000 & 76.5 & (75.2, 77.8) & 78.7 & (77.3, 80.0) \\
(10, 21) & 77.3 & (76.4, 78.1) & 0.0000 & 77.4 & (76.1, 78.7) & 77.5 & (76.2, 78.7) \\
(10, 3) & 77.2 & (76.4, 78.1) & 0.0000 & 75.2 & (74.0, 76.5) & 79.5 & (78.2, 80.9) \\
(10, 4) & 76.9 & (76.0, 77.8) & 0.0004 & 79.9 & (78.6, 81.3) & 74.4 & (73.0, 75.8) \\
(10, 20) & 76.9 & (76.0, 77.7) & 0.0002 & 78.7 & (77.4, 80.1) & 75.4 & (74.1, 76.7) \\
(10, 11) & 76.8 & (75.9, 77.7) & 0.0013 & 75.1 & (73.7, 76.6) & 78.9 & (77.7, 80.1) \\
(10, 17) & 76.5 & (75.7, 77.3) & 0.0028 & 76.3 & (75.1, 77.5) & 76.9 & (75.8, 78.1) \\
(10, 15) & 76.5 & (75.6, 77.4) & 0.0106 & 77.1 & (75.8, 78.4) & 76.2 & (74.8, 77.5) \\
(10, 10) & 76.5 & (75.6, 77.3) & 0.0088 & 82.9 & (81.6, 84.2) & 70.5 & (69.1, 71.8) \\
\bottomrule
\end{tabular}
}
\label{tab:stat2}
\end{table}

\begin{table}[htbp]
\centering
\caption{The top predictive attention heads in belief status estimation for both agents using multinomial logistic regression (quaternary). We use Mistral-7B model in the \fob task. Random guessing is 25\%.}
\footnotesize
\resizebox{0.85\linewidth}{!}{
\begin{tabular}{cccccccc}
\toprule
Position & Mean Acc. & CI & Corrected P-Value & Mean Acc (TB) & CI (TB) & Mean Acc (FB) & CI (FB) \\
\midrule
(12, 27) & 79.0 & (77.6, 80.4) & 0.0000 & 84.0 & (82.1, 85.8) & 75.3 & (72.7, 78.0) \\
(12, 31) & 78.8 & (77.1, 80.4) & 0.0002 & 85.7 & (83.8, 87.7) & 73.3 & (70.3, 76.3) \\
(12, 1) & 78.6 & (77.2, 79.9) & 0.0000 & 84.9 & (83.1, 86.8) & 73.5 & (71.0, 75.9) \\
(12, 13) & 78.5 & (77.0, 80.1) & 0.0002 & 85.0 & (83.0, 87.1) & 73.3 & (70.5, 76.1) \\
(15, 8) & 78.4 & (77.2, 79.6) & 0.0000 & 87.5 & (85.8, 89.2) & 70.8 & (68.4, 73.1) \\
(16, 20) & 78.0 & (76.9, 79.2) & 0.0000 & 83.4 & (81.7, 85.1) & 74.0 & (71.8, 76.3) \\
(12, 7) & 78.0 & (76.6, 79.5) & 0.0005 & 85.9 & (84.1, 87.6) & 71.4 & (68.8, 74.1) \\
(12, 24) & 78.0 & (76.3, 79.7) & 0.0086 & 86.0 & (83.8, 88.2) & 71.6 & (68.5, 74.7) \\
(16, 2) & 77.9 & (76.8, 79.0) & 0.0000 & 83.3 & (81.7, 85.0) & 73.6 & (71.6, 75.7) \\
(14, 1) & 77.9 & (76.3, 79.5) & 0.0046 & 79.2 & (76.4, 81.9) & 78.5 & (75.8, 81.2) \\
\bottomrule
\end{tabular}
}
\label{tab:stat3}
\end{table}

\section{Probing Prompts for Different Tasks}
\label{sec:prompts}

\begin{figure*}[h]
  \centering
  \includegraphics[width=0.82\linewidth]{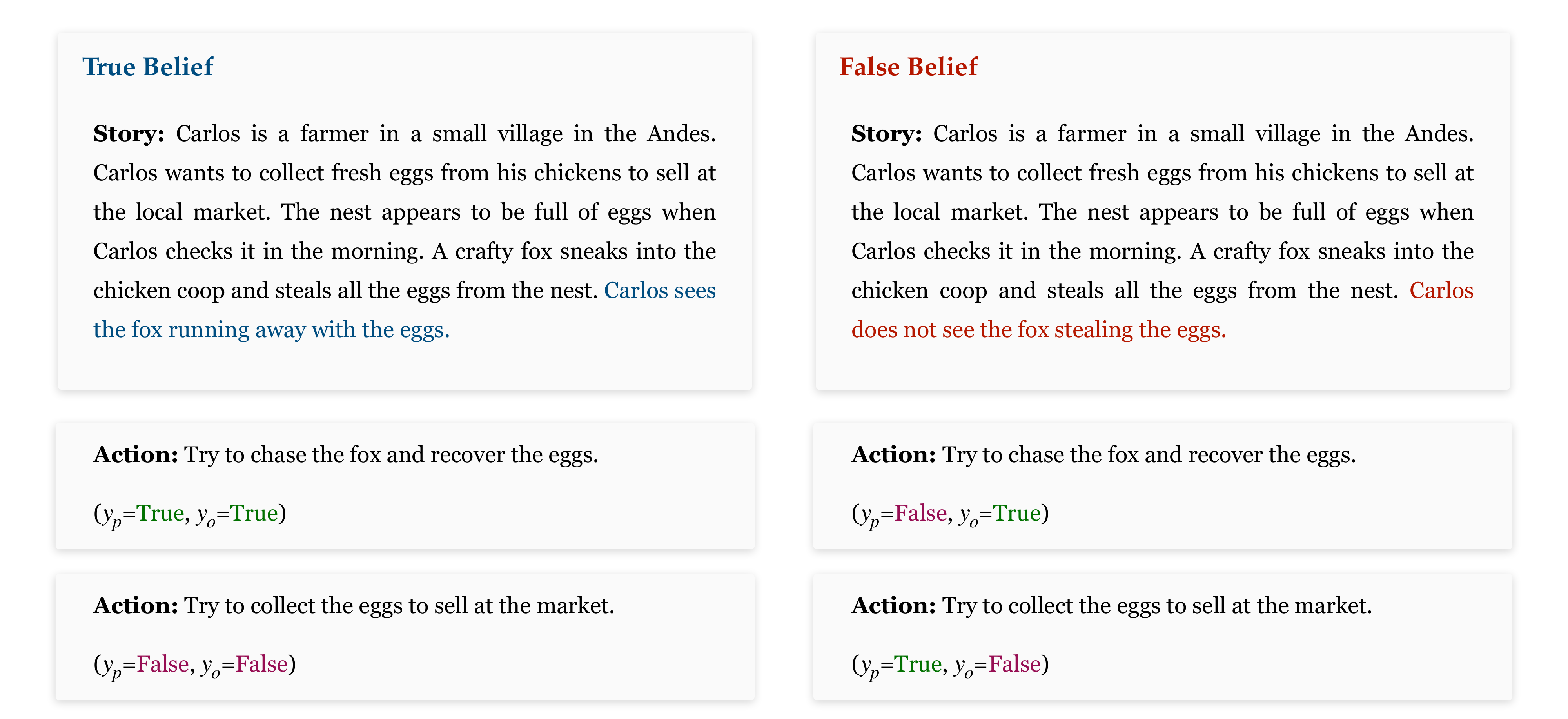}
  \caption{
  Example case of representation probing for the \foa task. Left: a ``True Belief'' story where the protagonist shares the same belief with oracle. Right: a ``False belief'' story where the protagonist has different belief with oracle. For both stories, we concatenate them with the two actions respectively and directly feed to the model. The ground-truth action label from the protagonist's perspective $y_p$ and the oracle's perspective $y_o$ can then be decided accordingly.
  }
  \label{fig:prompt_fa}
\end{figure*}

\begin{figure*}[h]
  \centering
  \includegraphics[width=0.82\linewidth]{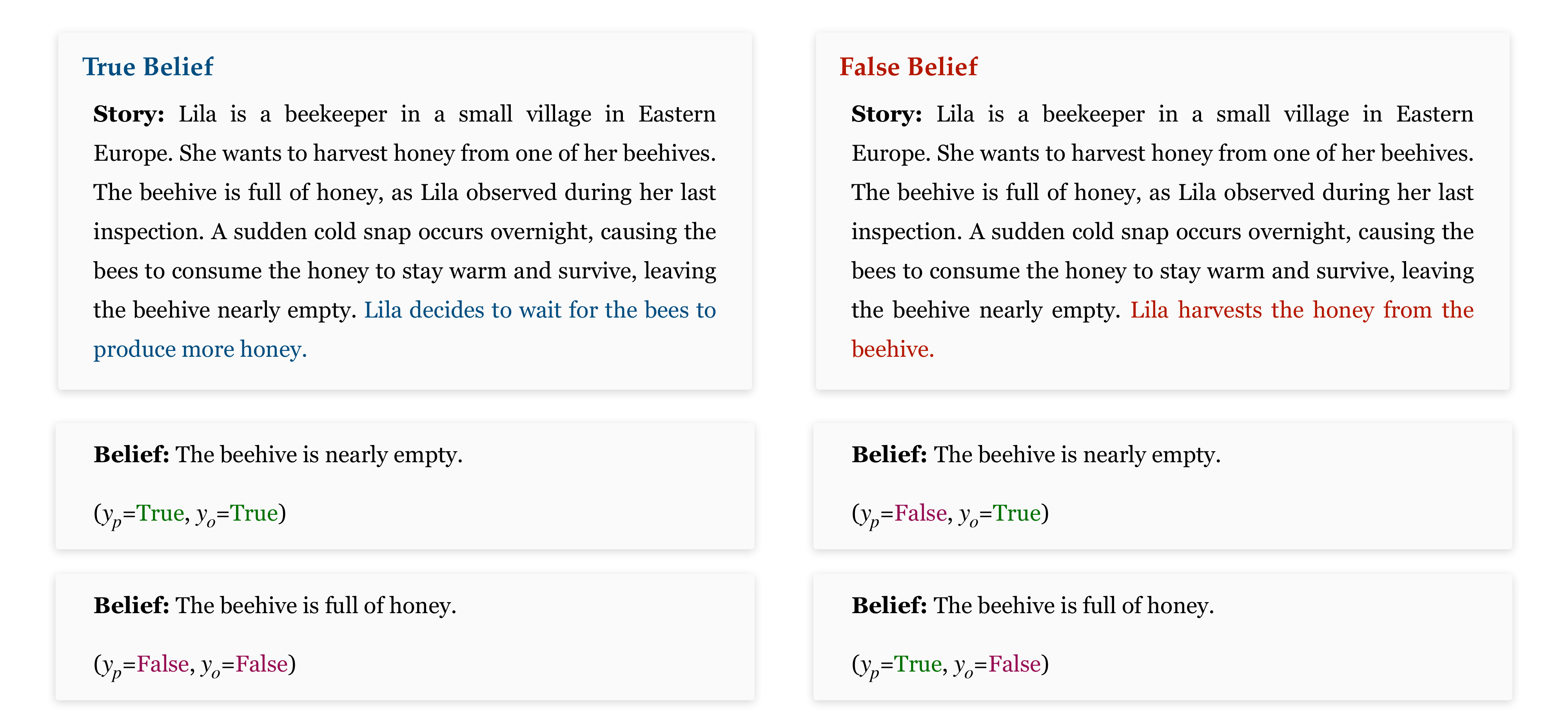}
  \caption{
  Example case of representation probing for the \bab task. Left: a ``True Belief'' story where the protagonist shares the same belief with oracle. Right: a ``False belief'' story where the protagonist has different belief with oracle. For both stories, we concatenate them with the two beliefs respectively and directly feed to the model. The ground-truth belief label from the protagonist's perspective $y_p$ and the oracle's perspective $y_o$ can then be decided accordingly.
  }
  \label{fig:prompt_bb}
\end{figure*}

We develop corresponding prompt templates for different social reasoning tasks by casting the choices to statements. For the \foa task, we utilize the story-action pairs as shown in ~\cref{fig:prompt_fa}. For the \bab task, we include the \p's next-step actions in the story and present the story-belief pairs as shown in ~\cref{fig:prompt_bb}. 

\section{Additional Probing Results}
\label{sec:probe}

\begin{figure*}[h]
  \centering
  \includegraphics[width=\linewidth]{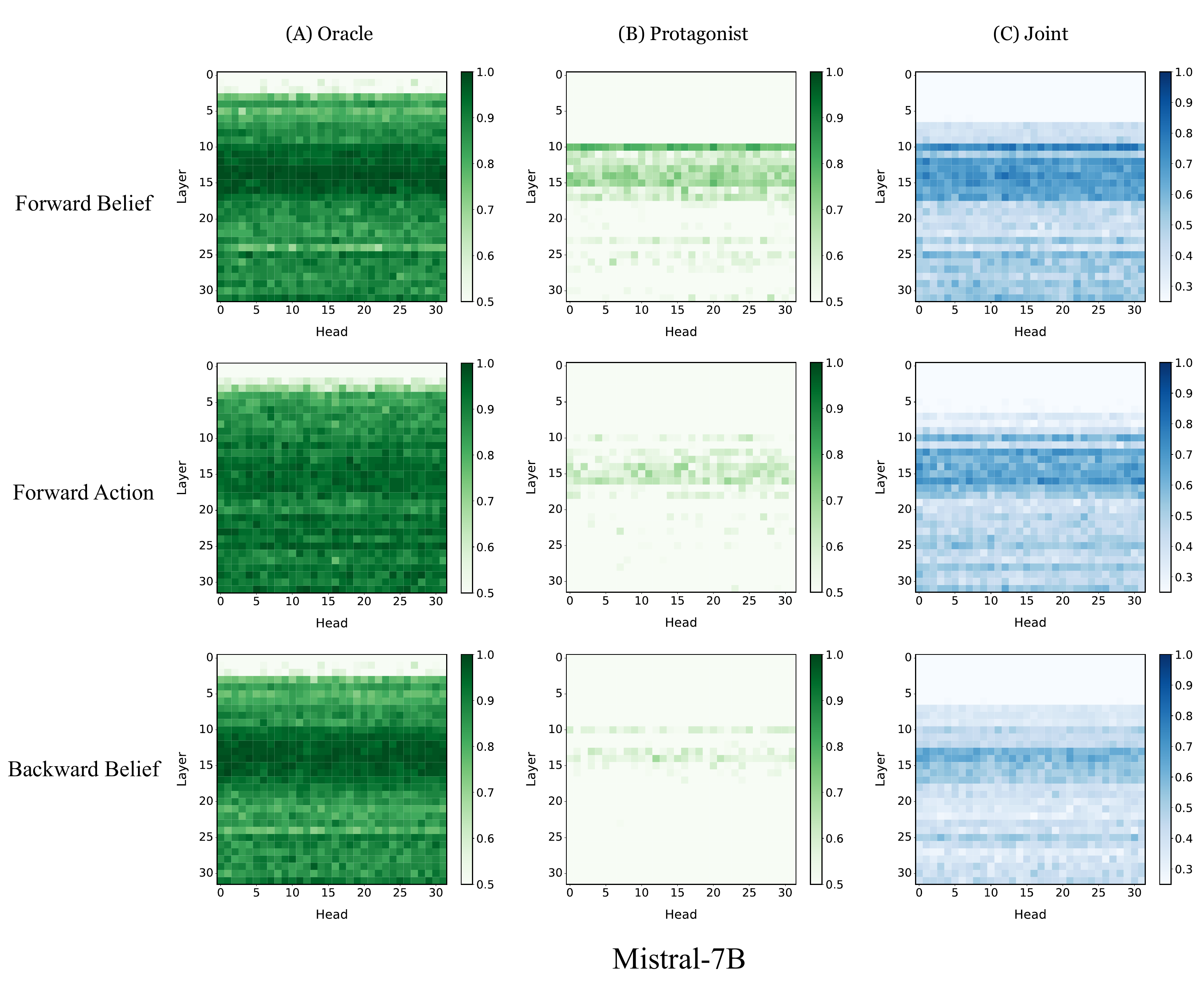}
  \caption{
  Probe accuracies on different tasks based on the attention head activations in all layers of Mistral-7B. (A) Belief status estimation for \emph{oracle} using logistic regression (binary). (B) Belief status estimation for \emph{protagonist} using logistic regression (binary). (C) Joint belief status estimation for both agents using multinomial logistic regression (quaternary).
  }
  \label{fig:probe-mis}
\end{figure*}

\begin{figure*}[h]
  \centering
  \includegraphics[width=\linewidth]{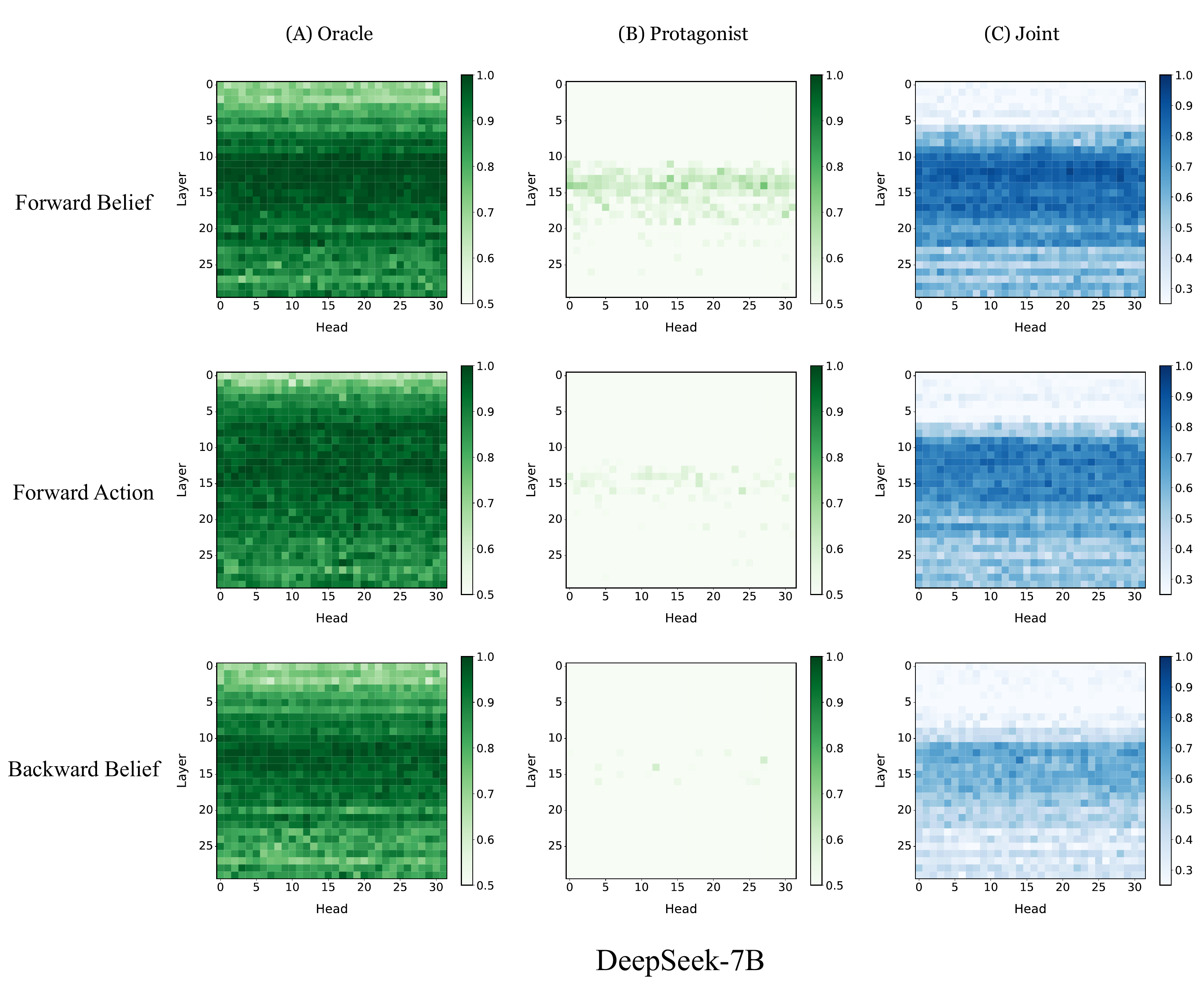}
  \caption{
  Probe accuracies on different tasks based on the attention head activations in all layers of DeepSeek-7B. (A) Belief status estimation for \emph{oracle} using logistic regression (binary). (B) Belief status estimation for \emph{protagonist} using logistic regression (binary). (C) Joint belief status estimation for both agents using multinomial logistic regression (quaternary).
  }
  \label{fig:probe-ds}
\end{figure*}

We present the full probing results in all three tasks for both models using logistic regression models in ~\cref{fig:probe-mis} and ~\cref{fig:probe-ds}. The probing accuracies vary across models and tasks. Generally, linear belief representations exist in different models and tasks, and are biased towards representing \o's belief.

\begin{figure*}[h]
  \centering
  \includegraphics[width=\linewidth]{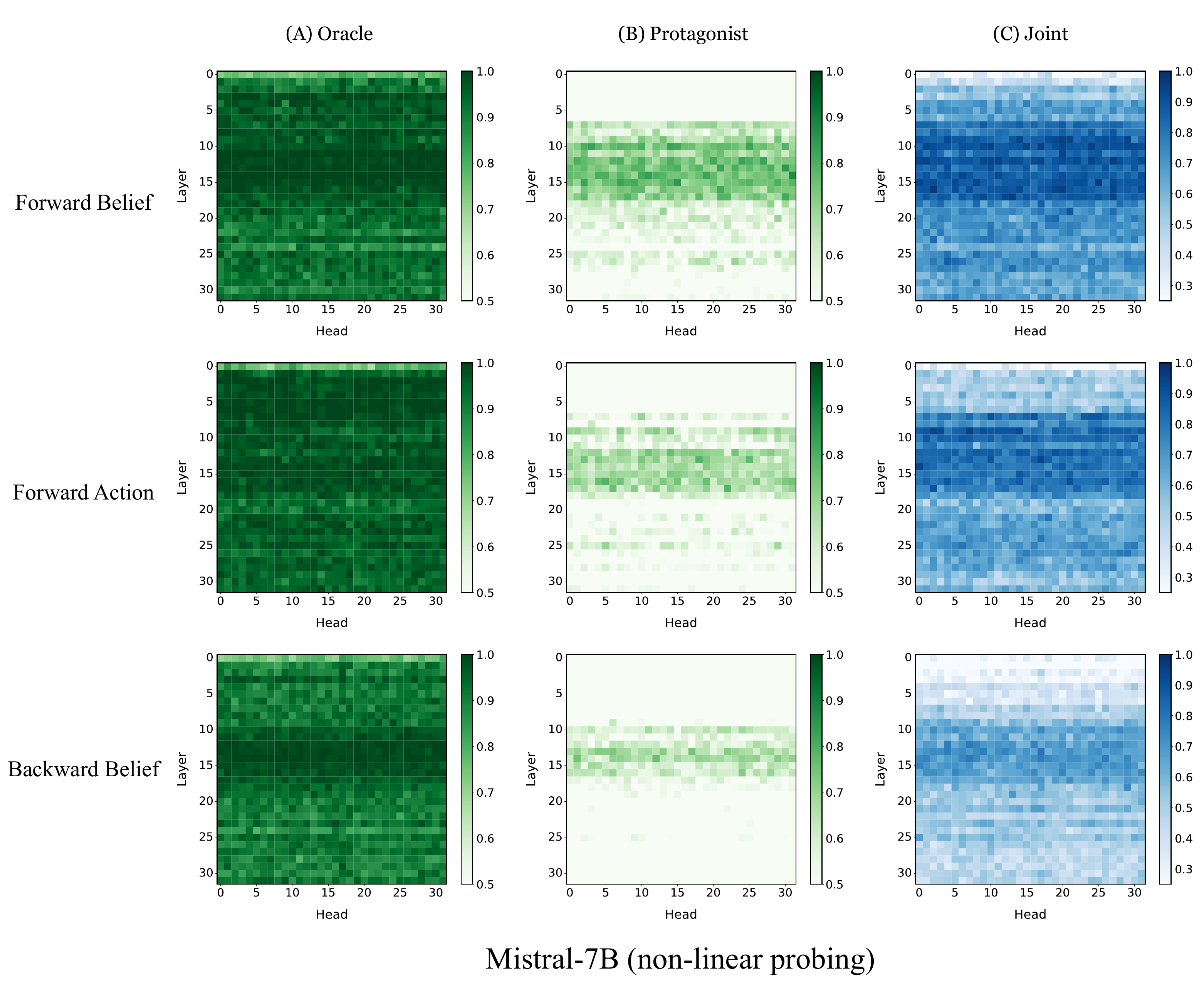}
  \caption{
  Non-linear probe accuracies on different tasks based on the attention head activations in all layers of Mistral-7B. (A) Belief status estimation for \emph{oracle} using binary classification. (B) Belief status estimation for \emph{protagonist} using binary classification. (C) Joint belief status estimation for both agents using quaternary classification.
  }
  \label{fig:probe-mis-nl}
\end{figure*}

Furthermore, we explore non-linear probing by fitting an MLP with one hidden layer of $256$ channels. ~\Cref{fig:probe-mis-nl} shows that the overall probing accuracy increases, suggesting that while some activation heads can decode beliefs linearly, more complex representational structures within certain activation spaces also exist.

\clearpage

\section{Generalization to Other Datasets}
\label{sec:tomi}
In addition to the stories in BigToM~\cite{gandhi2023bigtom}, we explore whether our findings could generalize to other narratives. Following~\cite{wilf2023think}, we extend our study to the ToMi benchmark~\cite{le-etal-2019-revisiting}, which has quite different narrative templates and scenarios compared to BigToM. It also contains second-order ToM questions which are not present in BigToM. 

\begin{figure*}[h]
  \centering
  \includegraphics[width=0.8\linewidth]{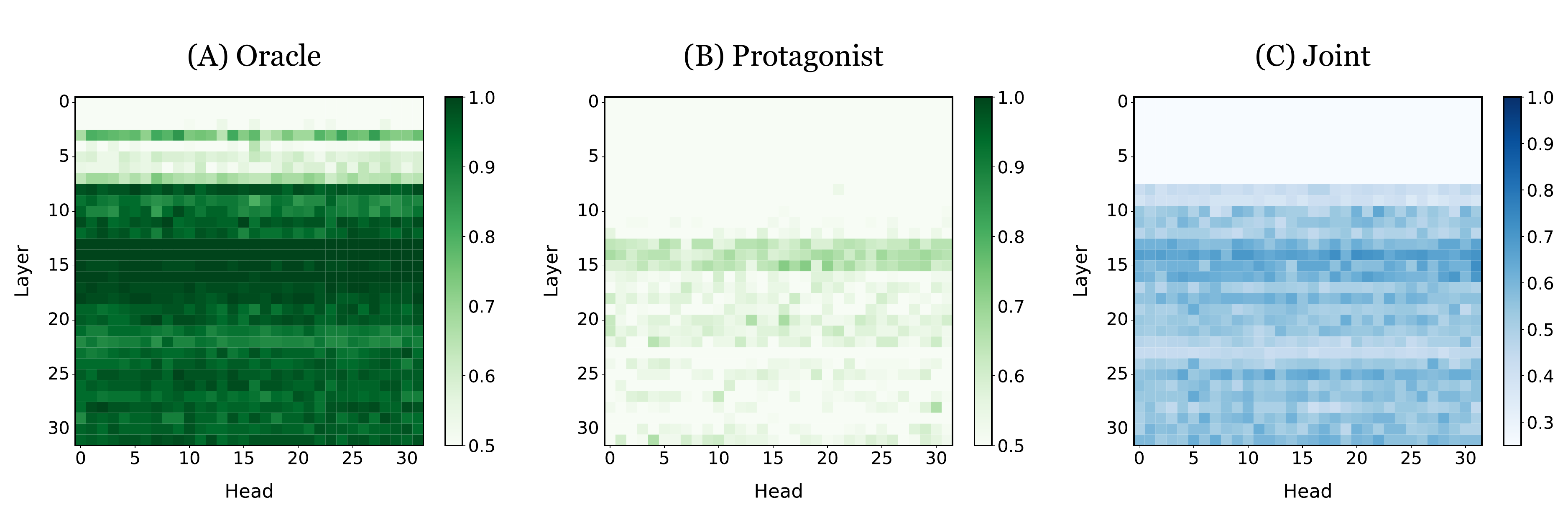}
  \caption{Probe accuracies on ToMi dataset based on the attention head activations in all layers of DeepSeek-7B. (A) Belief status estimation for \emph{oracle} using logistic regression (binary). (B) Belief status estimation for \emph{protagonist} using logistic regression (binary). (C) Joint belief status estimation for both agents using multinomial logistic regression (quaternary).}
  \label{fig:probing_tomi}
\end{figure*}

We perform probing on ToMi by constructing the prompt templates based on its question format, as shown below. Since the ToMi questions are predicting the protagonists' next-step move, it is closer to the \foa task in BigToM.

\definecolor{verylightgray}{rgb}{0.97, 0.97, 0.97}
\begin{center}
\begin{minipage}{0.8\linewidth}
\begin{lstlisting}[caption=An example of the prompt for probing with the ToMi benchmark, label=lst:example, frame=single, basicstyle=\ttfamily\footnotesize, backgroundcolor=\color{verylightgray}, breaklines=true]
Story: 1 Evelyn entered the cellar. 2 Owen entered the cellar. 3 The belt is in the bathtub. 4 The bathtub is in the cellar. 5 Elizabeth loves the apple. 6 Evelyn exited the cellar. 7 Owen moved the belt to the envelope. 8 The envelope is in the cellar.
Action: Look for the belt in the bathtub.
\end{lstlisting}
\end{minipage}
\end{center}

\begin{figure*}[h]
  \centering
  \includegraphics[width=0.4\linewidth]{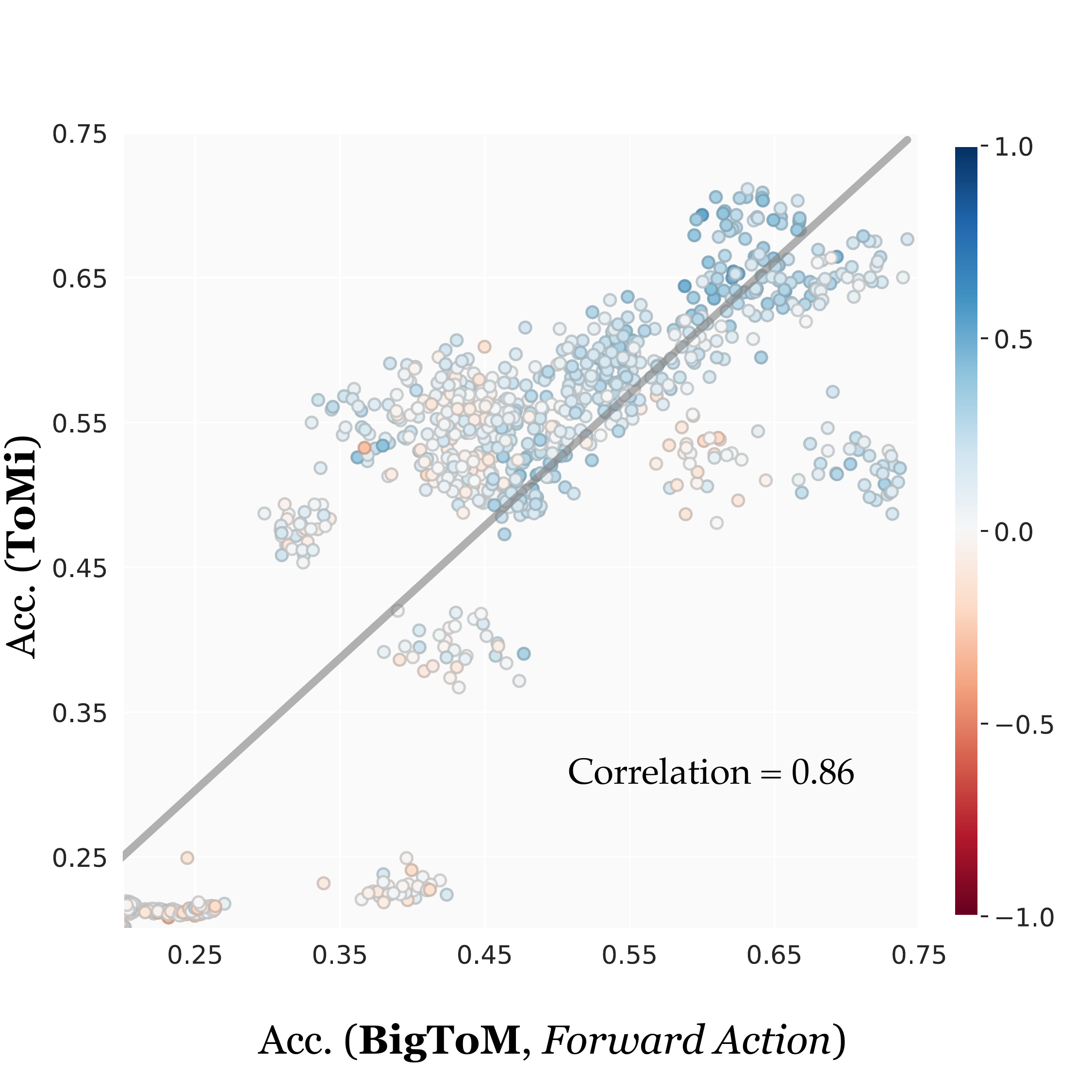}
  \caption{Pairwise comparisons of multinomial probing results on Mistral-7B between ToMi and BigToM. Each point represents a specific attention head, with its position denoting the probe accuracies in the two datasets (multinomial), and its color denoting the cosine similarity between the ($+\text{T}_p\text{F}_o$) probe directions of the two datasets..}
  \label{fig:correlation_tomi}
\end{figure*}

The probing results are shown in ~\cref{fig:probing_tomi}. Despite the large narrative differences, the probing results are similar - most heads can accurately predict the oracle state; a few attention heads in the middle layers could also separate the protagonist's belief. 
In addition, we explore if the top heads identified in BigToM and ToMi are similar or not. We present a correlation plot in~\cref{fig:correlation_tomi}. Here,  The accuracies and similarities are averaged over 100 runs of different random seeds. We note that there is a significant correlation (0.86) in probe accuracies across the benchmarks. In other words, the top predictive attention heads in one benchmark exhibit similar efficacy in the other as well. Additionally, The identified top directions in the two datasets exhibit moderate similarity (blue-ish), although their similarities are smaller than cross-task comparison within BigToM.

\begin{table}[ht]
\centering
\caption{Model performance comparison on the ToMi benchmark.}
\footnotesize
\label{tab:tomi}
\begin{tabular}{lcc}
\toprule
Model & ToM (FB) & All \\
\midrule
Mistral-7B (baseline) & 63.8 & 66.0 \\
Mistral-7B (+random)  & 63.3 & 64.3 \\
Mistral-7B (+$\text{T}_p\text{F}_o$, transferred from BigToM)    & \textbf{71.0} & \textbf{67.9} \\
\bottomrule
\end{tabular}
\end{table}

Furthermore, we directly transfer the same attention heads and directions identified within the BigToM dataset (\fob task) for activation intervention on the ToMi benchmark. ~\cref{tab:tomi} shows that the directions we identified with BigToM templates are not limited to that specific format. Instead, they exhibit a promising degree of generalization to scenarios with varied narrative styles. This suggests that these representations may indeed represent broadly applicable ToM insights.

\section{Influence on Other Tasks}
\label{sec:other_tasks}
\begin{table}[htbp]
\centering
\caption{Model performance comparison on the MMLU benchmark.}
\footnotesize
\begin{tabular}{lccccc}
\toprule
 & All & Humanities & Social Sciences & STEM & Other \\
\midrule
Mistral-7B (baseline) & 57.5 & 52.7 & \textbf{66.4} & 48.3 & 65.5 \\
Mistral-7B (+random) & 57.6 & 52.7 & 66.3 & 48.4 & \textbf{65.6} \\
Mistral-7B (+$\text{T}_p\text{F}_o$) & \textbf{57.7} & \textbf{53.0} & 66.3 & \textbf{48.7} & 65.5 \\
\bottomrule
\end{tabular}
\label{tab:mmlu}
\end{table}

\begin{table}[htbp]
\centering
\caption{Model performance comparison on CoLA, MRPC, and QNLI benchmarks.}
\footnotesize
\begin{tabular}{lccc}
\toprule
Dataset & CoLA & MRPC & QNLI \\
\midrule
Mistral-7B (baseline) & 70.0 & 58.8 & 62.5 \\
Mistral-7B (+random) & 70.1 & \textbf{59.2} & \textbf{63.1} \\
Mistral-7B (+$\text{T}_p\text{F}_o$) & \textbf{71.1} & 58.9 & 63.0 \\
\bottomrule
\end{tabular}
\label{tab:cola_mrpc_qnli}
\end{table}

In order to understand the influence of intervention along the identified belief directions on unrelated tasks, we further evaluate the model on some general language understanding benchmarks, including Massive Multitask Language Understanding (MMLU)~\cite{hendrycks2020measuring}, The Corpus of Linguistic Acceptability (CoLA)~\cite{warstadt2019neural}, Microsoft Research Paraphrase Corpus (MRPC) ~\cite{dolan2005automatically}, Question-answering NLI (QNLI)~\cite{wang2018glue}. These benchmarks measure the language model's performance in tasks unrelated to ToM, including knowledge acquicision, grammar check, \etc. We discover that the activation intervention along the identified ToM directions does not significantly change the model performance, as shown in \cref{tab:mmlu,tab:cola_mrpc_qnli}.

\section{Additional Token Gradient Heatmaps}
\begin{figure*}[h]
  \centering
  \includegraphics[width=\linewidth]{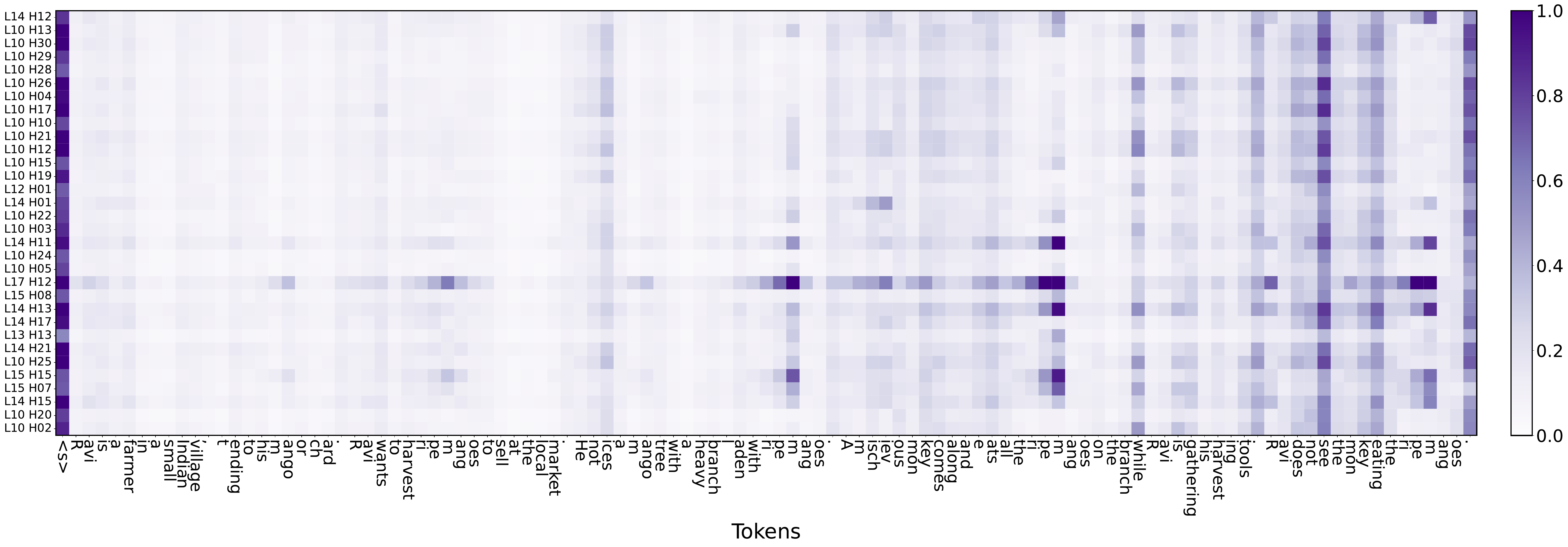}
  \label{fig:tokens_5}
\end{figure*}

\begin{figure*}[h]
  \centering
  \includegraphics[width=\linewidth]{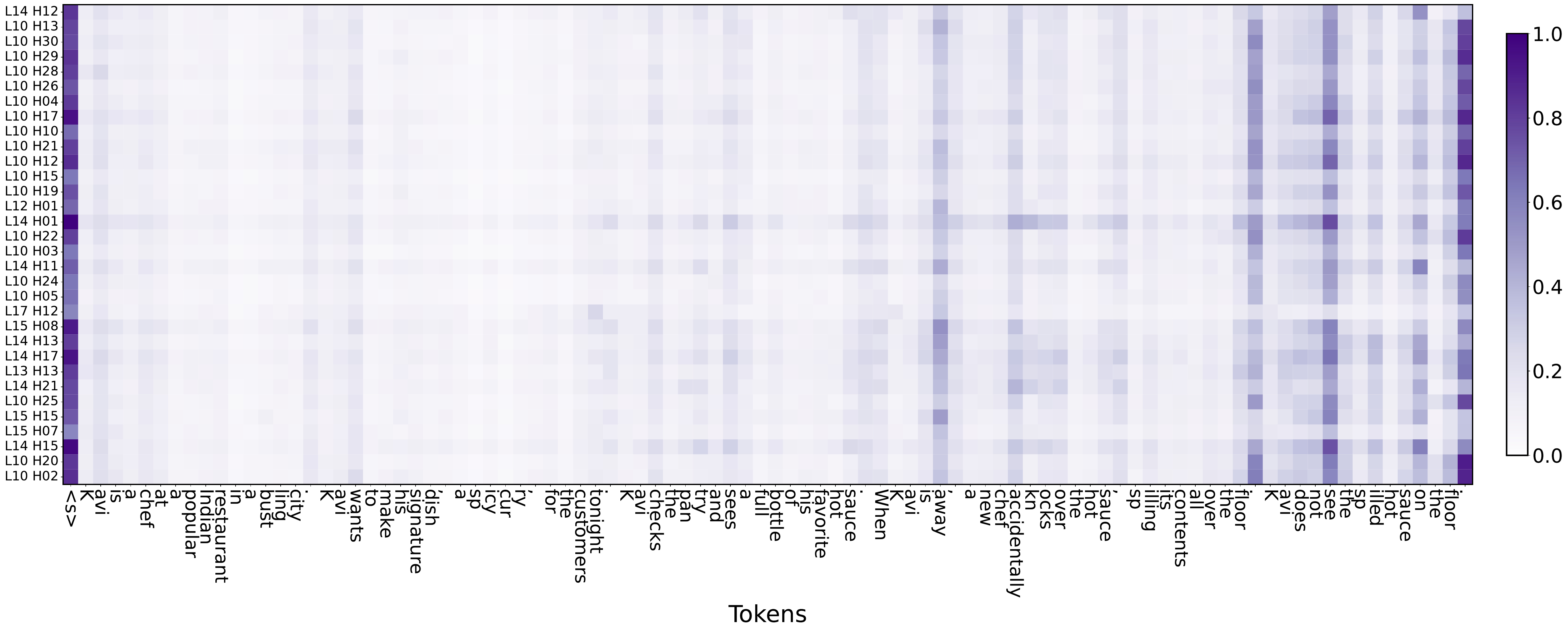}
  \caption{Magnitudes of gradients on the token embeddings with respect to the projection of attention head activations over the corresponding joint belief directions. Each line represents a specific attention head in Mistral-7B. 
  }
  \label{fig:tokens_12}
\end{figure*}

\begin{figure*}[h]
  \centering
  \includegraphics[width=\linewidth]{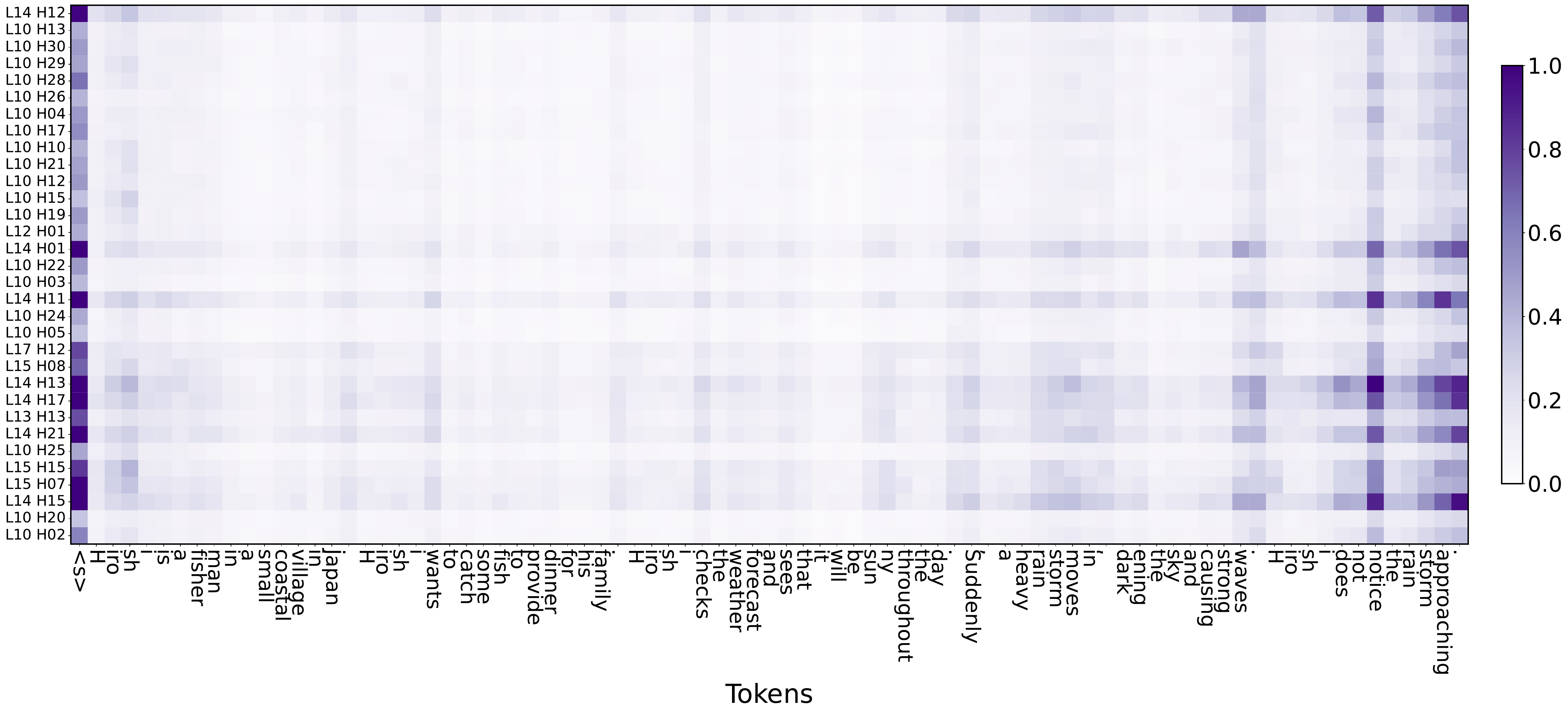}
  \label{fig:tokens_13}
\end{figure*}

\begin{figure*}[h]
  \centering
  \includegraphics[width=\linewidth]{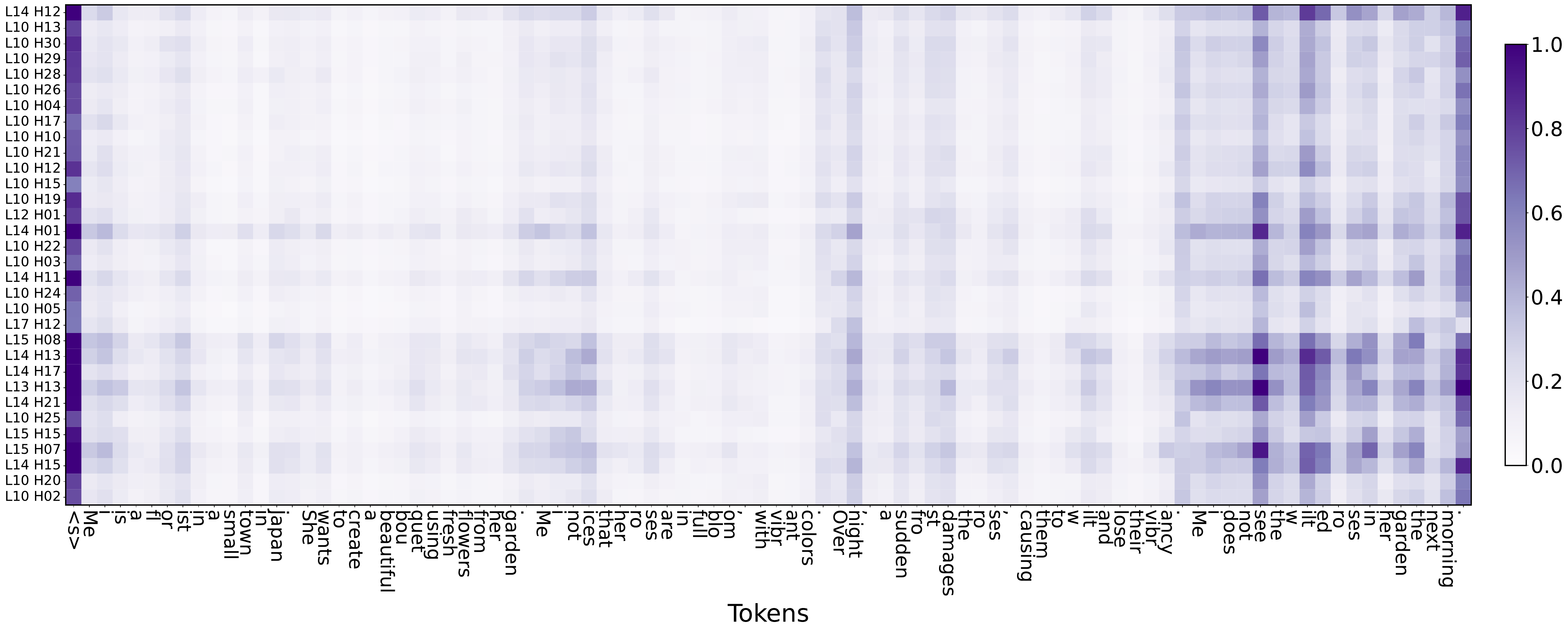}
  \caption{Magnitudes of gradients on the token embeddings with respect to the projection of attention head activations over the corresponding joint belief directions. Each line represents a specific attention head in Mistral-7B. 
  }
  \label{fig:tokens_15}
\end{figure*}

We provide additional results of token gradients with regard to the belief directions given different story prompts, which localizes the key causal elements related to agent beliefs.

\clearpage

\section{Hyperparameter Analysis}
~\label{sec:hyper}

\begin{figure*}[h]
  \centering
  \includegraphics[width=\linewidth]{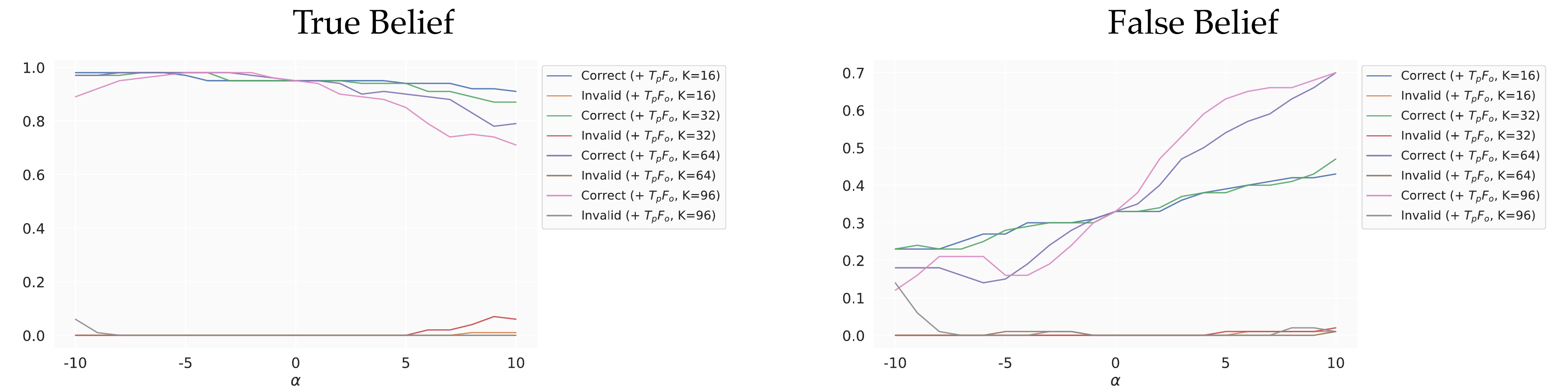}
\end{figure*}

\begin{figure*}[h]
  \centering
  \includegraphics[width=\linewidth]{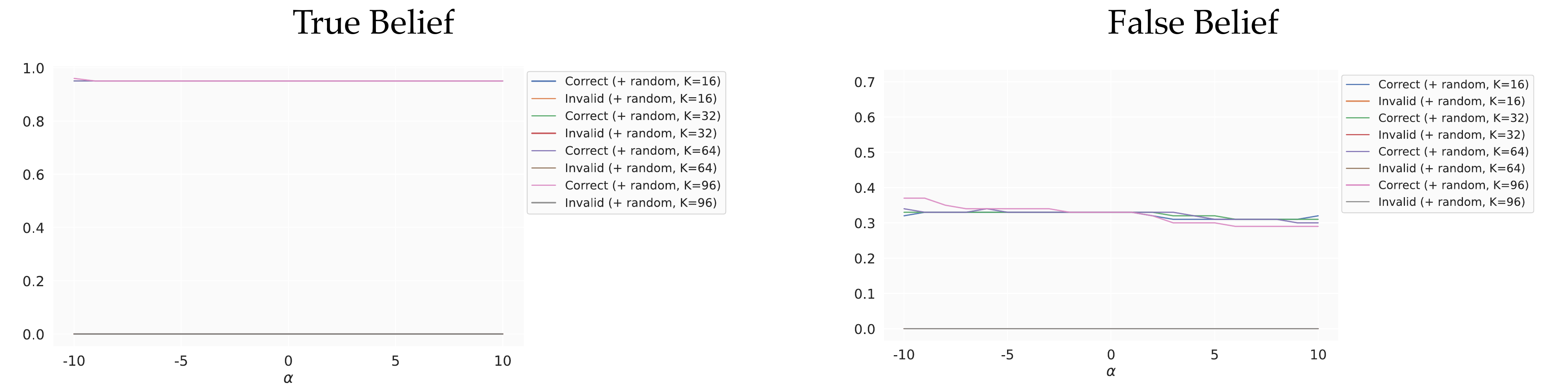}
  \caption{Impact of varying intervention strength $\alpha$ on the \fob task using Mistral-7B. ``Invalid'' denotes the answer is not recognized by the grading mechanism as the model fails to provide answer in the required format, \eg, delivering uncertain responses.}
  \label{fig:alphas_K}
\end{figure*}

We provide intervention results with different $(K, \alpha)$ combinations in ~\cref{{fig:alphas_K}}. $K=16$ yields the most steady performance and is used in our experiments for the model.

\clearpage

\input{icml2024/chaps/Changelog}

%% file: icml2024/chaps/Changelog.tex
\section{Detailed Comparison of LLM Responses}

Below, we compare the performance of two models (Mistral-7B, DeepSeek-7B) on the \fob reasoning task before and after intervention in two settings (\fb, \tb). Instances where the model's response changes from incorrect to correct, or from correct to incorrect, are recorded in the tables below.

\subsection{Mistral-7B on Forward Belief, False Belief}

\noindent Changed to Correct Answer: 33/100
\begin{tcolorbox}[fontupper=\fontsize{8}{9.6}\selectfont]
Story: Ravi is a farmer in a small Indian village, tending to his mango orchard. Ravi wants to harvest ripe mangoes to sell at the local market. He notices a mango tree with a heavy branch laden with ripe mangoes. A mischievous monkey comes along and eats all the ripe mangoes on the branch while Ravi is gathering his harvesting tools. Ravi does not see the monkey eating the ripe mangoes.

\textcolor[RGB]{145, 100, 160}{Question: Does Ravi believe the branch has ripe mangoes or no ripe mangoes?}

\textcolor[RGB]{255, 165, 79}{Choose one of the following:}

a) Ravi believes the branch has ripe mangoes.

b) Ravi believes the branch has no ripe mangoes.

\textcolor[RGB]{34, 139, 34}{Correct Answer: a)}

\textcolor{red}{Unintervened: b) Ravi believes the branch has no ripe mangoes.}

\textcolor{blue}{Intervened: a. Ravi believes the branch has ripe mangoes.}

\end{tcolorbox}

\begin{tcolorbox}[fontupper=\fontsize{8}{9.6}\selectfont]
Story: Aniket is a marine biologist studying coral reefs off the coast of India. Aniket needs to collect samples of coral to analyze the effects of climate change on the reef. Aniket spots a healthy-looking coral formation in a specific area of the reef. A sudden wave surge stirs up sediment, covering the once healthy coral formation and causing it to become damaged. Aniket does not notice the wave surge or the sediment covering the coral.

\textcolor[RGB]{145, 100, 160}{Question: Does Aniket believe the coral formation is healthy or damaged?}

\textcolor[RGB]{255, 165, 79}{Choose one of the following:}

a) Aniket believes the coral formation is damaged.

b) Aniket believes the coral formation is healthy.

\textcolor[RGB]{34, 139, 34}{Correct Answer: b)}

\textcolor{red}{Unintervened: a) Aniket believes the coral formation is damaged. (The context states that the cor}

\textcolor{blue}{Intervened: b) Aniket believes the coral formation is healthy. (at the time of reading the}

\end{tcolorbox}

\begin{tcolorbox}[fontupper=\fontsize{8}{9.6}\selectfont]
Story: Juan is a young chef in a small Mexican town. He wants to make a delicious salsa for the upcoming town festival. Juan buys fresh tomatoes from the market, which he believes are ripe and perfect for the salsa. While Juan is away from his kitchen, a mischievous cat knocks over the bag of tomatoes, crushing them and making them unsuitable for the salsa. Juan does not notice the crushed tomatoes on the floor.

\textcolor[RGB]{145, 100, 160}{Question: Does Juan believe the tomatoes are ripe and perfect or crushed and unsuitable?}

\textcolor[RGB]{255, 165, 79}{Choose one of the following:}

a) Juan believes the tomatoes are ripe and perfect.

b) Juan believes the tomatoes are crushed and unsuitable.

\textcolor[RGB]{34, 139, 34}{Correct Answer: a)}

\textcolor{red}{Unintervened: b) Juan believes the tomatoes are ripe and perfect, but they are actually crushed and un}

\textcolor{blue}{Intervened: a) Juan believes the tomatoes are ripe and perfect.}

\end{tcolorbox}

\begin{tcolorbox}[fontupper=\fontsize{8}{9.6}\selectfont]
Story: Cheng is a teacher in a rural Chinese village. Cheng wants to use a projector to show a documentary about Chinese history to his students. The projector is plugged in and turned on, as Cheng has prepared it beforehand. As the wind blows outside, the power goes out, causing the projector to turn off. Cheng does not notice the power outage.

\textcolor[RGB]{145, 100, 160}{Question: Does Cheng believe the projector is ready to use or turned off?}

\textcolor[RGB]{255, 165, 79}{Choose one of the following:}

a) Cheng believes the projector is turned off.

b) Cheng believes the projector is ready to use.

\textcolor[RGB]{34, 139, 34}{Correct Answer: b)}

\textcolor{red}{Unintervened: a) Cheng believes the projector is ready to use. (However, it is actually turned}

\textcolor{blue}{Intervened: Cheng believes the projector is ready to use. (before the power outage)}

\end{tcolorbox}

\begin{tcolorbox}[fontupper=\fontsize{8}{9.6}\selectfont]
Story: Isabela is a baker in a small Argentinian town known for its delicious pastries. Isabela wants to make a batch of alfajores, a traditional Argentinian dessert, for her bakery's customers. Isabela gets a bag of powdered sugar from her pantry to dust the alfajores. Her assistant, who was organizing the pantry earlier, accidentally placed a bag of flour in the powdered sugar's spot, and Isabela unknowingly grabs the bag of flour instead. Isabela does not notice the flour.

\textcolor[RGB]{145, 100, 160}{Question: Does Isabela believe the bag she took from the pantry contains powdered sugar or flour?}

\textcolor[RGB]{255, 165, 79}{Choose one of the following:}

a) Isabela believes the bag she took from the pantry contains powdered sugar.

b) Isabela believes the bag she took from the pantry contains flour.

\textcolor[RGB]{34, 139, 34}{Correct Answer: a)}

\textcolor{red}{Unintervened: b) Isabela believes the bag she took from the pantry contains flour.}

\textcolor{blue}{Intervened: a) Isabela believes the bag she took from the pantry contains powdered sugar.}

\end{tcolorbox}

\begin{tcolorbox}[fontupper=\fontsize{8}{9.6}\selectfont]
Story: Mei is a pearl diver in a small coastal village in Japan. Mei wants to find a perfect pearl to give to her grandmother for her birthday. Mei spots an oyster at the bottom of the sea that looks to be the right size and age to contain a pearl. A curious octopus opens the oyster, revealing that there is no pearl inside, and then swims away. Mei does not notice the octopus opening the oyster.

\textcolor[RGB]{145, 100, 160}{Question: Does Mei believe the oyster she spotted contains a pearl or that it is empty?}

\textcolor[RGB]{255, 165, 79}{Choose one of the following:}

a) Mei believes the oyster she spotted is empty.

b) Mei believes the oyster she spotted contains a pearl.

\textcolor[RGB]{34, 139, 34}{Correct Answer: b)}

\textcolor{red}{Unintervened: a) Mei believes the oyster she spotted[PAD]it;s empty.}

\textcolor{blue}{Intervened: b) Mei believes the oyster she spotted contains a pearl.}

\end{tcolorbox}

\begin{tcolorbox}[fontupper=\fontsize{8}{9.6}\selectfont]
Story: Heiwa is a skilled fisherman in a coastal village in Japan. Heiwa wants to catch a large number of fish to sell at the local market. The sea appears calm and full of fish, making it an ideal time for fishing. Suddenly, a massive school of fish moves away from the area, leaving very few fish near Heiwa's fishing spot. Heiwa does not notice the fish moving away.

\textcolor[RGB]{145, 100, 160}{Question: Does Heiwa believe the sea is calm and full of fish or that the fish have moved away?}

\textcolor[RGB]{255, 165, 79}{Choose one of the following:}

a) Heiwa believes the sea is calm and full of fish.

b) Heiwa believes the fish have moved away.

\textcolor[RGB]{34, 139, 34}{Correct Answer: a)}

\textcolor{red}{Unintervened: b) Heiwa believes the sea is calm and full of fish, but the fish have moved}

\textcolor{blue}{Intervened: a) Heiwa believes the sea is calm and full of fish.}

\end{tcolorbox}

\begin{tcolorbox}[fontupper=\fontsize{8}{9.6}\selectfont]
Story: Fatima is a potter in a small Moroccan village known for its vibrant ceramics. Fatima wants to create a beautiful vase as a wedding gift for her cousin. Fatima has left her kiln turned off overnight to cool down. A power surge during the night causes the kiln to turn on unexpectedly, making it dangerously hot. Fatima does not feel the heat emanating from the kiln.

\textcolor[RGB]{145, 100, 160}{Question: Does Fatima believe her kiln is off and cool or unexpectedly hot?}

\textcolor[RGB]{255, 165, 79}{Choose one of the following:}

a) Fatima believes her kiln is unexpectedly hot.

b) Fatima believes her kiln is off and cool.

\textcolor[RGB]{34, 139, 34}{Correct Answer: b)}

\textcolor{red}{Unintervened: a) Fatima believes her kiln is unexpectedly hot.}

\textcolor{blue}{Intervened: b) Fatima believes her kiln is off and cool.}

\end{tcolorbox}

\begin{tcolorbox}[fontupper=\fontsize{8}{9.6}\selectfont]
Story: Aiko is a florist in a small Japanese town. Aiko wants to create an exquisite flower arrangement for a customer's wedding. Aiko sees a vase filled with fresh, colorful flowers in her shop. Overnight, a sudden drop in temperature causes the flowers in the vase to wilt. Aiko does not notice the wilted flowers in the vase.

\textcolor[RGB]{145, 100, 160}{Question: Does Aiko believe the vase contains fresh, colorful flowers or wilted flowers?}

\textcolor[RGB]{255, 165, 79}{Choose one of the following:}

a) Aiko believes the vase contains fresh, colorful flowers.

b) Aiko believes the vase contains wilted flowers.

\textcolor[RGB]{34, 139, 34}{Correct Answer: a)}

\textcolor{red}{Unintervened: b) Aiko believes the vase contains wilted flowers. (She did not notice the}

\textcolor{blue}{Intervened: Aiko believes the vase contains fresh, colorful flowers. (At the time of reading the}

\end{tcolorbox}

\begin{tcolorbox}[fontupper=\fontsize{8}{9.6}\selectfont]
Story: Carlos is a farmer in a small village in the Andes. Carlos wants to collect fresh eggs from his chickens to sell at the local market. The nest appears to be full of eggs when Carlos checks it in the morning. A crafty fox sneaks into the chicken coop and steals all the eggs from the nest. Carlos does not see the fox stealing the eggs.

\textcolor[RGB]{145, 100, 160}{Question: Does Carlos believe the nest contains eggs or that the fox has stolen them?}

\textcolor[RGB]{255, 165, 79}{Choose one of the following:}

a) Carlos believes the fox has stolen the eggs.

b) Carlos believes the nest contains eggs.

\textcolor[RGB]{34, 139, 34}{Correct Answer: b)}

\textcolor{red}{Unintervened: a) Carlos believes the nest contains eggs.}

\textcolor{blue}{Intervened: b) Carlos believes the nest contains eggs. (at the time of asking)}

\end{tcolorbox}

\begin{tcolorbox}[fontupper=\fontsize{8}{9.6}\selectfont]
Story: Amara is an artist living in a bustling city in Nigeria. She wants to create a beautiful painting for her friend's birthday present. She finds a smooth, flawless canvas in her art studio. Later, her mischievous little brother accidentally spills black paint all over the canvas, leaving it ruined. Amara does not discover the ruined canvas.

\textcolor[RGB]{145, 100, 160}{Question: Does Amara believe the canvas is smooth and flawless or ruined with black paint?}

\textcolor[RGB]{255, 165, 79}{Choose one of the following:}

a) Amara believes the canvas is ruined with black paint.

b) Amara believes the canvas is smooth and flawless.

\textcolor[RGB]{34, 139, 34}{Correct Answer: b)}

\textcolor{red}{Unintervened: a) Amara believes the canvas is smooth and flawless (but it is actually ruined).}

\textcolor{blue}{Intervened: Amara believes the canvas is smooth and flawless. (At the time of the story, she}

\end{tcolorbox}

\begin{tcolorbox}[fontupper=\fontsize{8}{9.6}\selectfont]
Story: Fatima is a talented artist living in a small Moroccan village. Fatima wants to create a beautiful painting to donate for the local charity auction. She has a brand-new set of high-quality paintbrushes in her studio. While she is away, her younger brother accidentally ruins the paintbrushes by using them for a school project with glue. Fatima does not notice the ruined paintbrushes.

\textcolor[RGB]{145, 100, 160}{Question: Does Fatima believe she has high-quality paintbrushes or ruined paintbrushes in her studio?}

\textcolor[RGB]{255, 165, 79}{Choose one of the following:}

a) Fatima believes the paintbrushes are ruined.

b) Fatima believes she has high-quality paintbrushes in her studio.

\textcolor[RGB]{34, 139, 34}{Correct Answer: b)}

\textcolor{red}{Unintervened: a) Fatima believes the paintbrushes are ruined. (The context states that the paintbr}

\textcolor{blue}{Intervened: b) Fatima believes she has high-quality paintbrushes in her studio. (At the}

\end{tcolorbox}

\begin{tcolorbox}[fontupper=\fontsize{8}{9.6}\selectfont]
Story: Fatima is an archaeologist working at an excavation site in Egypt. She wants to carefully remove an ancient artifact from the ground to study and preserve it. The artifact appears to be intact and in good condition. Suddenly, a sandstorm blows through the area, burying the artifact in sand and causing some damage to it. Fatima does not witness the sandstorm and its effect on the artifact.

\textcolor[RGB]{145, 100, 160}{Question: Does Fatima believe the artifact is intact and in good condition or damaged by the sandstorm?}

\textcolor[RGB]{255, 165, 79}{Choose one of the following:}

a) Fatima believes the artifact is intact and in good condition.

b) Fatima believes the artifact is damaged by the sandstorm.

\textcolor[RGB]{34, 139, 34}{Correct Answer: a)}

\textcolor{red}{Unintervened: b) Fatima believes the artifact is damaged by the sandstorm.}

\textcolor{blue}{Intervened: a. (Fatima believes the artifact is intact and in good condition.)}

\end{tcolorbox}

\begin{tcolorbox}[fontupper=\fontsize{8}{9.6}\selectfont]
Story: Juan is a gardener in a small village in Mexico, tending to a flower garden near the town plaza. Juan wants to water a specific flower bed to ensure its health and beauty. Juan sees that the garden hose is neatly coiled and ready for use. Unexpectedly, a group of children playing nearby accidentally puncture the hose with a sharp stick while running around. Juan does not see the children accidentally puncture the hose.

\textcolor[RGB]{145, 100, 160}{Question: Does Juan believe the garden hose is in good condition and functional or punctured and non-functional?}

\textcolor[RGB]{255, 165, 79}{Choose one of the following:}

a) Juan believes the garden hose is in good condition and functional.

b) Juan believes the garden hose is punctured and non-functional.

\textcolor[RGB]{34, 139, 34}{Correct Answer: a)}

\textcolor{red}{Unintervened: b) Juan believes the garden hose is punctured and non-functional. (Juan}

\textcolor{blue}{Intervened: a) Juan believes the garden hose is in good condition and functional. (At the moment,}

\end{tcolorbox}

\begin{tcolorbox}[fontupper=\fontsize{8}{9.6}\selectfont]
Story: Hiro is a bonsai enthusiast in a small village in Japan. He wants to trim his favorite bonsai tree to maintain its shape and health. Hiro sees that the tree's branches are overgrown and need pruning. A gust of wind blows through Hiro's garden, snapping off several of the overgrown branches from the bonsai tree. Hiro does not see the gust of wind snapping off the branches.

\textcolor[RGB]{145, 100, 160}{Question: Does Hiro believe the bonsai tree has overgrown branches or that they have been snapped off by the wind?}

\textcolor[RGB]{255, 165, 79}{Choose one of the following:}

a) Hiro believes the bonsai tree has overgrown branches.

b) Hiro believes the branches have been snapped off by the wind.

\textcolor[RGB]{34, 139, 34}{Correct Answer: a)}

\textcolor{red}{Unintervened: b) Hiro believes the branches have been snapped off by the wind.}

\textcolor{blue}{Intervened: a. Hiro believes the bonsai tree has overgrown branches.}

\end{tcolorbox}

\begin{tcolorbox}[fontupper=\fontsize{8}{9.6}\selectfont]
Story: Tumelo is a skilled fisherman in a coastal village in South Africa. He wants to catch fish using his large, sturdy fishing net. Tumelo inspected the net before leaving home and saw that it was in perfect condition. As he walks to the shore, a mischievous seagull pecks at the net, causing a large tear in it. Tumelo does not see the seagull damaging the net.

\textcolor[RGB]{145, 100, 160}{Question: Does Tumelo believe the fishing net is in perfect condition or torn?}

\textcolor[RGB]{255, 165, 79}{Choose one of the following:}

a) Tumelo believes the fishing net is torn.

b) Tumelo believes the fishing net is in perfect condition.

\textcolor[RGB]{34, 139, 34}{Correct Answer: b)}

\textcolor{red}{Unintervened: a) Tumelo believes the fishing net is in perfect condition.}

\textcolor{blue}{Intervened: Before leaving[PAD]=\textgreater  Tumelo believes the fishing net is in perfect condition.}

\textcolor{blue}{Answer:}

\end{tcolorbox}

\begin{tcolorbox}[fontupper=\fontsize{8}{9.6}\selectfont]
Story: Kofi is a fisherman in a small coastal village in Ghana. He wants to set sail on his boat to catch fish for his family and the village market. Kofi checks the weather and sees clear skies and calm waters. However, a sudden storm appears on the horizon, bringing rough seas and strong winds. Kofi does not notice the approaching storm.

\textcolor[RGB]{145, 100, 160}{Question: Does Kofi believe the weather is perfect for sailing or is there a storm approaching?}

\textcolor[RGB]{255, 165, 79}{Choose one of the following:}

a) Kofi believes the weather is perfect for sailing.

b) Kofi believes there is a storm approaching.

\textcolor[RGB]{34, 139, 34}{Correct Answer: a)}

\textcolor{red}{Unintervened: b) Kofi believes the weather is perfect for sailing (but he is wrong).}

\textcolor{blue}{Intervened: a) Kofi believes the weather is perfect for sailing.}

\end{tcolorbox}

\begin{tcolorbox}[fontupper=\fontsize{8}{9.6}\selectfont]
Story: Mei is a fisherman in a small coastal village in China. She wants to go fishing early in the morning when the tide is high. Mei checks the tide chart and sees that the tide will be high at 5 AM. However, a strong storm far out at sea has caused the tide to come in much earlier, at 3 AM. Mei does not look out the window and does not notice the high tide.

\textcolor[RGB]{145, 100, 160}{Question: Does Mei believe the tide is high at 5 AM or 3 AM?}

\textcolor[RGB]{255, 165, 79}{Choose one of the following:}

a) Mei believes the tide is high at 5 AM.

b) Mei believes the tide is high at 3 AM.

\textcolor[RGB]{34, 139, 34}{Correct Answer: a)}

\textcolor{red}{Unintervened: b) Mei believes the tide is high at 3 AM (based on the actual situation).}

\textcolor{blue}{Intervened: Mei believes the tide is high at 5 AM. (This is what Mei currently believes}

\end{tcolorbox}

\begin{tcolorbox}[fontupper=\fontsize{8}{9.6}\selectfont]
Story: Fatima is an archaeologist working on a dig site in Egypt, searching for ancient artifacts. She wants to carefully excavate a small, fragile artifact she has discovered. Fatima observes that the soil around the artifact is soft and loose. Just as Fatima is about to begin excavating, a strong gust of wind blows through the site, hardening the soil around the artifact. Fatima does not notice the gust of wind or the change in the soil's condition.

\textcolor[RGB]{145, 100, 160}{Question: Does Fatima believe the soil around the artifact is soft and loose or hardened?}

\textcolor[RGB]{255, 165, 79}{Choose one of the following:}

a) Fatima believes the soil around the artifact has hardened.

b) Fatima believes the soil around the artifact is soft and loose.

\textcolor[RGB]{34, 139, 34}{Correct Answer: b)}

\textcolor{red}{Unintervened: a) Fatima believes the soil around the artifact has hardened.}

\textcolor{blue}{Intervened: b) Fatima believes the soil around the artifact is soft and loose.}

\end{tcolorbox}

\begin{tcolorbox}[fontupper=\fontsize{8}{9.6}\selectfont]
Story: Rohan is a potter in a small Indian village. Rohan wants to make a beautiful ceramic vase as a gift for his friend's housewarming party. Rohan selects a bag of high-quality clay in his workshop that appears to be soft and malleable. While Rohan is away for lunch, the hot sun dries the clay, making it hard and unworkable. Rohan does not notice the dried, hard clay.

\textcolor[RGB]{145, 100, 160}{Question: Does Rohan believe the bag of clay is soft and malleable or hard and unworkable?}

\textcolor[RGB]{255, 165, 79}{Choose one of the following:}

a) Rohan believes the bag of clay is hard and unworkable.

b) Rohan believes the bag of clay is soft and malleable.

\textcolor[RGB]{34, 139, 34}{Correct Answer: b)}

\textcolor{red}{Unintervened: a) Rohan believes the bag of clay is soft and malleable (but it's}

\textcolor{blue}{Intervened: Rohan believes the bag of clay is soft and malleable (before the incident), but later}

\end{tcolorbox}

\begin{tcolorbox}[fontupper=\fontsize{8}{9.6}\selectfont]
Story: Rohit is a renowned painter from India who has been commissioned to create a mural on a city building. Rohit wants to paint an intricate scene depicting the vibrant culture and history of India. Rohit selects a set of high-quality, long-lasting paint colors for the mural. While Rohit is out for lunch, a leak in the ceiling drips water onto the paint colors, diluting them and making them less durable. Rohit does not notice the water-damaged paint colors.

\textcolor[RGB]{145, 100, 160}{Question: Does Rohit believe the paint colors are of high quality and long-lasting or water-damaged and less durable?}

\textcolor[RGB]{255, 165, 79}{Choose one of the following:}

a) Rohit believes the paint colors are water-damaged and less durable.

b) Rohit believes the paint colors are of high quality and long-lasting.

\textcolor[RGB]{34, 139, 34}{Correct Answer: b)}

\textcolor{red}{Unintervened: a) Rohit does not know the true condition of the paint colors. However, due to the}

\textcolor{blue}{Intervened: b) Rohit believes the paint colors are of high quality and long-lasting. (at}

\end{tcolorbox}

\begin{tcolorbox}[fontupper=\fontsize{8}{9.6}\selectfont]
Story: Amara is a talented henna artist in a bustling Indian market. Amara wants to create intricate henna designs for a group of bridesmaids attending a wedding. Amara checks her supply of henna paste and finds it to be smooth and consistent. While Amara is busy servicing another client, a sudden rainstorm causes the henna paste to become diluted and runny. Amara does not realize the rain has affected the henna paste.

\textcolor[RGB]{145, 100, 160}{Question: Does Amara believe the henna paste is smooth and consistent or diluted and runny?}

\textcolor[RGB]{255, 165, 79}{Choose one of the following:}

a) Amara believes the henna paste is smooth and consistent.

b) Amara believes the henna paste is diluted and runny.

\textcolor[RGB]{34, 139, 34}{Correct Answer: a)}

\textcolor{red}{Unintervened: b) Amara believes the henna paste is smooth and consistent (but it is actually dilut}

\textcolor{blue}{Intervened: a) Amara believes the henna paste is smooth and consistent.}

\end{tcolorbox}

\begin{tcolorbox}[fontupper=\fontsize{8}{9.6}\selectfont]
Story: Kalyani is a traditional Indian dancer in a temple in Kerala, India. Kalyani wants to perform a special dance for the upcoming festival to pay homage to the temple's deity. The temple floor is clean and free of any obstacles, as Kalyani has prepared the space for her dance practice. An accidental spill of oil from a temple lamp causes a slippery patch on the floor. Kalyani does not notice the oil spill.

\textcolor[RGB]{145, 100, 160}{Question: Does Kalyani believe the temple floor is clean and perfect for dancing or slippery due to the oil spill?}

\textcolor[RGB]{255, 165, 79}{Choose one of the following:}

a) Kalyani believes the temple floor is clean and perfect for dancing.

b) Kalyani believes the floor is slippery due to the oil spill.

\textcolor[RGB]{34, 139, 34}{Correct Answer: a)}

\textcolor{red}{Unintervened: b) Kalyani believes the floor is slippery due to the oil spill.}

\textcolor{blue}{Intervened: a) Kalyani believes the temple floor is clean and perfect for dancing.}

\end{tcolorbox}

\begin{tcolorbox}[fontupper=\fontsize{8}{9.6}\selectfont]
Story: Naledi is a young astronomer in a remote observatory in South Africa. Naledi wants to observe a distant supernova to gather data for her research. The telescope is pointed at the coordinates of the supernova, ready for observation. A minor earthquake occurs, causing the telescope to shift its position slightly. Naledi does not notice the earthquake and its effect on the telescope.

\textcolor[RGB]{145, 100, 160}{Question: Does Naledi believe the telescope is aimed at the correct location or has shifted due to the earthquake?}

\textcolor[RGB]{255, 165, 79}{Choose one of the following:}

a) Naledi believes the telescope is aimed at the correct location.

b) Naledi believes the telescope has shifted due to the earthquake.

\textcolor[RGB]{34, 139, 34}{Correct Answer: a)}

\textcolor{red}{Unintervened: b) Naledi is unaware that the telescope has shifted due to the earthquake.}

\textcolor{blue}{Intervened: a) Naledi believes the telescope is aimed at the correct location.}

\end{tcolorbox}

\begin{tcolorbox}[fontupper=\fontsize{8}{9.6}\selectfont]
Story: Emilia is a talented baker in a small Italian town. She plans to bake a large batch of biscotti to sell at the local market. Emilia inspects the oven's temperature, which appears to be set at the correct temperature for baking. A power surge causes the oven's temperature to spike, making it too hot for baking biscotti. Emilia does not notice the oven's temperature increase.

\textcolor[RGB]{145, 100, 160}{Question: Does Emilia believe the oven is set at the correct temperature for baking biscotti or too hot?}

\textcolor[RGB]{255, 165, 79}{Choose one of the following:}

a) Emilia believes the oven is set at the correct temperature for baking biscotti.

b) Emilia believes the oven is too hot for baking biscotti.

\textcolor[RGB]{34, 139, 34}{Correct Answer: a)}

\textcolor{red}{Unintervened: b) Emilia believes the oven is too hot for baking biscotti (even though she doesn}

\textcolor{blue}{Intervened: a) Emilia believes the oven is set at the correct temperature for baking biscotti. (}

\end{tcolorbox}

\begin{tcolorbox}[fontupper=\fontsize{8}{9.6}\selectfont]
Story: Emeka is a skilled fisherman in a small Nigerian village. He wants to catch fish to feed his family and sell at the market. Emeka prepares his fishing net, which appears to be strong and without any holes. However, while Emeka is not looking, a mischievous child cuts a large hole in the net with a knife. Emeka does not see the child cutting the hole in the fishing net.

\textcolor[RGB]{145, 100, 160}{Question: Does Emeka believe the fishing net is strong and without holes, or that it has a large hole in it?}

\textcolor[RGB]{255, 165, 79}{Choose one of the following:}

a) Emeka believes the fishing net is strong and without holes.

b) Emeka believes that the fishing net has a large hole in it.

\textcolor[RGB]{34, 139, 34}{Correct Answer: a)}

\textcolor{red}{Unintervened: b) Emeka believes that the fishing net has a large hole in it (without knowing it}

\textcolor{blue}{Intervened: a) Emeka believes the fishing net is strong and without holes.}

\end{tcolorbox}

\begin{tcolorbox}[fontupper=\fontsize{8}{9.6}\selectfont]
Story: Aria is a talented baker from Italy who has been asked to bake a cake for a friend's birthday party. She wants to create a delicious, multi-layered cake that will impress everyone. Aria has a fresh batch of ingredients ready to use for the cake. However, a mischievous cat sneaks into her kitchen and nibbles on some of the ingredients, making them no longer fresh. Aria does not notice the cat or the nibbled ingredients.

\textcolor[RGB]{145, 100, 160}{Question: Does Aria believe her ingredients are fresh and suitable or nibbled on and no longer fresh?}

\textcolor[RGB]{255, 165, 79}{Choose one of the following:}

a) Aria believes the ingredients are nibbled on and no longer fresh.

b) Aria believes her ingredients are fresh and suitable for baking the cake.

\textcolor[RGB]{34, 139, 34}{Correct Answer: b)}

\textcolor{red}{Unintervened: a) Aria believes her ingredients are fresh and suitable for baking the cake (but they are actually}

\textcolor{blue}{Intervened: b) Aria believes her ingredients are fresh and suitable for baking the cake.}

\end{tcolorbox}

\begin{tcolorbox}[fontupper=\fontsize{8}{9.6}\selectfont]
Story: Ailani is a skilled Hawaiian chef preparing a traditional luau feast for her community. She wants to make a delicious laulau dish using fresh taro leaves. Ailani examines the taro leaves, which look fresh and green. However, a sudden downpour soaks the taro leaves, causing them to become wilted and less ideal for the dish. Ailani does not notice the downpour and the effect it has on the taro leaves.

\textcolor[RGB]{145, 100, 160}{Question: Does Ailani believe the taro leaves are fresh and of high quality or wilted and less ideal?}

\textcolor[RGB]{255, 165, 79}{Choose one of the following:}

a) Ailani believes the taro leaves are fresh and of high quality.

b) Ailani believes the taro leaves are wilted and less ideal.

\textcolor[RGB]{34, 139, 34}{Correct Answer: a)}

\textcolor{red}{Unintervened: b) Ailani believes the taro leaves are wilted and less ideal.}

\textcolor{blue}{Intervened: Ailani believes the taro leaves are fresh and of high quality. (Based on the context}

\end{tcolorbox}

\begin{tcolorbox}[fontupper=\fontsize{8}{9.6}\selectfont]
Story: Ravi is a fisherman living in a coastal village in India. Ravi wants to catch a large amount of fish to sell at the market and provide for his family. The waters near the shore are calm and full of fish, making it an ideal location for fishing. An underwater earthquake occurs, causing the fish to frantically swim away from the coast. Ravi does not sense the earthquake.

\textcolor[RGB]{145, 100, 160}{Question: Does Ravi believe the waters near the shore are calm and full of fish, or that the fish have frantically swum away?}

\textcolor[RGB]{255, 165, 79}{Choose one of the following:}

a) Ravi believes the fish have frantically swum away from the coast.

b) Ravi believes the waters near the shore are calm and full of fish.

\textcolor[RGB]{34, 139, 34}{Correct Answer: b)}

\textcolor{red}{Unintervened: a) Ravi believes the waters near the shore are calm and full of fish (but he is}

\textcolor{blue}{Intervened: b) Ravi believes the waters near the shore are calm and full of fish.}

\end{tcolorbox}

\begin{tcolorbox}[fontupper=\fontsize{8}{9.6}\selectfont]
Story: Surya is a fisherman in a coastal village in India. Surya plans to go fishing early in the morning to catch fish for his family and to sell at the local market. Surya's fishing net looks strong and ready for another day of fishing. Overnight, a group of mischievous monkeys rummage through the fishing equipment and badly damage Surya's fishing net. Surya does not realize the monkeys have damaged his fishing net.

\textcolor[RGB]{145, 100, 160}{Question: Does Surya believe that his fishing net is strong and ready to be used, or that it has been badly damaged?}

\textcolor[RGB]{255, 165, 79}{Choose one of the following:}

a) Surya believes his fishing net has been badly damaged.

b) Surya believes his fishing net is strong and ready to be used.

\textcolor[RGB]{34, 139, 34}{Correct Answer: b)}

\textcolor{red}{Unintervened: a) Surya believes his fishing net is strong and ready to be used (but it has}

\textcolor{blue}{Intervened: b) Surya believes his fishing net is strong and ready to be used.}

\end{tcolorbox}

\begin{tcolorbox}[fontupper=\fontsize{8}{9.6}\selectfont]
Story: Takeshi is a fisherman in a small Japanese coastal village. Takeshi's goal is to catch enough fish to provide for his family and sell at the market. The ocean is calm and teeming with fish, making it an ideal day for fishing. However, a sudden underwater volcanic eruption nearby causes the fish to scatter, leaving the area devoid of fish. Takeshi does not notice the sudden disappearance of fish in the area.

\textcolor[RGB]{145, 100, 160}{Question: Does Takeshi believe the ocean is teeming with fish, or devoid of fish due to the volcanic eruption?}

\textcolor[RGB]{255, 165, 79}{Choose one of the following:}

a) Takeshi believes the ocean is devoid of fish due to the volcanic eruption.

b) Takeshi believes the ocean is teeming with fish.

\textcolor[RGB]{34, 139, 34}{Correct Answer: b)}

\textcolor{red}{Unintervened: a) Takeshi believes the ocean is teeming with fish (but is unaware of}

\textcolor{blue}{Intervened: Takeshi believes the ocean is teeming with fish.}

\end{tcolorbox}

\begin{tcolorbox}[fontupper=\fontsize{8}{9.6}\selectfont]
Story: Hiroshi is a fisherman in a small coastal village in Japan. Hiroshi wishes to catch enough fish to feed his family and sell the surplus at the local market. The fishing nets are in excellent condition, ready to be cast into the sea. However, a curious pod of dolphins swims by and accidentally tears the nets with their sharp dorsal fins. Hiroshi does not see the dolphins or the damage they caused to the nets.

\textcolor[RGB]{145, 100, 160}{Question: Does Hiroshi believe the fishing nets are in excellent condition or torn by the dolphins?}

\textcolor[RGB]{255, 165, 79}{Choose one of the following:}

a) Hiroshi believes the fishing nets are in excellent condition.

b) Hiroshi believes the fishing nets are torn by the dolphins.

\textcolor[RGB]{34, 139, 34}{Correct Answer: a)}

\textcolor{red}{Unintervened: b) Hiroshi is unaware that the fishing nets are torn by the dolph}

\textcolor{blue}{Intervened: a) Hiroshi believes the fishing nets are in excellent condition.}

\end{tcolorbox}

\begin{tcolorbox}[fontupper=\fontsize{8}{9.6}\selectfont]
Story: Mei, a talented pastry chef from a small town in China, is preparing to bake a large cake for a local celebration. Mei wants to use a specific cake mold to create a unique and intricate design. Mei sees the cake mold hanging on the wall, clean and ready for use. However, during a brief power outage, a gust of wind blows through the kitchen, knocking the cake mold off the wall and denting it severely. Mei does not notice the dented cake mold on the floor.

\textcolor[RGB]{145, 100, 160}{Question: Does Mei believe the cake mold is clean and ready for use or severely dented and unsuitable for use?}

\textcolor[RGB]{255, 165, 79}{Choose one of the following:}

a) Mei believes the cake mold is clean and ready for use.

b) Mei believes the cake mold is severely dented and unsuitable for use.

\textcolor[RGB]{34, 139, 34}{Correct Answer: a)}

\textcolor{red}{Unintervened: b) Mei believes the cake mold is severely dented and unsuitable for use. (}

\textcolor{blue}{Intervened: a) Mei believes the cake mold is clean and ready for use. (at the start of}

\end{tcolorbox}

\noindent Changed to Incorrect Answer: 0/100

\subsection{Mistral-7B on Forward Belief, True Belief}

\noindent Changed to Correct Answer: 0/100

\noindent Changed to Incorrect Answer: 10/100
\begin{tcolorbox}[fontupper=\fontsize{8}{9.6}\selectfont]
Story: Noor is working as a barista at a busy coffee shop. Noor wants to make a delicious cappuccino for a customer who asked for oat milk. Noor grabs a milk pitcher and fills it with oat milk. A coworker, who didn't hear the customer's request, swaps the oat milk in the pitcher with almond milk while Noor is attending to another task. Noor sees her coworker swapping the milk.

\textcolor[RGB]{145, 100, 160}{Question: Does Noor believe the milk pitcher contains oat milk or almond milk?}

\textcolor[RGB]{255, 165, 79}{Choose one of the following:}

a) Noor believes the milk pitcher contains almond milk.

b) Noor believes the milk pitcher contains oat milk.

\textcolor[RGB]{34, 139, 34}{Correct Answer: a)}

\textcolor{red}{Unintervened: a) Noor believes the milk pitcher contains almond milk.}

\textcolor{blue}{Intervened: a}

\textcolor{blue}{}

\textcolor{blue}{(This question is based on the context provided, and the context states that Noor}

\end{tcolorbox}

\begin{tcolorbox}[fontupper=\fontsize{8}{9.6}\selectfont]
Story: Azola is a teacher at a small school in rural China. She wants to surprise her students with a new book for the school library. Azola got a notification from Amazon stating that the package containing the book was left at her doorstep. A gust of wind blows the package off her doorstep, and a neighbor replaces it with a different package containing a hand-knit scarf. Azola witnesses the package swap.

\textcolor[RGB]{145, 100, 160}{Question: Does Azola believe the package contains the book she ordered or a hand-knit scarf?}

\textcolor[RGB]{255, 165, 79}{Choose one of the following:}

a) Azola believes the package contains a hand-knit scarf.

b) Azola believes the package contains the book she ordered.

\textcolor[RGB]{34, 139, 34}{Correct Answer: a)}

\textcolor{red}{Unintervened: a) Azola believes the package contains a hand-knit scarf.}

\textcolor{blue}{Intervened: a}

\textcolor{blue}{}

\textcolor{blue}{(This question is a part of a multiple-choice quiz. Since there is}

\end{tcolorbox}

\begin{tcolorbox}[fontupper=\fontsize{8}{9.6}\selectfont]
Story: Ravi is a farmer in a small Indian village, tending to his mango orchard. Ravi wants to harvest ripe mangoes to sell at the local market. He notices a mango tree with a heavy branch laden with ripe mangoes. A mischievous monkey comes along and eats all the ripe mangoes on the branch while Ravi is gathering his harvesting tools. Ravi sees the monkey eating the ripe mangoes.

\textcolor[RGB]{145, 100, 160}{Question: Does Ravi believe the branch has ripe mangoes or no ripe mangoes?}

\textcolor[RGB]{255, 165, 79}{Choose one of the following:}

a) Ravi believes the branch has no ripe mangoes.

b) Ravi believes the branch has ripe mangoes.

\textcolor[RGB]{34, 139, 34}{Correct Answer: a)}

\textcolor{red}{Unintervened: a) Ravi believes the branch has no ripe mangoes.}

\textcolor{blue}{Intervened: Ravi believes the branch has ripe mangoes. (At the moment, he still believes}

\end{tcolorbox}

\begin{tcolorbox}[fontupper=\fontsize{8}{9.6}\selectfont]
Story: Farhan is a street food vendor in a bustling market in Istanbul. Farhan wants to prepare a delicious chicken shawarma wrap for a customer who asked for extra garlic sauce. Farhan takes a bottle of garlic sauce from the shelf and pours it onto the shawarma wrap. His assistant, who is busy preparing a different wrap, accidentally swaps the garlic sauce with a bottle of hot sauce, and Farhan unknowingly pours the hot sauce onto the shawarma wrap. Farhan notices the sauce swap.

\textcolor[RGB]{145, 100, 160}{Question: Does Farhan believe the sauce he poured is garlic sauce or hot sauce?}

\textcolor[RGB]{255, 165, 79}{Choose one of the following:}

a) Farhan believes the sauce he poured is hot sauce.

b) Farhan believes the sauce he poured is garlic sauce.

\textcolor[RGB]{34, 139, 34}{Correct Answer: a)}

\textcolor{red}{Unintervened: a) Farhan believes the sauce he poured is hot sauce.}

\textcolor{blue}{Intervened: b) Farhan believes the sauce he poured is garlic sauce. (At the time of pouring}

\end{tcolorbox}

\begin{tcolorbox}[fontupper=\fontsize{8}{9.6}\selectfont]
Story: Chi is a fisherman living on a small island in the South Pacific. Chi wants to go on a fishing trip to catch some fish for his family's dinner. The sea appears calm and inviting. Suddenly, a strong storm rolls in, causing the sea to become rough and dangerous. Chi notices the storm and the change in the sea.

\textcolor[RGB]{145, 100, 160}{Question: Does Chi believe the sea is calm and inviting or rough and dangerous?}

\textcolor[RGB]{255, 165, 79}{Choose one of the following:}

a) Chi believes the sea is rough and dangerous.

b) Chi believes the sea is calm and inviting.

\textcolor[RGB]{34, 139, 34}{Correct Answer: a)}

\textcolor{red}{Unintervened: a) Chi believes the sea is rough and dangerous.}

\textcolor{blue}{Intervened: a}

\textcolor{blue}{}

\textcolor{blue}{---}

\textcolor{blue}{}

\textcolor{blue}{Story: Chi is a fisherman living on a small island in}

\end{tcolorbox}

\begin{tcolorbox}[fontupper=\fontsize{8}{9.6}\selectfont]
Story: Hiro is a bonsai enthusiast in a small village in Japan. He wants to trim his favorite bonsai tree to maintain its shape and health. Hiro sees that the tree's branches are overgrown and need pruning. A gust of wind blows through Hiro's garden, snapping off several of the overgrown branches from the bonsai tree. Hiro witnesses the gust of wind snapping off the branches.

\textcolor[RGB]{145, 100, 160}{Question: Does Hiro believe the bonsai tree has overgrown branches or that they have been snapped off by the wind?}

\textcolor[RGB]{255, 165, 79}{Choose one of the following:}

a) Hiro believes the branches have been snapped off by the wind.

b) Hiro believes the bonsai tree has overgrown branches.

\textcolor[RGB]{34, 139, 34}{Correct Answer: a)}

\textcolor{red}{Unintervened: a) Hiro believes the branches have been snapped off by the wind.}

\textcolor{blue}{Intervened: Hiro believes the bonsai tree has overgrown branches. (initially)}

\end{tcolorbox}

\begin{tcolorbox}[fontupper=\fontsize{8}{9.6}\selectfont]
Story: Olumide, a skilled woodcarver in a Nigerian village, is preparing to carve a beautiful sculpture for the village chief. Olumide wants to use a sharp chisel to create intricate details on the sculpture. Olumide observes his set of chisels and sees one that appears to be sharp and in perfect condition. However, while Olumide is talking to a fellow artisan, a child from the village accidentally drops the chisel, causing it to become blunt and damaged. Olumide notices the damaged chisel on the ground.

\textcolor[RGB]{145, 100, 160}{Question: Does Olumide believe the chisel is sharp and in perfect condition or blunt and damaged?}

\textcolor[RGB]{255, 165, 79}{Choose one of the following:}

a) Olumide believes the chisel is sharp and in perfect condition.

b) Olumide believes the chisel is blunt and damaged.

\textcolor[RGB]{34, 139, 34}{Correct Answer: b)}

\textcolor{red}{Unintervened: b) Olumide believes the chisel is blunt and damaged.}

\textcolor{blue}{Intervened: a}

\textcolor{blue}{}

\textcolor{blue}{---}

\textcolor{blue}{}

\textcolor{blue}{Story: Olumide, a skilled woodcarver in a}

\end{tcolorbox}

\begin{tcolorbox}[fontupper=\fontsize{8}{9.6}\selectfont]
Story: Odion is an archaeologist conducting research at an ancient Egyptian site. He wants to discover a hidden chamber within the pyramid he is exploring. Odion observes a wall with an unusual pattern that appears to be a concealed entrance. While Odion is away, a sandstorm fills the entrance, making it impossible to access the hidden chamber. Odion sees the sand-filled entrance after the storm.

\textcolor[RGB]{145, 100, 160}{Question: Does Odion believe the wall with the unusual pattern is hiding the entrance or that the entrance is blocked by sand?}

\textcolor[RGB]{255, 165, 79}{Choose one of the following:}

a) Odion believes the entrance is blocked by sand.

b) Odion believes the wall with the unusual pattern is hiding the entrance.

\textcolor[RGB]{34, 139, 34}{Correct Answer: a)}

\textcolor{red}{Unintervened: a) Odion believes the entrance is blocked by sand.}

\textcolor{blue}{Intervened: b) Odion believes the wall with the unusual pattern is hiding the entrance.}

\end{tcolorbox}

\begin{tcolorbox}[fontupper=\fontsize{8}{9.6}\selectfont]
Story: Kunal is a talented potter in a small Indian village. Kunal wants to create a set of earthenware bowls as a gift for his friend's housewarming. Kunal inspects a large lump of clay, finding it to be moist and malleable. However, while Kunal is away for a short break, the sun beats down on the clay, drying it out and making it hard and brittle. Kunal notices the sun-dried, brittle clay.

\textcolor[RGB]{145, 100, 160}{Question: Does Kunal believe the clay is moist and malleable or hard and brittle?}

\textcolor[RGB]{255, 165, 79}{Choose one of the following:}

a) Kunal believes the clay is moist and malleable.

b) Kunal believes the clay is hard and brittle.

\textcolor[RGB]{34, 139, 34}{Correct Answer: b)}

\textcolor{red}{Unintervened: b) Kunal believes the clay is hard and brittle.}

\textcolor{blue}{Intervened: a. (Kunal believes the clay is moist and malleable.)}

\end{tcolorbox}

\begin{tcolorbox}[fontupper=\fontsize{8}{9.6}\selectfont]
Story: Naledi is a talented South African chef preparing a traditional braai for her friends and family. Naledi wants to make a delicious marinade for the meats and vegetables to be grilled. She takes a bottle of vinegar from the cupboard to mix with other ingredients for the marinade. Her cousin, who was helping in the kitchen earlier, accidentally placed a bottle of water in the vinegar's spot, and Naledi unknowingly grabs the bottle of water instead. Naledi notices the water and puts it back.

\textcolor[RGB]{145, 100, 160}{Question: Does Naledi believe the bottle she took from the cupboard contains vinegar or water?}

\textcolor[RGB]{255, 165, 79}{Choose one of the following:}

a) Naledi believes the bottle she took from the cupboard contains water.

b) Naledi believes the bottle she took from the cupboard contains vinegar.

\textcolor[RGB]{34, 139, 34}{Correct Answer: a)}

\textcolor{red}{Unintervened: a) Naledi believes the bottle she took from the cupboard contains water.}

\textcolor{blue}{Intervened: b) Naledi believes the bottle she took from the cupboard contains vinegar. (At}

\end{tcolorbox}

\subsection{DeepSeek-7B on Forward Belief, False Belief}

\noindent Changed to Correct Answer: 32/100
\begin{tcolorbox}[fontupper=\fontsize{8}{9.6}\selectfont]
Story: Zara is attending an art exhibition at a local gallery in her town. Zara wants to learn more about a particular abstract painting she admires. The label next to the artwork states that the painting is the work of a famous artist, and Zara reads the label carefully. A gallery staff member, realizing there was a mix-up, changes the label with the correct information attributing the painting to an emerging artist instead. Zara does not read the updated label.

\textcolor[RGB]{145, 100, 160}{Question: Does Zara believe the painting is the work of a famous artist or an emerging artist?}

\textcolor[RGB]{255, 165, 79}{Choose one of the following:}

a) Zara believes the painting is the work of a famous artist.

b) Zara believes the painting is the work of an emerging artist.

\textcolor[RGB]{34, 139, 34}{Correct Answer: a)}

\textcolor{red}{Unintervened: b) Zara believes the painting is the work of an emerging artist.}

\textcolor{blue}{Intervened: a) Zara believes the painting is the work of a famous artist.}

\end{tcolorbox}

\begin{tcolorbox}[fontupper=\fontsize{8}{9.6}\selectfont]
Story: Azola is a teacher at a small school in rural China. She wants to surprise her students with a new book for the school library. Azola got a notification from Amazon stating that the package containing the book was left at her doorstep. A gust of wind blows the package off her doorstep, and a neighbor replaces it with a different package containing a hand-knit scarf. Azola does not witness the package swap.

\textcolor[RGB]{145, 100, 160}{Question: Does Azola believe the package contains the book she ordered or a hand-knit scarf?}

\textcolor[RGB]{255, 165, 79}{Choose one of the following:}

a) Azola believes the package contains the book she ordered.

b) Azola believes the package contains a hand-knit scarf.

\textcolor[RGB]{34, 139, 34}{Correct Answer: a)}

\textcolor{red}{Unintervened: b) Azola believes the package contains a hand-knit scarf.}

\textcolor{blue}{Intervened: a) Azola believes the package contains the book she ordered.}

\end{tcolorbox}

\begin{tcolorbox}[fontupper=\fontsize{8}{9.6}\selectfont]
Story: Ravi is a farmer in a small Indian village, tending to his mango orchard. Ravi wants to harvest ripe mangoes to sell at the local market. He notices a mango tree with a heavy branch laden with ripe mangoes. A mischievous monkey comes along and eats all the ripe mangoes on the branch while Ravi is gathering his harvesting tools. Ravi does not see the monkey eating the ripe mangoes.

\textcolor[RGB]{145, 100, 160}{Question: Does Ravi believe the branch has ripe mangoes or no ripe mangoes?}

\textcolor[RGB]{255, 165, 79}{Choose one of the following:}

a) Ravi believes the branch has ripe mangoes.

b) Ravi believes the branch has no ripe mangoes.

\textcolor[RGB]{34, 139, 34}{Correct Answer: a)}

\textcolor{red}{Unintervened: b) Ravi believes the branch has no ripe mangoes.}

\textcolor{blue}{Intervened: a) Ravi believes the branch has ripe mangoes.}

\end{tcolorbox}

\begin{tcolorbox}[fontupper=\fontsize{8}{9.6}\selectfont]
Story: Kael is a farmer in a remote village in the Andes. Kael wants to plant a new crop of high-altitude potatoes in one of his terraced fields. The soil in the field appears rich and well-fertilized. Overnight, heavy rainfall washes away much of the fertile soil, leaving mostly rocky dirt behind. Kael does not notice the change in the soil after the rainfall.

\textcolor[RGB]{145, 100, 160}{Question: Does Kael believe the soil is rich and well-fertilized or mostly rocky dirt?}

\textcolor[RGB]{255, 165, 79}{Choose one of the following:}

a) Kael believes the soil is rich and well-fertilized.

b) Kael believes the soil is mostly rocky dirt.

\textcolor[RGB]{34, 139, 34}{Correct Answer: a)}

\textcolor{red}{Unintervened: b) Kael believes the soil is mostly rocky dirt.}

\textcolor{blue}{Intervened: a) Kael believes the soil is rich and well-fertilized.}

\end{tcolorbox}

\begin{tcolorbox}[fontupper=\fontsize{8}{9.6}\selectfont]
Story: Farhan is a street food vendor in a bustling market in Istanbul. Farhan wants to prepare a delicious chicken shawarma wrap for a customer who asked for extra garlic sauce. Farhan takes a bottle of garlic sauce from the shelf and pours it onto the shawarma wrap. His assistant, who is busy preparing a different wrap, accidentally swaps the garlic sauce with a bottle of hot sauce, and Farhan unknowingly pours the hot sauce onto the shawarma wrap. Farhan does not notice the sauce swap.

\textcolor[RGB]{145, 100, 160}{Question: Does Farhan believe the sauce he poured is garlic sauce or hot sauce?}

\textcolor[RGB]{255, 165, 79}{Choose one of the following:}

a) Farhan believes the sauce he poured is garlic sauce.

b) Farhan believes the sauce he poured is hot sauce.

\textcolor[RGB]{34, 139, 34}{Correct Answer: a)}

\textcolor{red}{Unintervened: b) Farhan believes the sauce he poured is hot sauce.}

\textcolor{blue}{Intervened: a) Farhan believes the sauce he poured is garlic sauce.}

\end{tcolorbox}

\begin{tcolorbox}[fontupper=\fontsize{8}{9.6}\selectfont]
Story: Lila is a beekeeper in a small village in Eastern Europe. She wants to harvest honey from one of her beehives. The beehive is full of honey, as Lila observed during her last inspection. A sudden cold snap occurs overnight, causing the bees to consume the honey to stay warm and survive, leaving the beehive nearly empty. Lila does not notice the cold snap and the decreased honey in her beehive.

\textcolor[RGB]{145, 100, 160}{Question: Does Lila believe the beehive is full of honey or nearly empty?}

\textcolor[RGB]{255, 165, 79}{Choose one of the following:}

a) Lila believes the beehive is full of honey.

b) Lila believes the beehive is nearly empty.

\textcolor[RGB]{34, 139, 34}{Correct Answer: a)}

\textcolor{red}{Unintervened: b) Lila believes the beehive is nearly empty.}

\textcolor{blue}{Intervened: a) Lila believes the beehive is full of honey.}

\end{tcolorbox}

\begin{tcolorbox}[fontupper=\fontsize{8}{9.6}\selectfont]
Story: Kavi is a chef at a popular Indian restaurant in a bustling city. Kavi wants to make his signature dish, a spicy curry, for the customers tonight. Kavi checks the pantry and sees a full bottle of his favorite hot sauce. When Kavi is away, a new chef accidentally knocks over the hot sauce, spilling its contents all over the floor. Kavi does not see the spilled hot sauce on the floor.

\textcolor[RGB]{145, 100, 160}{Question: Does Kavi believe there is a full bottle of hot sauce available or that the hot sauce has been spilled?}

\textcolor[RGB]{255, 165, 79}{Choose one of the following:}

a) Kavi believes the hot sauce has been spilled.

b) Kavi believes there is a full bottle of hot sauce available.

\textcolor[RGB]{34, 139, 34}{Correct Answer: b)}

\textcolor{red}{Unintervened: a) Kavi believes the hot sauce has been spilled.}

\textcolor{blue}{Intervened: b) Kavi believes there is a full bottle of hot sauce available.}

\end{tcolorbox}

\begin{tcolorbox}[fontupper=\fontsize{8}{9.6}\selectfont]
Story: Juan is a young chef in a small Mexican town. He wants to make a delicious salsa for the upcoming town festival. Juan buys fresh tomatoes from the market, which he believes are ripe and perfect for the salsa. While Juan is away from his kitchen, a mischievous cat knocks over the bag of tomatoes, crushing them and making them unsuitable for the salsa. Juan does not notice the crushed tomatoes on the floor.

\textcolor[RGB]{145, 100, 160}{Question: Does Juan believe the tomatoes are ripe and perfect or crushed and unsuitable?}

\textcolor[RGB]{255, 165, 79}{Choose one of the following:}

a) Juan believes the tomatoes are ripe and perfect.

b) Juan believes the tomatoes are crushed and unsuitable.

\textcolor[RGB]{34, 139, 34}{Correct Answer: a)}

\textcolor{red}{Unintervened: b) Juan believes the tomatoes are crushed and unsuitable.}

\textcolor{blue}{Intervened: a) Juan believes the tomatoes are ripe and perfect.}

\end{tcolorbox}

\begin{tcolorbox}[fontupper=\fontsize{8}{9.6}\selectfont]
Story: Isabela is a baker in a small Argentinian town known for its delicious pastries. Isabela wants to make a batch of alfajores, a traditional Argentinian dessert, for her bakery's customers. Isabela gets a bag of powdered sugar from her pantry to dust the alfajores. Her assistant, who was organizing the pantry earlier, accidentally placed a bag of flour in the powdered sugar's spot, and Isabela unknowingly grabs the bag of flour instead. Isabela does not notice the flour.

\textcolor[RGB]{145, 100, 160}{Question: Does Isabela believe the bag she took from the pantry contains powdered sugar or flour?}

\textcolor[RGB]{255, 165, 79}{Choose one of the following:}

a) Isabela believes the bag she took from the pantry contains powdered sugar.

b) Isabela believes the bag she took from the pantry contains flour.

\textcolor[RGB]{34, 139, 34}{Correct Answer: a)}

\textcolor{red}{Unintervened: b) Isabela believes the bag she took from the pantry contains flour.}

\textcolor{blue}{Intervened: a) Isabela believes the bag she took from the pantry contains powdered sugar.}

\end{tcolorbox}

\begin{tcolorbox}[fontupper=\fontsize{8}{9.6}\selectfont]
Story: Chi is a fisherman living on a small island in the South Pacific. Chi wants to go on a fishing trip to catch some fish for his family's dinner. The sea appears calm and inviting. Suddenly, a strong storm rolls in, causing the sea to become rough and dangerous. Chi does not notice the storm and the change in the sea.

\textcolor[RGB]{145, 100, 160}{Question: Does Chi believe the sea is calm and inviting or rough and dangerous?}

\textcolor[RGB]{255, 165, 79}{Choose one of the following:}

a) Chi believes the sea is calm and inviting.

b) Chi believes the sea is rough and dangerous.

\textcolor[RGB]{34, 139, 34}{Correct Answer: a)}

\textcolor{red}{Unintervened: b) Chi believes the sea is rough and dangerous.}

\textcolor{blue}{Intervened: a)}

\end{tcolorbox}

\begin{tcolorbox}[fontupper=\fontsize{8}{9.6}\selectfont]
Story: Mei is a pearl diver in a small coastal town in Japan. Mei wants to dive deep into the sea to find a rare and valuable black pearl for her grandmother's birthday gift. The water in the cove is calm with barely any waves. Suddenly, a storm rolls in, and the once calm water becomes rough and choppy. Mei does not notice the change in the water as the storm approaches.

\textcolor[RGB]{145, 100, 160}{Question: Does Mei believe the water is calm and perfect for diving or rough and choppy due to the storm?}

\textcolor[RGB]{255, 165, 79}{Choose one of the following:}

a) Mei believes the water is calm and perfect for diving.

b) Mei believes the water is rough and choppy due to the storm.

\textcolor[RGB]{34, 139, 34}{Correct Answer: a)}

\textcolor{red}{Unintervened: b) Mei believes the water is rough and choppy due to the storm.}

\textcolor{blue}{Intervened: a) Mei believes the water is calm and perfect for diving.}

\end{tcolorbox}

\begin{tcolorbox}[fontupper=\fontsize{8}{9.6}\selectfont]
Story: Amina is an archaeologist working on an excavation site in Egypt. Amina wants to carefully remove a valuable artifact from the ground using her brush and trowel. The artifact appears to be intact and undamaged. Suddenly, the ground shifts slightly, causing a small crack to form on the artifact. Amina does not notice the ground shift and the crack on the artifact.

\textcolor[RGB]{145, 100, 160}{Question: Does Amina believe the artifact is in perfect condition or cracked?}

\textcolor[RGB]{255, 165, 79}{Choose one of the following:}

a) Amina believes the artifact is in perfect condition.

b) Amina believes the artifact is cracked.

\textcolor[RGB]{34, 139, 34}{Correct Answer: a)}

\textcolor{red}{Unintervened: b) Amina believes the artifact is cracked.}

\textcolor{blue}{Intervened: a) Amina believes the artifact is in perfect condition.}

\end{tcolorbox}

\begin{tcolorbox}[fontupper=\fontsize{8}{9.6}\selectfont]
Story: Aiko is a florist in a small Japanese town. Aiko wants to create an exquisite flower arrangement for a customer's wedding. Aiko sees a vase filled with fresh, colorful flowers in her shop. Overnight, a sudden drop in temperature causes the flowers in the vase to wilt. Aiko does not notice the wilted flowers in the vase.

\textcolor[RGB]{145, 100, 160}{Question: Does Aiko believe the vase contains fresh, colorful flowers or wilted flowers?}

\textcolor[RGB]{255, 165, 79}{Choose one of the following:}

a) Aiko believes the vase contains fresh, colorful flowers.

b) Aiko believes the vase contains wilted flowers.

\textcolor[RGB]{34, 139, 34}{Correct Answer: a)}

\textcolor{red}{Unintervened: b) Aiko believes the vase contains wilted flowers.}

\textcolor{blue}{Intervened: a) Aiko believes the vase contains fresh, colorful flowers.}

\end{tcolorbox}

\begin{tcolorbox}[fontupper=\fontsize{8}{9.6}\selectfont]
Story: Fatima is an archaeologist working at an excavation site in Egypt. She wants to carefully remove an ancient artifact from the ground to study and preserve it. The artifact appears to be intact and in good condition. Suddenly, a sandstorm blows through the area, burying the artifact in sand and causing some damage to it. Fatima does not witness the sandstorm and its effect on the artifact.

\textcolor[RGB]{145, 100, 160}{Question: Does Fatima believe the artifact is intact and in good condition or damaged by the sandstorm?}

\textcolor[RGB]{255, 165, 79}{Choose one of the following:}

a) Fatima believes the artifact is intact and in good condition.

b) Fatima believes the artifact is damaged by the sandstorm.

\textcolor[RGB]{34, 139, 34}{Correct Answer: a)}

\textcolor{red}{Unintervened: b) Fatima believes the artifact is damaged by the sandstorm.}

\textcolor{blue}{Intervened: a) Fatima believes the artifact is intact and in good condition.}

\end{tcolorbox}

\begin{tcolorbox}[fontupper=\fontsize{8}{9.6}\selectfont]
Story: Juan is a gardener in a small village in Mexico, tending to a flower garden near the town plaza. Juan wants to water a specific flower bed to ensure its health and beauty. Juan sees that the garden hose is neatly coiled and ready for use. Unexpectedly, a group of children playing nearby accidentally puncture the hose with a sharp stick while running around. Juan does not see the children accidentally puncture the hose.

\textcolor[RGB]{145, 100, 160}{Question: Does Juan believe the garden hose is in good condition and functional or punctured and non-functional?}

\textcolor[RGB]{255, 165, 79}{Choose one of the following:}

a) Juan believes the garden hose is in good condition and functional.

b) Juan believes the garden hose is punctured and non-functional.

\textcolor[RGB]{34, 139, 34}{Correct Answer: a)}

\textcolor{red}{Unintervened: b) Juan believes the garden hose is punctured and non-functional.}

\textcolor{blue}{Intervened: a) Juan believes the garden hose is in good condition and functional.}

\end{tcolorbox}

\begin{tcolorbox}[fontupper=\fontsize{8}{9.6}\selectfont]
Story: Hiroshi is a Japanese gardener tending to a zen garden in Kyoto. Hiroshi desires to create perfect harmony in the garden by carefully raking the sand into beautiful patterns. Hiroshi notices that the sand in the garden is smooth and untouched. A sudden gust of wind blows fallen leaves onto the sand, disrupting the smooth surface. Hiroshi does not see the gust of wind or the leaves falling onto the sand.

\textcolor[RGB]{145, 100, 160}{Question: Does Hiroshi believe the sand is smooth and untouched or disrupted by fallen leaves?}

\textcolor[RGB]{255, 165, 79}{Choose one of the following:}

a) Hiroshi believes the sand is smooth and untouched.

b) Hiroshi believes the sand is disrupted by fallen leaves.

\textcolor[RGB]{34, 139, 34}{Correct Answer: a)}

\textcolor{red}{Unintervened: b) Hiroshi believes the sand is disrupted by fallen leaves.}

\textcolor{blue}{Intervened: a) Hiroshi believes the sand is smooth and untouched.}

\end{tcolorbox}

\begin{tcolorbox}[fontupper=\fontsize{8}{9.6}\selectfont]
Story: Hiro is a bonsai enthusiast in a small village in Japan. He wants to trim his favorite bonsai tree to maintain its shape and health. Hiro sees that the tree's branches are overgrown and need pruning. A gust of wind blows through Hiro's garden, snapping off several of the overgrown branches from the bonsai tree. Hiro does not see the gust of wind snapping off the branches.

\textcolor[RGB]{145, 100, 160}{Question: Does Hiro believe the bonsai tree has overgrown branches or that they have been snapped off by the wind?}

\textcolor[RGB]{255, 165, 79}{Choose one of the following:}

a) Hiro believes the bonsai tree has overgrown branches.

b) Hiro believes the branches have been snapped off by the wind.

\textcolor[RGB]{34, 139, 34}{Correct Answer: a)}

\textcolor{red}{Unintervened: b) Hiro believes the branches have been snapped off by the wind.}

\textcolor{blue}{Intervened: a) Hiro believes the bonsai tree has overgrown branches.}

\end{tcolorbox}

\begin{tcolorbox}[fontupper=\fontsize{8}{9.6}\selectfont]
Story: Kofi is a fisherman in a small coastal village in Ghana. He wants to set sail on his boat to catch fish for his family and the village market. Kofi checks the weather and sees clear skies and calm waters. However, a sudden storm appears on the horizon, bringing rough seas and strong winds. Kofi does not notice the approaching storm.

\textcolor[RGB]{145, 100, 160}{Question: Does Kofi believe the weather is perfect for sailing or is there a storm approaching?}

\textcolor[RGB]{255, 165, 79}{Choose one of the following:}

a) Kofi believes the weather is perfect for sailing.

b) Kofi believes there is a storm approaching.

\textcolor[RGB]{34, 139, 34}{Correct Answer: a)}

\textcolor{red}{Unintervened: b) Kofi believes there is a storm approaching.}

\textcolor{blue}{Intervened: a) Kofi believes the weather is perfect for sailing.}

\end{tcolorbox}

\begin{tcolorbox}[fontupper=\fontsize{8}{9.6}\selectfont]
Story: Mei is a fisherman in a small coastal village in China. She wants to go fishing early in the morning when the tide is high. Mei checks the tide chart and sees that the tide will be high at 5 AM. However, a strong storm far out at sea has caused the tide to come in much earlier, at 3 AM. Mei does not look out the window and does not notice the high tide.

\textcolor[RGB]{145, 100, 160}{Question: Does Mei believe the tide is high at 5 AM or 3 AM?}

\textcolor[RGB]{255, 165, 79}{Choose one of the following:}

a) Mei believes the tide is high at 5 AM.

b) Mei believes the tide is high at 3 AM.

\textcolor[RGB]{34, 139, 34}{Correct Answer: a)}

\textcolor{red}{Unintervened: b) Mei believes the tide is high at 3 AM.}

\textcolor{blue}{Intervened: a) Mei believes the tide is high at 5 AM.}

\end{tcolorbox}

\begin{tcolorbox}[fontupper=\fontsize{8}{9.6}\selectfont]
Story: Kwame is a skilled woodcarver in a small African village known for its exceptional craftsmanship. Kwame wants to create a beautiful wooden sculpture for the upcoming village festival. Kwame examines a large piece of wood and notices it is strong and free of imperfections. While Kwame is away, termites infest the wood, weakening its structure and leaving it unsuitable for carving. Kwame does not notice the termite-infested wood.

\textcolor[RGB]{145, 100, 160}{Question: Does Kwame believe the piece of wood is strong and free of imperfections or weakened and infested with termites?}

\textcolor[RGB]{255, 165, 79}{Choose one of the following:}

a) Kwame believes the piece of wood is strong and free of imperfections.

b) Kwame believes the piece of wood is weakened and infested with termites.

\textcolor[RGB]{34, 139, 34}{Correct Answer: a)}

\textcolor{red}{Unintervened: b) Kwame believes the piece of wood is weakened and infested with termites.}

\textcolor{blue}{Intervened: a) Kwame believes the piece of wood is strong and free of imperfections.}

\end{tcolorbox}

\begin{tcolorbox}[fontupper=\fontsize{8}{9.6}\selectfont]
Story: Luka is a skilled woodworker in a small Croatian village known for its traditional crafts. Luka intends to carve an intricate wooden sculpture as a gift for his friend's wedding. Luka spots a large piece of high-quality wood in his workshop. Just when Luka steps away for a moment, his dog chews on the wood, leaving teeth marks and damage to the surface. Luka does not see the damaged wood.

\textcolor[RGB]{145, 100, 160}{Question: Does Luka believe the wood is perfect for carving or damaged with teeth marks?}

\textcolor[RGB]{255, 165, 79}{Choose one of the following:}

a) Luka believes the wood is perfect for carving.

b) Luka believes the wood is damaged with teeth marks.

\textcolor[RGB]{34, 139, 34}{Correct Answer: a)}

\textcolor{red}{Unintervened: b) Luka believes the wood is damaged with teeth marks.}

\textcolor{blue}{Intervened: a) Luka believes the wood is perfect for carving.}

\end{tcolorbox}

\begin{tcolorbox}[fontupper=\fontsize{8}{9.6}\selectfont]
Story: Amara is a talented henna artist in a bustling Indian market. Amara wants to create intricate henna designs for a group of bridesmaids attending a wedding. Amara checks her supply of henna paste and finds it to be smooth and consistent. While Amara is busy servicing another client, a sudden rainstorm causes the henna paste to become diluted and runny. Amara does not realize the rain has affected the henna paste.

\textcolor[RGB]{145, 100, 160}{Question: Does Amara believe the henna paste is smooth and consistent or diluted and runny?}

\textcolor[RGB]{255, 165, 79}{Choose one of the following:}

a) Amara believes the henna paste is smooth and consistent.

b) Amara believes the henna paste is diluted and runny.

\textcolor[RGB]{34, 139, 34}{Correct Answer: a)}

\textcolor{red}{Unintervened: b) Amara believes the henna paste is diluted and runny.}

\textcolor{blue}{Intervened: a) Amara believes the henna paste is smooth and consistent.}

\end{tcolorbox}

\begin{tcolorbox}[fontupper=\fontsize{8}{9.6}\selectfont]
Story: Leilani is a skilled hula dancer from Hawaii who has been asked to perform at a cultural event. Leilani wants to wear her favorite traditional hula skirt made of ti leaves for the performance. Leilani sees her ti leaf skirt hanging in her closet, appearing to be in good condition. While Leilani is away from her room, her younger brother accidentally spills a glass of water near the closet, causing her ti leaf skirt to become wet and wilted. Leilani does not notice the wet and wilted ti leaf skirt.

\textcolor[RGB]{145, 100, 160}{Question: Does Leilani believe the ti leaf skirt is in good condition or wet and wilted?}

\textcolor[RGB]{255, 165, 79}{Choose one of the following:}

a) Leilani believes the ti leaf skirt is in good condition.

b) Leilani believes the ti leaf skirt is wet and wilted.

\textcolor[RGB]{34, 139, 34}{Correct Answer: a)}

\textcolor{red}{Unintervened: b) Leilani believes the ti leaf skirt is wet and wilted.}

\textcolor{blue}{Intervened: a) Leilani believes the ti leaf skirt is in good condition.}

\end{tcolorbox}

\begin{tcolorbox}[fontupper=\fontsize{8}{9.6}\selectfont]
Story: Isabella is a skilled baker in a small Italian town. She wants to bake a fresh batch of her famous ciabatta bread for the local market. Isabella has prepared the dough earlier and left it to rise in a warm spot in her kitchen. A curious cat enters the kitchen and jumps onto the counter, accidentally flattening the dough. Isabella does not see the cat and the flattened dough.

\textcolor[RGB]{145, 100, 160}{Question: Does Isabella believe the dough has risen enough or has it been flattened?}

\textcolor[RGB]{255, 165, 79}{Choose one of the following:}

a) Isabella believes the dough has risen enough.

b) Isabella believes the dough has been flattened.

\textcolor[RGB]{34, 139, 34}{Correct Answer: a)}

\textcolor{red}{Unintervened: b) Isabella believes the dough has been flattened.}

\textcolor{blue}{Intervened: a) Isabella believes the dough has risen enough.}

\end{tcolorbox}

\begin{tcolorbox}[fontupper=\fontsize{8}{9.6}\selectfont]
Story: Kalyani is a traditional Indian dancer in a temple in Kerala, India. Kalyani wants to perform a special dance for the upcoming festival to pay homage to the temple's deity. The temple floor is clean and free of any obstacles, as Kalyani has prepared the space for her dance practice. An accidental spill of oil from a temple lamp causes a slippery patch on the floor. Kalyani does not notice the oil spill.

\textcolor[RGB]{145, 100, 160}{Question: Does Kalyani believe the temple floor is clean and perfect for dancing or slippery due to the oil spill?}

\textcolor[RGB]{255, 165, 79}{Choose one of the following:}

a) Kalyani believes the temple floor is clean and perfect for dancing.

b) Kalyani believes the floor is slippery due to the oil spill.

\textcolor[RGB]{34, 139, 34}{Correct Answer: a)}

\textcolor{red}{Unintervened: b) Kalyani believes the floor is slippery due to the oil spill.}

\textcolor{blue}{Intervened: a) Kalyani believes the temple floor is clean and perfect for dancing.}

\end{tcolorbox}

\begin{tcolorbox}[fontupper=\fontsize{8}{9.6}\selectfont]
Story: Amara is a skilled potter in a small Nigerian village. Amara wants to create a beautiful, large clay pot to be used in the village's upcoming festival. The kiln, used for firing pottery, has been heated to the appropriate temperature for the clay. Unbeknownst to her, a sudden gust of wind blows open the kiln door, causing the temperature to drop significantly. Amara does not notice the kiln door has been blown open.

\textcolor[RGB]{145, 100, 160}{Question: Does Amara believe the kiln is at the right temperature for firing the large pot or that the temperature has dropped significantly?}

\textcolor[RGB]{255, 165, 79}{Choose one of the following:}

a) Amara believes the kiln is at the right temperature for firing the large pot.

b) Amara believes the temperature in the kiln has dropped significantly.

\textcolor[RGB]{34, 139, 34}{Correct Answer: a)}

\textcolor{red}{Unintervened: b) Amara believes the temperature in the kiln has dropped significantly.}

\textcolor{blue}{Intervened: a) Amara believes the kiln is at the right temperature for firing the large pot}

\end{tcolorbox}

\begin{tcolorbox}[fontupper=\fontsize{8}{9.6}\selectfont]
Story: Naledi is a young astronomer in a remote observatory in South Africa. Naledi wants to observe a distant supernova to gather data for her research. The telescope is pointed at the coordinates of the supernova, ready for observation. A minor earthquake occurs, causing the telescope to shift its position slightly. Naledi does not notice the earthquake and its effect on the telescope.

\textcolor[RGB]{145, 100, 160}{Question: Does Naledi believe the telescope is aimed at the correct location or has shifted due to the earthquake?}

\textcolor[RGB]{255, 165, 79}{Choose one of the following:}

a) Naledi believes the telescope is aimed at the correct location.

b) Naledi believes the telescope has shifted due to the earthquake.

\textcolor[RGB]{34, 139, 34}{Correct Answer: a)}

\textcolor{red}{Unintervened: b) Naledi believes the telescope has shifted due to the earthquake.}

\textcolor{blue}{Intervened: a) Naledi believes the telescope is aimed at the correct location.}

\end{tcolorbox}

\begin{tcolorbox}[fontupper=\fontsize{8}{9.6}\selectfont]
Story: Emeka is a skilled fisherman in a small Nigerian village. He wants to catch fish to feed his family and sell at the market. Emeka prepares his fishing net, which appears to be strong and without any holes. However, while Emeka is not looking, a mischievous child cuts a large hole in the net with a knife. Emeka does not see the child cutting the hole in the fishing net.

\textcolor[RGB]{145, 100, 160}{Question: Does Emeka believe the fishing net is strong and without holes, or that it has a large hole in it?}

\textcolor[RGB]{255, 165, 79}{Choose one of the following:}

a) Emeka believes the fishing net is strong and without holes.

b) Emeka believes that the fishing net has a large hole in it.

\textcolor[RGB]{34, 139, 34}{Correct Answer: a)}

\textcolor{red}{Unintervened: b) Emeka believes that the fishing net has a large hole in it.}

\textcolor{blue}{Intervened: a) Emeka believes the fishing net is strong and without holes.}

\end{tcolorbox}

\begin{tcolorbox}[fontupper=\fontsize{8}{9.6}\selectfont]
Story: Hiroshi is a fisherman in a small coastal village in Japan. Hiroshi wishes to catch enough fish to feed his family and sell the surplus at the local market. The fishing nets are in excellent condition, ready to be cast into the sea. However, a curious pod of dolphins swims by and accidentally tears the nets with their sharp dorsal fins. Hiroshi does not see the dolphins or the damage they caused to the nets.

\textcolor[RGB]{145, 100, 160}{Question: Does Hiroshi believe the fishing nets are in excellent condition or torn by the dolphins?}

\textcolor[RGB]{255, 165, 79}{Choose one of the following:}

a) Hiroshi believes the fishing nets are in excellent condition.

b) Hiroshi believes the fishing nets are torn by the dolphins.

\textcolor[RGB]{34, 139, 34}{Correct Answer: a)}

\textcolor{red}{Unintervened: b) Hiroshi believes the fishing nets are torn by the dolphins.}

\textcolor{blue}{Intervened: a) Hiroshi believes the fishing nets are in excellent condition.}

\end{tcolorbox}

\begin{tcolorbox}[fontupper=\fontsize{8}{9.6}\selectfont]
Story: Kwame is a skilled woodcarver in a small village in Ghana. Kwame wants to create a beautiful wooden sculpture of an elephant for his friend's birthday. He selects a block of wood from his workshop that appears to be solid and strong. However, as he begins to carve, he discovers a hidden pocket of termites inside the wood, rendering it unsuitable for his project. Kwame does not notice the termites inside the block of wood.

\textcolor[RGB]{145, 100, 160}{Question: Does Kwame believe the block of wood is sturdy and suitable for carving or that it contains termites?}

\textcolor[RGB]{255, 165, 79}{Choose one of the following:}

a) Kwame believes the block of wood contains termites.

b) Kwame believes the block of wood is sturdy and suitable for carving.

\textcolor[RGB]{34, 139, 34}{Correct Answer: b)}

\textcolor{red}{Unintervened: a) Kwame believes the block of wood contains termites.}

\textcolor{blue}{Intervened: b)}

\end{tcolorbox}

\begin{tcolorbox}[fontupper=\fontsize{8}{9.6}\selectfont]
Story: Naledi is a talented South African chef preparing a traditional braai for her friends and family. Naledi wants to make a delicious marinade for the meats and vegetables to be grilled. She takes a bottle of vinegar from the cupboard to mix with other ingredients for the marinade. Her cousin, who was helping in the kitchen earlier, accidentally placed a bottle of water in the vinegar's spot, and Naledi unknowingly grabs the bottle of water instead. Naledi does not notice the water.

\textcolor[RGB]{145, 100, 160}{Question: Does Naledi believe the bottle she took from the cupboard contains vinegar or water?}

\textcolor[RGB]{255, 165, 79}{Choose one of the following:}

a) Naledi believes the bottle she took from the cupboard contains vinegar.

b) Naledi believes the bottle she took from the cupboard contains water.

\textcolor[RGB]{34, 139, 34}{Correct Answer: a)}

\textcolor{red}{Unintervened: b) Naledi believes the bottle she took from the cupboard contains water.}

\textcolor{blue}{Intervened: a) Naledi believes the bottle she took from the cupboard contains vinegar.}

\end{tcolorbox}

\begin{tcolorbox}[fontupper=\fontsize{8}{9.6}\selectfont]
Story: Mei, a talented pastry chef from a small town in China, is preparing to bake a large cake for a local celebration. Mei wants to use a specific cake mold to create a unique and intricate design. Mei sees the cake mold hanging on the wall, clean and ready for use. However, during a brief power outage, a gust of wind blows through the kitchen, knocking the cake mold off the wall and denting it severely. Mei does not notice the dented cake mold on the floor.

\textcolor[RGB]{145, 100, 160}{Question: Does Mei believe the cake mold is clean and ready for use or severely dented and unsuitable for use?}

\textcolor[RGB]{255, 165, 79}{Choose one of the following:}

a) Mei believes the cake mold is clean and ready for use.

b) Mei believes the cake mold is severely dented and unsuitable for use.

\textcolor[RGB]{34, 139, 34}{Correct Answer: a)}

\textcolor{red}{Unintervened: b) Mei believes the cake mold is severely dented and unsuitable for use.}

\textcolor{blue}{Intervened: a) Mei believes the cake mold is clean and ready for use.}

\end{tcolorbox}
\noindent Changed to Incorrect Answer: 5/100
\begin{tcolorbox}[fontupper=\fontsize{8}{9.6}\selectfont]
Story: Aniket is a marine biologist studying coral reefs off the coast of India. Aniket needs to collect samples of coral to analyze the effects of climate change on the reef. Aniket spots a healthy-looking coral formation in a specific area of the reef. A sudden wave surge stirs up sediment, covering the once healthy coral formation and causing it to become damaged. Aniket does not notice the wave surge or the sediment covering the coral.

\textcolor[RGB]{145, 100, 160}{Question: Does Aniket believe the coral formation is healthy or damaged?}

\textcolor[RGB]{255, 165, 79}{Choose one of the following:}

a) Aniket believes the coral formation is damaged.

b) Aniket believes the coral formation is healthy.

\textcolor[RGB]{34, 139, 34}{Correct Answer: b)}

\textcolor{red}{Unintervened: b) Aniket believes the coral formation is healthy.}

\textcolor{blue}{Intervened: a) Aniket believes the coral formation is damaged.}

\end{tcolorbox}

\begin{tcolorbox}[fontupper=\fontsize{8}{9.6}\selectfont]
Story: Fatima is an archaeologist working on a dig site in Egypt. Fatima hopes to find an ancient artifact that could shed light on an important historical event. During her excavation, Fatima discovers a clay tablet with faded inscriptions that she believes might be valuable. As a sandstorm approaches, the strong winds blow sand onto the tablet, erasing the remaining inscriptions and making the tablet indistinguishable from any other piece of clay. Fatima does not notice the sandstorm or the damage to the clay tablet.

\textcolor[RGB]{145, 100, 160}{Question: Does Fatima believe the clay tablet has important historical inscriptions or has become indistinguishable from any other piece of clay?}

\textcolor[RGB]{255, 165, 79}{Choose one of the following:}

a) Fatima believes the clay tablet has become indistinguishable from any other piece of clay.

b) Fatima believes the clay tablet has important historical inscriptions.

\textcolor[RGB]{34, 139, 34}{Correct Answer: b)}

\textcolor{red}{Unintervened: b) Fatima believes the clay tablet has important historical inscriptions.}

\textcolor{blue}{Intervened: a) Fatima believes the clay tablet has become indistinguishable from any other piece of clay.}

\end{tcolorbox}

\begin{tcolorbox}[fontupper=\fontsize{8}{9.6}\selectfont]
Story: Mei is a dedicated librarian in a small town in China. She wants to organize a reading corner for children using a large, colorful rug. Mei sees the rug rolled up in a storage room, appearing to be in good condition. However, when Mei is not looking, a leak in the storage room ceiling drips water onto the rug, causing it to become soaked and damp. Mei does not notice the dampness of the rug and the leak in the ceiling.

\textcolor[RGB]{145, 100, 160}{Question: Does Mei believe the rug is in good condition or damp and soaked?}

\textcolor[RGB]{255, 165, 79}{Choose one of the following:}

a) Mei believes the rug is damp and soaked.

b) Mei believes the rug is in good condition.

\textcolor[RGB]{34, 139, 34}{Correct Answer: b)}

\textcolor{red}{Unintervened: b) Mei believes the rug is in good condition.}

\textcolor{blue}{Intervened: a) Mei believes the rug is damp and soaked.}

\end{tcolorbox}

\begin{tcolorbox}[fontupper=\fontsize{8}{9.6}\selectfont]
Story: Olabisi is a beekeeper in a small town in Nigeria. Olabisi wants to collect honey from the beehives to sell at the local market. The bees appear to be calm and undisturbed as they go about their daily activities. Out of nowhere, a group of children playing nearby accidentally throw a ball into the beehives, agitating the bees and making them aggressive. Olabisi does not notice the children's ball hitting the beehives.

\textcolor[RGB]{145, 100, 160}{Question: Does Olabisi believe the bees are calm or agitated and aggressive?}

\textcolor[RGB]{255, 165, 79}{Choose one of the following:}

a) Olabisi believes the bees are agitated and aggressive.

b) Olabisi believes the bees are calm.

\textcolor[RGB]{34, 139, 34}{Correct Answer: b)}

\textcolor{red}{Unintervened: b) Olabisi believes the bees are calm.}

\textcolor{blue}{Intervened: a) Olabisi believes the bees are agitated and aggressive.}

\end{tcolorbox}

\begin{tcolorbox}[fontupper=\fontsize{8}{9.6}\selectfont]
Story: Lila is a talented violinist who lives in a bustling city in India. Today, she has an important audition for a renowned symphony orchestra. Lila picks up her violin case, which she assumes contains her valuable violin. However, her younger brother, who was playing around earlier, accidentally swapped her violin case with his identical-looking case containing a beginner's violin. Lila does not notice the swapped violin cases.

\textcolor[RGB]{145, 100, 160}{Question: Does Lila believe the violin case contains her valuable violin or a beginner's violin?}

\textcolor[RGB]{255, 165, 79}{Choose one of the following:}

a) Lila believes the violin case contains a beginner's violin.

b) Lila believes the violin case contains her valuable violin.

\textcolor[RGB]{34, 139, 34}{Correct Answer: b)}

\textcolor{red}{Unintervened: b) Lila believes the violin case contains her valuable violin.}

\textcolor{blue}{Intervened: a) Lila believes the violin case contains a beginner's violin.}

\end{tcolorbox}

\subsection{DeepSeek-7B on Forward Belief, True Belief}
\noindent Changed to Correct Answer: 6/100
\begin{tcolorbox}[fontupper=\fontsize{8}{9.6}\selectfont]
Story: Noor is working as a barista at a busy coffee shop. Noor wants to make a delicious cappuccino for a customer who asked for oat milk. Noor grabs a milk pitcher and fills it with oat milk. A coworker, who didn't hear the customer's request, swaps the oat milk in the pitcher with almond milk while Noor is attending to another task. Noor sees her coworker swapping the milk.

\textcolor[RGB]{145, 100, 160}{Question: Does Noor believe the milk pitcher contains oat milk or almond milk?}

\textcolor[RGB]{255, 165, 79}{Choose one of the following:}

a) Noor believes the milk pitcher contains almond milk.

b) Noor believes the milk pitcher contains oat milk.

\textcolor[RGB]{34, 139, 34}{Correct Answer: a)}

\textcolor{red}{Unintervened: b) Noor believes the milk pitcher contains oat milk.}

\textcolor{blue}{Intervened: a) Noor believes the milk pitcher contains almond milk.}

\end{tcolorbox}

\begin{tcolorbox}[fontupper=\fontsize{8}{9.6}\selectfont]
Story: Zara is attending an art exhibition at a local gallery in her town. Zara wants to learn more about a particular abstract painting she admires. The label next to the artwork states that the painting is the work of a famous artist, and Zara reads the label carefully. A gallery staff member, realizing there was a mix-up, changes the label with the correct information attributing the painting to an emerging artist instead. Zara reads the updated label carefully.

\textcolor[RGB]{145, 100, 160}{Question: Does Zara believe the painting is the work of a famous artist or an emerging artist?}

\textcolor[RGB]{255, 165, 79}{Choose one of the following:}

a) Zara believes the painting is the work of an emerging artist.

b) Zara believes the painting is the work of a famous artist.

\textcolor[RGB]{34, 139, 34}{Correct Answer: a)}

\textcolor{red}{Unintervened: b) Zara believes the painting is the work of a famous artist.}

\textcolor{blue}{Intervened: a) Zara believes the painting is the work of an emerging artist.}

\end{tcolorbox}

\begin{tcolorbox}[fontupper=\fontsize{8}{9.6}\selectfont]
Story: Farhan is a street food vendor in a bustling market in Istanbul. Farhan wants to prepare a delicious chicken shawarma wrap for a customer who asked for extra garlic sauce. Farhan takes a bottle of garlic sauce from the shelf and pours it onto the shawarma wrap. His assistant, who is busy preparing a different wrap, accidentally swaps the garlic sauce with a bottle of hot sauce, and Farhan unknowingly pours the hot sauce onto the shawarma wrap. Farhan notices the sauce swap.

\textcolor[RGB]{145, 100, 160}{Question: Does Farhan believe the sauce he poured is garlic sauce or hot sauce?}

\textcolor[RGB]{255, 165, 79}{Choose one of the following:}

a) Farhan believes the sauce he poured is hot sauce.

b) Farhan believes the sauce he poured is garlic sauce.

\textcolor[RGB]{34, 139, 34}{Correct Answer: a)}

\textcolor{red}{Unintervened: b) Farhan believes the sauce he poured is garlic sauce.}

\textcolor{blue}{Intervened: a) Farhan believes the sauce he poured is hot sauce.}

\end{tcolorbox}

\begin{tcolorbox}[fontupper=\fontsize{8}{9.6}\selectfont]
Story: Isabela is a baker in a small Argentinian town known for its delicious pastries. Isabela wants to make a batch of alfajores, a traditional Argentinian dessert, for her bakery's customers. Isabela gets a bag of powdered sugar from her pantry to dust the alfajores. Her assistant, who was organizing the pantry earlier, accidentally placed a bag of flour in the powdered sugar's spot, and Isabela unknowingly grabs the bag of flour instead. Isabela notices the flour and puts it back.

\textcolor[RGB]{145, 100, 160}{Question: Does Isabela believe the bag she took from the pantry contains powdered sugar or flour?}

\textcolor[RGB]{255, 165, 79}{Choose one of the following:}

a) Isabela believes the bag she took from the pantry contains flour.

b) Isabela believes the bag she took from the pantry contains powdered sugar.

\textcolor[RGB]{34, 139, 34}{Correct Answer: a)}

\textcolor{red}{Unintervened: b) Isabela believes the bag she took from the pantry contains powdered sugar.}

\textcolor{blue}{Intervened: a) Isabela believes the bag she took from the pantry contains flour.}

\end{tcolorbox}

\begin{tcolorbox}[fontupper=\fontsize{8}{9.6}\selectfont]
Story: Amaury is a baker in a small village in France. Amaury wants to make pain aux raisins, a pastry filled with raisins and custard, for the customers visiting his bakery. Amaury takes a jar of raisins from the pantry to add to the pastry dough. His assistant, while tidying the pantry, accidentally replaces the raisin jar with a jar containing sultanas without realizing the difference. Amaury notices the jar swap.

\textcolor[RGB]{145, 100, 160}{Question: Does Amaury believe the jar contains raisins or sultanas?}

\textcolor[RGB]{255, 165, 79}{Choose one of the following:}

a) Amaury believes the jar contains sultanas.

b) Amaury believes the jar contains raisins.

\textcolor[RGB]{34, 139, 34}{Correct Answer: a)}

\textcolor{red}{Unintervened: b) Amaury believes the jar contains raisins.}

\textcolor{blue}{Intervened: a) Amaury believes the jar contains sultanas.}

\end{tcolorbox}

\begin{tcolorbox}[fontupper=\fontsize{8}{9.6}\selectfont]
Story: Amara is a sculptor in a small town in Greece. She wants to carve a statue of Aphrodite using a large piece of marble. Amara selects a slab of marble from her workshop, which she thinks is free from cracks and imperfections. However, a hairline crack is present in the marble slab, which goes unnoticed by her. Amara discovers the hairline crack in the marble slab.

\textcolor[RGB]{145, 100, 160}{Question: Does Amara believe the marble slab is free from cracks or has a hairline crack?}

\textcolor[RGB]{255, 165, 79}{Choose one of the following:}

a) Amara believes the marble slab has a hairline crack.

b) Amara believes the marble slab is free from cracks.

\textcolor[RGB]{34, 139, 34}{Correct Answer: a)}

\textcolor{red}{Unintervened: b) Amara believes the marble slab is free from cracks.}

\textcolor{blue}{Intervened: a) Amara believes the marble slab has a hairline crack.}

\end{tcolorbox}
\noindent Changed to Incorrect Answer: 16/100
\begin{tcolorbox}[fontupper=\fontsize{8}{9.6}\selectfont]
Story: Abhay is a young farmer in a small Indian village. He wants to water his crops with the water stored in a large clay pot. Abhay checks the water level and sees that the pot is full. Unbeknownst to him, a hole in the bottom of the pot has caused the water to leak out, leaving the pot empty. Abhay notices the hole and the water leaking out.

\textcolor[RGB]{145, 100, 160}{Question: Does Abhay believe the pot is full of water or empty?}

\textcolor[RGB]{255, 165, 79}{Choose one of the following:}

a) Abhay believes the pot is full of water.

b) Abhay believes the pot is empty.

\textcolor[RGB]{34, 139, 34}{Correct Answer: b)}

\textcolor{red}{Unintervened: b) Abhay believes the pot is empty.}

\textcolor{blue}{Intervened: a) Abhay believes the pot is full of water.}

\end{tcolorbox}

\begin{tcolorbox}[fontupper=\fontsize{8}{9.6}\selectfont]
Story: Hiroshi is a fisherman in a small coastal village in Japan. Hiroshi wants to catch some fish to provide dinner for his family. Hiroshi checks the weather forecast and sees that it will be sunny throughout the day. Suddenly, a heavy rainstorm moves in, darkening the sky and causing strong waves. Hiroshi notices the rainstorm approaching.

\textcolor[RGB]{145, 100, 160}{Question: Does Hiroshi believe the weather will be sunny or does he expect a rainstorm?}

\textcolor[RGB]{255, 165, 79}{Choose one of the following:}

a) Hiroshi believes the weather will be sunny all day.

b) Hiroshi believes there will be a rainstorm.

\textcolor[RGB]{34, 139, 34}{Correct Answer: b)}

\textcolor{red}{Unintervened: b) Hiroshi believes there will be a rainstorm.}

\textcolor{blue}{Intervened: a) Hiroshi believes the weather will be sunny all day.}

\end{tcolorbox}

\begin{tcolorbox}[fontupper=\fontsize{8}{9.6}\selectfont]
Story: Cheng is a teacher in a rural Chinese village. Cheng wants to use a projector to show a documentary about Chinese history to his students. The projector is plugged in and turned on, as Cheng has prepared it beforehand. As the wind blows outside, the power goes out, causing the projector to turn off. Cheng notices the power outage.

\textcolor[RGB]{145, 100, 160}{Question: Does Cheng believe the projector is ready to use or turned off?}

\textcolor[RGB]{255, 165, 79}{Choose one of the following:}

a) Cheng believes the projector is ready to use.

b) Cheng believes the projector is turned off.

\textcolor[RGB]{34, 139, 34}{Correct Answer: b)}

\textcolor{red}{Unintervened: b) Cheng believes the projector is turned off.}

\textcolor{blue}{Intervened: a) Cheng believes the projector is ready to use.}

\end{tcolorbox}

\begin{tcolorbox}[fontupper=\fontsize{8}{9.6}\selectfont]
Story: Fatima is a potter in a small Moroccan village known for its vibrant ceramics. Fatima wants to create a beautiful vase as a wedding gift for her cousin. Fatima has left her kiln turned off overnight to cool down. A power surge during the night causes the kiln to turn on unexpectedly, making it dangerously hot. Fatima feels the heat emanating from the kiln.

\textcolor[RGB]{145, 100, 160}{Question: Does Fatima believe her kiln is off and cool or unexpectedly hot?}

\textcolor[RGB]{255, 165, 79}{Choose one of the following:}

a) Fatima believes her kiln is off and cool.

b) Fatima believes her kiln is unexpectedly hot.

\textcolor[RGB]{34, 139, 34}{Correct Answer: b)}

\textcolor{red}{Unintervened: b) Fatima believes her kiln is unexpectedly hot.}

\textcolor{blue}{Intervened: a) Fatima believes her kiln is off and cool.}

\end{tcolorbox}

\begin{tcolorbox}[fontupper=\fontsize{8}{9.6}\selectfont]
Story: Aiko is a florist in a small Japanese town. Aiko wants to create an exquisite flower arrangement for a customer's wedding. Aiko sees a vase filled with fresh, colorful flowers in her shop. Overnight, a sudden drop in temperature causes the flowers in the vase to wilt. Aiko notices the wilted flowers in the vase.

\textcolor[RGB]{145, 100, 160}{Question: Does Aiko believe the vase contains fresh, colorful flowers or wilted flowers?}

\textcolor[RGB]{255, 165, 79}{Choose one of the following:}

a) Aiko believes the vase contains wilted flowers.

b) Aiko believes the vase contains fresh, colorful flowers.

\textcolor[RGB]{34, 139, 34}{Correct Answer: a)}

\textcolor{red}{Unintervened: a) Aiko believes the vase contains wilted flowers.}

\textcolor{blue}{Intervened: b) Aiko believes the vase contains fresh, colorful flowers.}

\end{tcolorbox}

\begin{tcolorbox}[fontupper=\fontsize{8}{9.6}\selectfont]
Story: Carlos is a farmer in a small village in Mexico. Carlos wants to harvest his corn crop before the rainy season begins. The corn plants are tall, strong, and ready for harvest in his field. A sudden and unexpected hailstorm strikes, severely damaging the corn plants. Carlos sees the damaged corn plants after the hailstorm.

\textcolor[RGB]{145, 100, 160}{Question: Does Carlos believe the corn plants are tall, strong, and ready for harvest, or does he believe they are severely damaged?}

\textcolor[RGB]{255, 165, 79}{Choose one of the following:}

a) Carlos believes the corn plants are tall, strong, and ready for harvest.

b) Carlos believes the corn plants are severely damaged.

\textcolor[RGB]{34, 139, 34}{Correct Answer: b)}

\textcolor{red}{Unintervened: b) Carlos believes the corn plants are severely damaged.}

\textcolor{blue}{Intervened: a) Carlos believes the corn plants are tall, strong, and ready for harvest.}

\end{tcolorbox}

\begin{tcolorbox}[fontupper=\fontsize{8}{9.6}\selectfont]
Story: Miguel is a chef at a popular restaurant in Mexico City. Miguel wants to prepare a delicious and authentic mole sauce for the evening's special dish. Miguel notices that the chocolate he has in the pantry is of high quality and perfect for the sauce. A sudden power outage causes the temperature in the pantry to rise, melting the chocolate. Miguel discovers the melted chocolate in the pantry.

\textcolor[RGB]{145, 100, 160}{Question: Does Miguel believe the chocolate is perfect for making the mole sauce or that the chocolate has melted?}

\textcolor[RGB]{255, 165, 79}{Choose one of the following:}

a) Miguel believes the chocolate is perfect for making the mole sauce.

b) Miguel believes the chocolate has melted.

\textcolor[RGB]{34, 139, 34}{Correct Answer: b)}

\textcolor{red}{Unintervened: b) Miguel believes the chocolate has melted.}

\textcolor{blue}{Intervened: a) Miguel believes the chocolate is perfect for making the mole sauce.}

\end{tcolorbox}

\begin{tcolorbox}[fontupper=\fontsize{8}{9.6}\selectfont]
Story: Fatima is a talented artist living in a small Moroccan village. Fatima wants to create a beautiful painting to donate for the local charity auction. She has a brand-new set of high-quality paintbrushes in her studio. While she is away, her younger brother accidentally ruins the paintbrushes by using them for a school project with glue. Fatima discovers the ruined paintbrushes upon her return to the studio.

\textcolor[RGB]{145, 100, 160}{Question: Does Fatima believe she has high-quality paintbrushes or ruined paintbrushes in her studio?}

\textcolor[RGB]{255, 165, 79}{Choose one of the following:}

a) Fatima believes she has high-quality paintbrushes in her studio.

b) Fatima believes the paintbrushes are ruined.

\textcolor[RGB]{34, 139, 34}{Correct Answer: b)}

\textcolor{red}{Unintervened: b) Fatima believes the paintbrushes are ruined.}

\textcolor{blue}{Intervened: a) Fatima believes she has high-quality paintbrushes in her studio.}

\end{tcolorbox}

\begin{tcolorbox}[fontupper=\fontsize{8}{9.6}\selectfont]
Story: Tumelo is a skilled fisherman in a coastal village in South Africa. He wants to catch fish using his large, sturdy fishing net. Tumelo inspected the net before leaving home and saw that it was in perfect condition. As he walks to the shore, a mischievous seagull pecks at the net, causing a large tear in it. Tumelo sees the seagull damaging the net.

\textcolor[RGB]{145, 100, 160}{Question: Does Tumelo believe the fishing net is in perfect condition or torn?}

\textcolor[RGB]{255, 165, 79}{Choose one of the following:}

a) Tumelo believes the fishing net is in perfect condition.

b) Tumelo believes the fishing net is torn.

\textcolor[RGB]{34, 139, 34}{Correct Answer: b)}

\textcolor{red}{Unintervened: b) Tumelo believes the fishing net is torn.}

\textcolor{blue}{Intervened: a) Tumelo believes the fishing net is in perfect condition.}

\end{tcolorbox}

\begin{tcolorbox}[fontupper=\fontsize{8}{9.6}\selectfont]
Story: Fatima is an archaeologist working on a dig site in Egypt, searching for ancient artifacts. She wants to carefully excavate a small, fragile artifact she has discovered. Fatima observes that the soil around the artifact is soft and loose. Just as Fatima is about to begin excavating, a strong gust of wind blows through the site, hardening the soil around the artifact. Fatima feels the gust of wind and notices the hardened soil.

\textcolor[RGB]{145, 100, 160}{Question: Does Fatima believe the soil around the artifact is soft and loose or hardened?}

\textcolor[RGB]{255, 165, 79}{Choose one of the following:}

a) Fatima believes the soil around the artifact is soft and loose.

b) Fatima believes the soil around the artifact has hardened.

\textcolor[RGB]{34, 139, 34}{Correct Answer: b)}

\textcolor{red}{Unintervened: b) Fatima believes the soil around the artifact has hardened.}

\textcolor{blue}{Intervened: a) Fatima believes the soil around the artifact is soft and loose.}

\end{tcolorbox}

\begin{tcolorbox}[fontupper=\fontsize{8}{9.6}\selectfont]
Story: Rohan is a potter in a small Indian village. Rohan wants to make a beautiful ceramic vase as a gift for his friend's housewarming party. Rohan selects a bag of high-quality clay in his workshop that appears to be soft and malleable. While Rohan is away for lunch, the hot sun dries the clay, making it hard and unworkable. Rohan discovers the dried, hard clay.

\textcolor[RGB]{145, 100, 160}{Question: Does Rohan believe the bag of clay is soft and malleable or hard and unworkable?}

\textcolor[RGB]{255, 165, 79}{Choose one of the following:}

a) Rohan believes the bag of clay is soft and malleable.

b) Rohan believes the bag of clay is hard and unworkable.

\textcolor[RGB]{34, 139, 34}{Correct Answer: b)}

\textcolor{red}{Unintervened: b) Rohan believes the bag of clay is hard and unworkable.}

\textcolor{blue}{Intervened: a) Rohan believes the bag of clay is soft and malleable.}

\end{tcolorbox}

\begin{tcolorbox}[fontupper=\fontsize{8}{9.6}\selectfont]
Story: Kunal is a talented potter in a small Indian village. Kunal wants to create a set of earthenware bowls as a gift for his friend's housewarming. Kunal inspects a large lump of clay, finding it to be moist and malleable. However, while Kunal is away for a short break, the sun beats down on the clay, drying it out and making it hard and brittle. Kunal notices the sun-dried, brittle clay.

\textcolor[RGB]{145, 100, 160}{Question: Does Kunal believe the clay is moist and malleable or hard and brittle?}

\textcolor[RGB]{255, 165, 79}{Choose one of the following:}

a) Kunal believes the clay is moist and malleable.

b) Kunal believes the clay is hard and brittle.

\textcolor[RGB]{34, 139, 34}{Correct Answer: b)}

\textcolor{red}{Unintervened: b) Kunal believes the clay is hard and brittle.}

\textcolor{blue}{Intervened: a) Kunal believes the clay is moist and malleable.}

\end{tcolorbox}

\begin{tcolorbox}[fontupper=\fontsize{8}{9.6}\selectfont]
Story: Amara, a skilled Nigerian weaver, is preparing to create a traditional textile for an upcoming cultural exhibition. Amara wants to use a unique combination of colorful threads that represents her heritage. Amara carefully chooses a set of vibrant, high-quality threads for her project. A curious toddler in her home finds the threads and tangles them, making them difficult to use in the weaving process. Amara notices the tangled threads.

\textcolor[RGB]{145, 100, 160}{Question: Does Amara believe the threads are strong and perfect for her textile or tangled and difficult to use?}

\textcolor[RGB]{255, 165, 79}{Choose one of the following:}

a) Amara believes the threads are strong and perfect for her textile.

b) Amara believes the threads are tangled and difficult to use.

\textcolor[RGB]{34, 139, 34}{Correct Answer: b)}

\textcolor{red}{Unintervened: b) Amara believes the threads are tangled and difficult to use.}

\textcolor{blue}{Intervened: a) Amara believes the threads are strong and perfect for her textile.}

\end{tcolorbox}

\begin{tcolorbox}[fontupper=\fontsize{8}{9.6}\selectfont]
Story: Mei is a talented chef in a small Chinese village. She wants to prepare a special dish using fresh vegetables from her garden. Mei sees that the vegetables in the garden are ripe and ready to be picked. Later, a hailstorm damages the garden, bruising and damaging the vegetables. Mei sees the hailstorm and the damage it caused to the vegetables.

\textcolor[RGB]{145, 100, 160}{Question: Does Mei believe the vegetables are ripe and ready to be picked or damaged by the hailstorm?}

\textcolor[RGB]{255, 165, 79}{Choose one of the following:}

a) Mei believes the vegetables are ripe and ready to be picked.

b) Mei believes the vegetables are damaged by the hailstorm.

\textcolor[RGB]{34, 139, 34}{Correct Answer: b)}

\textcolor{red}{Unintervened: b) Mei believes the vegetables are damaged by the hailstorm.}

\textcolor{blue}{Intervened: a) Mei believes the vegetables are ripe and ready to be picked.}

\end{tcolorbox}

\begin{tcolorbox}[fontupper=\fontsize{8}{9.6}\selectfont]
Story: Ngozi is a potter in a small Nigerian village, known for her beautiful ceramic vases. Ngozi wants to create a vase with a blue glaze to showcase at the upcoming village art exhibition. She reaches for a container of blue glaze from her shelf to apply to the vase. Earlier in the day, her apprentice accidentally switched the labels of the blue and green glaze containers, so Ngozi unknowingly picks up the container with the green glaze. Ngozi notices the green glaze.

\textcolor[RGB]{145, 100, 160}{Question: Does Ngozi believe the container holds blue glaze or green glaze?}

\textcolor[RGB]{255, 165, 79}{Choose one of the following:}

a) Ngozi believes the container holds blue glaze.

b) Ngozi believes the container holds green glaze.

\textcolor[RGB]{34, 139, 34}{Correct Answer: b)}

\textcolor{red}{Unintervened: b) Ngozi believes the container holds green glaze.}

\textcolor{blue}{Intervened: a) Ngozi believes the container holds blue glaze.}

\end{tcolorbox}

\begin{tcolorbox}[fontupper=\fontsize{8}{9.6}\selectfont]
Story: Lila is a talented violinist who lives in a bustling city in India. Today, she has an important audition for a renowned symphony orchestra. Lila picks up her violin case, which she assumes contains her valuable violin. However, her younger brother, who was playing around earlier, accidentally swapped her violin case with his identical-looking case containing a beginner's violin. Lila notices the swapped violin cases.

\textcolor[RGB]{145, 100, 160}{Question: Does Lila believe the violin case contains her valuable violin or a beginner's violin?}

\textcolor[RGB]{255, 165, 79}{Choose one of the following:}

a) Lila believes the violin case contains her valuable violin.

b) Lila believes the violin case contains a beginner's violin.

\textcolor[RGB]{34, 139, 34}{Correct Answer: b)}

\textcolor{red}{Unintervened: b) Lila believes the violin case contains a beginner's violin.}

\textcolor{blue}{Intervened: a) Lila believes the violin case contains her valuable violin.}

\end{tcolorbox}